\RequirePackage{fix-cm}
\documentclass{svjour3}                     
\smartqed  
\usepackage{graphicx}
\graphicspath{ {} }

\usepackage[numbers]{natbib}

\usepackage{multirow}
\usepackage{hhline}
\usepackage[para,online,flushleft]{threeparttable}
\usepackage{array}
\newcolumntype{P}[1]{>{\raggedright\arraybackslash}p{#1}}

\usepackage[table]{xcolor}
\usepackage{mathptmx} 
\usepackage{anyfontsize}

\usepackage{enumitem}
\usepackage{amsmath}
\usepackage{amsfonts}

 \usepackage{relsize}

\newcommand{\eg}{\emph{e.g. }}
\newcommand{\ie}{\emph{i.e. }}
\newcommand{\etc}{\emph{etc. }}

\usepackage{longtable} 
\usepackage{footnote}
\makesavenoteenv{tabular}
\makesavenoteenv{table}

\usepackage{amssymb}
\usepackage{pifont}
\usepackage{xcolor}
\definecolor{OliveGreen}{rgb}{0,0.6,0}
\definecolor{Maroon}{rgb}{0.9, 0.17, 0.31}
\definecolor{Amber}{rgb}{1.0, 0.49, 0.0}

\newcommand{\Eq}{Equation}

\newcommand{\Eqs}{Equations}

\newcommand{\fig}{Figure}

\newcommand{\sect}{Section}

\usepackage{leftidx}

\begin{document}

\title{EEGS: A Transparent Model of Emotions
}


\author{Suman Ojha         \and
        Jonathan Vitale      \and
        Mary-Anne Williams
}


\institute{$^{1}$S. Ojha \at
              \email{Suman.Ojha@uts.edu.au}          
           \and
           $^{2}$J. Vitale \at
              \email{Jonathan.Vitale@uts.edu.au}   
            \and
              $^{3}$M-A. Williams \at
              \email{Mary-Anne.Williams@uts.edu.au}\\ \\
            $^{1,2,3}$Centre for Artificial Intelligence (CAI) \\
            Faculty of Engineering and Information Technology \\
            University of Technology Sydney (UTS) \\
              Australia\\
}

\date{Received: date / Accepted: date}

\maketitle

\begin{abstract}
This paper presents the computational details of our emotion model, EEGS, and also provides an overview of a three-stage validation methodology used for the evaluation of our model, which can also be applicable for other computational models of emotion. A major gap in existing emotion modelling literature has been the lack of computational/technical details of the implemented models, which not only makes it difficult for  early-stage researchers to  understand the area but also prevents benchmarking of the developed models for expert researchers. We partly addressed these issues by presenting technical details for the  computation of appraisal variables in our previous work. In this paper, we present mathematical formulas for the calculation of emotion intensities based on the theoretical premises of appraisal theory. Moreover, we will discuss how we enable our emotion model to reach to a regulated emotional state for social acceptability of autonomous agents.  We  hope  this  paper  will  allow  a  better transparency  of  knowledge,  accurate  benchmarking and  further  evolution  of  the  field  of emotion modelling.

\end{abstract}

\section{Introduction and Background}
\label{sec:introduction}

According to appraisal theory, emotion in an individual is the result of cognitive evaluation (appraisal) of the given event or situation by the individual \cite{Ortony1990,Scherer2001,Smith1990,Roseman1984}. This theory suggests that the evaluation of a situation is performed considering a set of criteria called \emph{appraisal variables}. Appraisal variables can be considered as the basis on which a particular event is evaluated beneficial or harmful by the individual experiencing the situation, which in turn leads to a particular emotional experience as a result of the evaluation. Thus, the resulting emotional state of an individual is determined by how s/he performs the appraisal of the given event based on various appraisal variables. Since the process of appraisal is person specific and hence subjective in nature, an emotion eliciting stimulus is not always guaranteed to trigger same emotional response for two different individuals \cite{Smith1993}. For example, imagine a family with children happening to be in a nudist beach without knowing that they may experience embarrassment and concern, whereas a young couple happening in the same situation may experience curiosity and laugh about it. It is because how a person perceives and analyses the environment depends on the individual assessment of the situation, which leads to various emotional experiences. However, this is not to deny that physio-anatomic processes do not at all take part in the process of emotion generation. \citet{Lambie2002} suggest that the process of appraisal indeed can occur in two levels: (i) early first-order phenomenological evaluation, which reflects the bodily reactions to the event, and (ii) conscious second-order cognitive appraisal, which denotes higher level awareness of the situation. This view is further supported by other researchers like \citet{Frijda2005}. The earliest proponent of appraisal theory of emotion is regarded as \citet{Arnold1960}. She made a revolutionary proposal of the cognitive analysis of the process of emotional experience in individuals and hence put forward the concept of appraisal in the mechanism of emotion generation \cite{Reisenzein2006}. Influenced by this concept of appraisal, several other psychologists proposed variations of appraisal theory thereafter \cite{Frijda1986,Ortony1990,Smith1990,Scherer2001}.

\subsection{Understanding Appraisal Dynamics}
\label{sec:system>revisiting_appraisal_dynamics}

\begin{figure}[!t]
\centering
\includegraphics[width=0.98\textwidth]{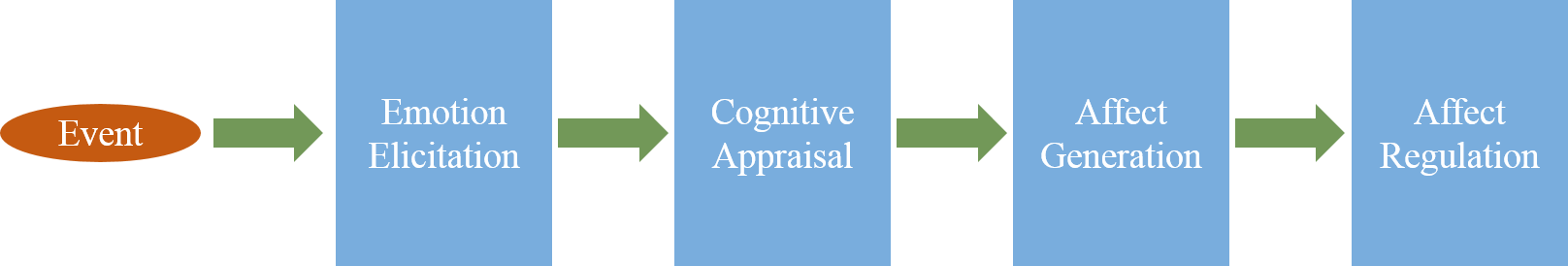}
\caption{Process flow from an stimulus event to (1) emotion elicitation, (2) cognitive appraisal, (3) affect generation, and (4) affect regulation.}
\label{fig:appraisal_dynamics_revisited}
\end{figure}

Figure~\ref{fig:appraisal_dynamics_revisited} presents an overview of the flow of various processes that occur when a stimulus event occurs in the surrounding of an individual. The overall process of emotion generation and regulation from the occurrence of an event can be divided into four main phases namely (1) \emph{Emotion Elicitation}, (2) \emph{Cognitive Appraisal}, (3) \emph{Affect Generation}, and (4) \emph{Affect Regulation}. \emph{Emotion elicitation} phase is a process of being aware of the occurrence of the event and realising whether the event has positive or negative impact on the agent (natural or artificial). This process is defined as first-order phenomenological experience by \citet{Lambie2002}. Once an individual becomes aware of the stimulus event through the non-cognitive first-order evaluation, then a second-order \emph{cognitive appraisal} of the situation is performed \cite{Lambie2002}. After the mechanism of cognitive appraisal is completed, the appraisals performed then determine the emotion intensities which may also be affected by individual specific factors such as personality \cite{Corr2008,Revelle1995} and mood \cite{Neumann2001,Morris1992}. This process is called \emph{affect generation}. Since according to appraisal theories, an event can lead to the generation of more than one emotions, the generated emotions should be regulated to allow an artificial agent to reach a single optimal emotional state. This kind of regulatory mechanism is important because emotions of the agent may have significant effect on the decision it makes -- as suggested in the literature \cite{Callahan1988, Gaudine2001}. It has been suggested by the researchers that humans normally perform a conscious and deliberative process of handling conflicting emotional states and reach to a stable and situation-congruent emotional state \cite{Gross2007}. This process is called \emph{affect regulation}. A computational model of emotion should be able to account for these basic processes underlying the emotional episode. In line with this assertion, the following sections will discuss how our model model, EEGS, performs the above mentioned processes in a computational implementation. It is important to note that, in this work, the emotion elicitation process has not been computationally realised. It has been assumed that such a process occurs in an individual as suggested by \citet{Lambie2002} and has been applied to infer signed scores to indicate the negative or positive connotation of the surrounding event. We discuss about how these scores are determined in our previous work \cite{Ojha2017a}. These signed scores are then provided to the cognitive appraisal process which provides necessary ingredients for the affect generation process and ultimately to the process of affect regulation. The latter three stages in Figure~\ref{fig:appraisal_dynamics_revisited} have been investigated and their computational details presented in this paper.

We have to admit that our primary goal in this paper is to present the internal technical details of our computational model of emotion, EEGS, to encourage a better exchange of knowledge among researchers and enable further benchmarking and advancement of the field of emotion modelling. As such, in this paper, we will not be presenting the discussion of how the various components of our model were evaluated. In the remaining of the paper, we will focus our discussion on explaining the mathematical and computational aspects of EEGS and readers will be directed to relevant publications for a better understanding of the evaluation of a component, wherever necessary.

\section{EEGS: Ethical Emotion Generation System}
\label{sec:system>EEGS_ethical_emotion_generation_system}

EEGS stands for \textbf{E}thical \textbf{E}motion \textbf{G}eneration \textbf{S}ystem which is a computational model of emotion intended to provide an ability to generate ethically-guided emotional responses and behavioural tendencies in social robots and virtual agents. By saying \emph{ethically-guided emotional responses}, we refer to the process of emotion regulation that allows an intelligent agent to have adequate control over the computed emotions and regulate the emotions based on its ability to reason ethically thereby reaching to an emotional state that is socially acceptable.

\subsection{Overall System Architecture}
\label{sec:system>overall_system_srchitecture}

\begin{figure}[!t]
\centering
\includegraphics[width=\textwidth]{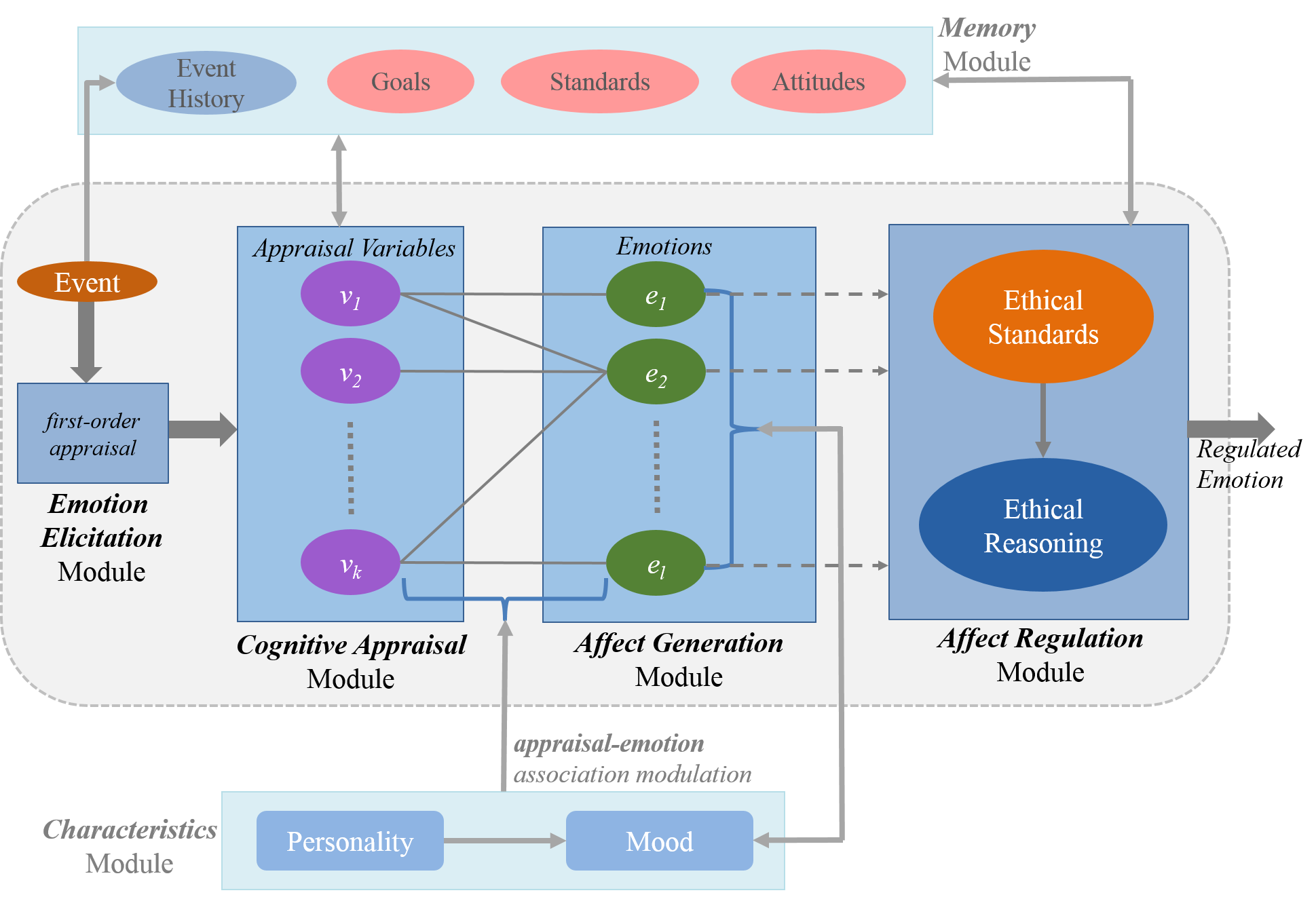}
\caption{Overall architecture of EEGS.}
\label{fig:system_architecture_EEGS}
\end{figure}


Figure~\ref{fig:system_architecture_EEGS} shows the overall relationship of various processes involved in the mechanism of emotion generation in EEGS. The four core processes involved in the emotion generation mechanism are handled by (i) \emph{Emotion Elicitation Module}, (ii) \emph{Cognitive Appraisal Module}, (iii) \emph{Affect Generation Module}, and (iv) \emph{Affect Regulation Module}. These modules should depend on other modules for their effective functioning. For example, since cognitive appraisal process needs information about the goals, standards, and attitudes of the agent \cite{Ortony1990}, (v) \emph{Memory Module} helps in the completion of this process by providing the data related to goals, standards and attitudes of the agent. Moreover, mood and personality can play a significant role in the process of emotion generation. Such person-specific factors are handled by (vi) \emph{Characteristics Module} in EEGS. Although literature suggests that the experience of emotion may be affected by several other factors \cite{Hong2000,Scollon2004,Aleman2008,Canli2009,Corr2008,Revelle1995,Neumann2001,Morris1992}, we have considered the influence of only the mood and personality factors in EEGS since these factors are more widely agreed to have influence on the process of emotion. A detailed description of the working of each module and its relationship with other modules in an agent running EEGS is presented below.

\subsubsection{Emotion Elicitation Module}
\label{sec:system>emotion_elicitation_module}

An event in the environment of an agent is first perceived by the \emph{emotion elicitation module}. This module performs the early first-order \cite{Lambie2002} lower-level \cite{Scherer2001} phenomenological appraisal of the situation based on the situational context. This is an instantaneous non-cognitive process where an individual evaluates the positivity or negativity of the perceived event as an experienced valenced bodily reaction \cite{James1884,Lange1885}. Such a first-order non-cognitive evaluation of the event provides initial information about the event to perform the conscious cognitive appraisal of the situation in relation to the agent's emotions generation. As mentioned earlier in Section~\ref{sec:system>revisiting_appraisal_dynamics}, this research does not fully realise the process of first-order reaction. Such a reaction has been operationalised as numerical scores based on the data collected from humans for specific scenarios. The details of this process is presented in our publication \cite{Ojha2017a}.

\subsubsection{Cognitive Appraisal Module}
\label{sec:system>cognitive_appraisal_module}

The first-order phenomenological information \cite{Lambie2002} received from the \emph{emotion elicitation module} is utilised by the \emph{cognitive appraisal module} where the event is assessed based on the \emph{goals}, \emph{standards}, and \emph{attitudes} of the agent -- as suggested by \citet{Ortony1990}. Once the first-order appraisal is received, \emph{cognitive appraisal module} performs a parallel processing of several appraisals (see Section~\ref{sec:system>parallel_processing_of_appraisals} to understand the parallel processing mechanism). Each of these processes compute an \emph{appraisal variable}, which is an assessment criteria to evaluate the event. Each of the appraisal variables is associated with one or more emotions and the values of these appraisal variables determine the intensities of the associated emotions (see Section~\ref{sec:system>appraisal-emotion_network} for more details). This association between appraisal variables and emotions is weighted and may be affected by several individual-specific characteristics such as personality and mood \cite{Corr2008,Revelle1995,Neumann2001}. In this research, the association between appraisal variables and emotions has been influenced by only the factors of mood and personality to limit the scope of investigation.

\subsubsection{Memory Module}
\label{sec:system>memory_module}

Memory module in EEGS plays a central role in managing the goals, standards and attitudes of the agent. Additionally, an \emph{event history} is also maintained which keeps track of the previously experienced events from a particular external agent. The goals, standards and attitudes of the agent not only affect the cognitive appraisal process but are also affected by the resulting appraisals of the agent (as indicated by double pointing arrow between \emph{memory module} and \emph{cognitive appraisal module}). In the course of interaction, the goals, standards and attitudes are updated based on the how the external agent interacts with the agent (see Section~\ref{sec:system>goals_standards_and_attitudes} for more details).

\subsubsection{Characteristics Module}
\label{sec:system>characteristics_module}

In EEGS, the role of \emph{characteristics module} is to manage the interaction of the emotion generation process with other person-specific factors such as personality \cite{Corr2008,Revelle1995} and mood \cite{Watson1985}. Although there may be other factors that influence the process of emotion generation mainly in terms of determining the intensities of various emotions \cite{Hong2000,Scollon2004,Aleman2008,Bevilacqua2011}, this research implements only the effect of personality and mood in EEGS. However, this is not to disagree that others factors do not participate in this process. By providing a specific \emph{characteristics module} to accommodate factors affecting emotion elicitation process, EEGS can be extended in future works to enable interactions with other person-specific factors that have not been the focus of current research. In EEGS, mood and personality factors take part in the process of mapping the appraisal variables generated in \emph{cognitive appraisal} module to the emotions in \emph{affect generation module}. These factors determine to what degree an appraisal variable is associated with an emotion thereby influencing the effect of an appraisal variable on the intensity of an emotion (see Section~\ref{sec:system>appraisal-emotion_network} for more details).

Since the personality of a person is considered a non-dynamic characteristic \cite{Costa1988,Dweck2008}, the personality of an agent running EEGS is not expected to change with time. Yet, mood is believed to change over the course of time with the influence of multiple emotional experiences \cite{Beedie2005,Ekman1994,Parkinson1996}. Therefore, emotions and mood in EEGS affect one another in the course of interaction (as indicated by double-headed arrow between emotions and mood in Figure~\ref{fig:system_architecture_EEGS}). As such, mood affects emotions in two ways -- one (i) \emph{while mapping the appraisal variables to emotion intensities}, and another (ii) \emph{by modulating the intensities of the emotions depending on whether the mood is congruent with the valence of the emotion}. This helps to ensure that mood congruent emotions are more likely to be activated instead of mood incongruent (see Section~\ref{sec:system>revisiting_interaction_among_emotion_mood_and_personality} for technical details of this process in EEGS).

\subsubsection{Affect Generation Module}
\label{sec:system>affect_generation_module}

The central role in generating emotions in EEGS is played by \emph{affect generation module}. We use the term `affect' instead of `emotion' because this module is responsible for the generation of emotion intensities as well as the mood of an agent. While the mood state is handled by \emph{characteristics module}, its dynamics is influenced by the emotion intensities generated in the \emph{affect generation module} (as indicated by double sided arrow between emotions and mood). Since the appraisal of an event can lead to the generation of more than one emotions with effective intensities \cite{Ortony1990,Scherer2001,Roseman1990}, EEGS may have more than one active emotion after the completion of affect generation process. It is accepted that emotional state of an individual influences the decision s/he makes \cite{Callahan1988,Gaudine2001,Isen1983}. Therefore, an artificial agent with an ability to generate emotions should reach to a final stable emotional state if it has to be inolved in a decision making task that may be influenced by its emotional state. In such a situation, if there are multiple conflicting emotions active within the agent, the agent may not be able to make a right decision. Hence, it is necessary to converge the active emotions to a stable and regulated emotional state. This process is handled by \emph{affect regulation module} discussed below.

\subsubsection{Affect Regulation Module}
\label{sec:system>affect_regulation_module}

When the \emph{affect generation module} generates more than one conflicting emotions, these emotions should be converged to a final regulated emotional state for an agent to influence its decision based on its emotional state. This is because, the agent should also ensure that the emotional state triggered in response to a surrounding event is also socially acceptable in the given context (see Section~\ref{sec:system>emotion_convergence_and_regulation} for more discussion on this). Although, researchers have previously proposed either \emph{highest intensity approach} \cite{Gratch2004Domain} or \emph{blended intensity approach} \cite{Reilly2006,Marinier2007} (see Section~\ref{sec:system>emotion_convergence_in_computational_models} for more discussion on this), we propose a better selection of emotional state can be achieved by \emph{ethical reasoning mechanism}. In EEGS, ethical reasoning process is supported by its ethical standards which is constructed by the agent in the course of interaction with external agents (see Section~\ref{sec:system>ethical_reasoning_for_emotion_regulation_in_EEGS} for technical details of this mechanism in EEGS). With the help of ethical reasoning capability, EEGS allows an agent to reach to a single emotional state that, when expressed, reduces the risks of socially unacceptable behavioural responses by the agent.

\vspace{1cm}
It is important to note that the \emph{goals}, \emph{standards} and \emph{attitudes} of an agent not only influence the process of appraisal (hence, emotion generation) and ethical reasoning, but are also influenced by these processes in return (as indicated by double-sided arrow between \emph{memory module} and \emph{cognitive appraisal module} and also between \emph{memory module} and \emph{affect regulation module}). The rationale behind this choice is that goals, standards, and attitudes of an individual are dynamic aspects and tend to change with the change in the environmental conditions and situations around the individual.

Although we do not present other modules beyond \emph{affect regulation module} in the overall architecture of EEGS presented in Figure~\ref{fig:system_architecture_EEGS}, it should be understood that the regulated emotional state output from the \emph{affect regulation module} can be used as an input to any other expressive, behavioural or decision making unit an agent may have. We have left the choice open to any researcher who may further wish to extend our emotion model and integrate with other intelligent systems. This flexibility allows EEGS to be used as a plug-and-play component of other autonomous agents where selecting a socially acceptable emotional state may be needed as an input to other cognitive capabilities.

In the remaining of the chapter, we will discuss each of the modules presented in the overall architecture (Figure~\ref{fig:system_architecture_EEGS}) in detail. Since this research is heavily inspired by the appraisal theory of \citet{Ortony1990}, which considers events, actions of agents and objects at the core of appraisal mechanism, we will start the discussion with the introduction of these concepts.

\section{Events, Actions and Objects}
\label{sec:system>events_actions_and_objects}

An event can be defined as a stimulus that has the potential to cause emotion elicitation in an individual. There can be numerous information associated with an event. For example, there can be several questions associated with an event: what happened in the event? was it likely to cause positive or negative impact on the agent? who is responsible for the incidence? when did the event occur? who was affected by the event? and many more. From the computational perspective, the notion of \emph{event} should contain sufficient information that allows an artificial agent to perform situation assessment leading to emotional and behavioural responses. In the scenario of interaction between a human and an artificial agent or in the case of interaction between an agent and another or even in the situation of multi-agent interaction, the necessary information that should be present in an event include (i) \emph{agent/person initiating the event}, (ii) \emph{details about what was involved in the event}, (iii) \emph{when did the event occur}, (iv) \emph{who was primarily affected by the event}. The information about an event allows an individual (or an artificial agent) to perform cognitive appraisal of the situation before generating situation congruent emotional states \cite{Ortony1990}. The structure of an event operationalised in EEGS is presented in detail in Section~\ref{sec:system>structure_of_events_actions_and_objects}.

The notion of events is closely related to the concepts of \emph{actions} and \emph{objects} \cite{Ortony1990}. It is because an event is usually accompanied by some action(s) and objects (persons, animals, vehicles, \etc) involved in its occurrence. An action performed by an agent A toward an agent B can have a positive or negative impact on agent B, where A and B may coincide as the same agent. For example, an action involving \emph{helping} someone is considered positive by the person being helped while an action of \emph{scolding} is considered negative by the individual being scolded. However, same action can be considered as positive or negative depending on the situation. For example, the action of \emph{helping} may also be considered negative in some situations. Consider a situation where your best friend helps your worst enemy in a difficult task. In this case, even the action of helping which would normally be considered as positive is considered negative and it is likely to cause you anger. What determines this low-level \cite{Scherer2001} first-order assessment of an action \cite{Lambie2002} is its relationship with the object involved in the event.

As mentioned before, an \emph{object} can be a person, animal, vehicle, tree, wind, \etc An individual may have certain level of \emph{familiarity} about an object. For example, in case of human object, an agent might have interacted with the person quite a number of times before thereby having a high degree of familiarity with the person. In case of non-human objects, an agent might not have recognised an object before thereby having a low degree of familiarity about the object. Moreover, an agent can build its own \emph{perception} about an object based on its experience with the object. For example, a person who continuously misbehaves with an agent for someone liked by the agent, then the agent will gradually develop negative perception about the person.

\subsection[Structure of Events, Actions and Objects]{Structure of Events, Actions and Objects\footnote{Parts of the discussion in this section have been previously published in our paper \cite{Ojha2017b}}}
\label{sec:system>structure_of_events_actions_and_objects}

In the previous section, we presented an understanding of the relationship among the events, actions and objects -- drawing the inspirations from OCC theory \cite{Ortony1990}. In this section, we will present the structural organisation of events, actions and objects in EEGS.

\subsubsection{Event Structure}

Events in EEGS are stored in the following form:

\vskip 0.1in

\centerline{{\tt (<Source>,<Action>,<Target>,<DateTime>,<OtherInformation[]>)}}

\vskip 0.1in

\noindent {\tt{Source}} denotes the object/person involved in the {\tt{Action}} within the event. It should be noted that an {\tt{Action}} is itself an object and can contain other parameters (to be dicussed soon after). {\tt{Target}} denotes the object/person who is primarily affected by the action of the {\tt{Source}}. {\tt{DateTime}} denotes the date and time the event occurred. {\tt{OtherInformation[]}} is an array that allows the storage of several other event/action/object related information. For example, familiarity and perception of the source and target after the occurrence of the event, overall impact of the event on the agent after the completion of the appraisal process, and so on. What information shall be contained in {\tt{OtherInformation[]}} depends on the needs of the agent on which our emotion model is implemented. We offer this flexibility to allow broad range of applications of the proposed system.

\begin{table}
\caption{An example of some events. The last column may contain other information like perception and familiarity of the agent with the source and target of action in the event, impact of the event on the agent, \etc}
\label{tbl:event_structure}       
\begin{tabular}{p{0.15\textwidth}p{0.15\textwidth}p{0.15\textwidth}p{0.2\textwidth}p{0.15\textwidth}}
\hline\noalign{\smallskip}
    Source & Action & Target & DateTime & OtherInformation[]\\
\noalign{\smallskip}\hline\noalign{\smallskip}
    PAUL & Kick & DAVID & 30/12/2017 10:30 & [...]\\
    JOHN & Help & KATE & 25/09/2017 20:45 & [...] \\
    JULIE & Scold & NICK & 11/07/2016 00:25 & [...] \\
\noalign{\smallskip}\hline
\end{tabular}
\end{table}

The first row in Table~\ref{tbl:event_structure} shows a record of an event in agent's memory where a person named PAUL Kicked another person named DAVID on $30^{th}$ of December 2017 at 10:30 AM. It should be noted that although {\tt{Action}} in Table~\ref{tbl:event_structure} is shown as a string (\eg ``Kick''), an action is itself an object with more than one parameters. The internal structure of an action in EEGS is explained below. 

\subsubsection{Action Structure}

Actions in EEGS are structured in the following form:

\vskip 0.1in

\centerline{{\tt (<ActionName>,<ActionValence>,<ActionDegree>)}}

\vskip 0.1in

\noindent {\tt{ActionName}} is an identifier used to denote type of action that occurred in the surrounding of the agent. For example, an action of kicking can be denoted by an identifier ``{\tt{Kick}}''. {\tt{ActionValence}} denotes whether the action is considered positive or negative. It is a binary identifier that can be {\tt{POSITIVE}} or {\tt{NEGATIVE}}. {\tt{ActionDegree}} signifies the degree of positivity or negativity of the action from {\tt{Source}} to the {\tt{Target}}. Its value can lie in the range [-1, +1], where -1 denotes extremely negative action and +1 denotes extremely positive action.

\begin{table}
\caption{An example of some actions. {\tt{ActionValence}} and {\tt{ActionDegree}} presented in the table are based on the data obtained from a survey (to be discussed later). }
\label{tbl:action_structure}       
\begin{tabular}{p{0.3\textwidth}p{0.3\textwidth}p{0.2\textwidth}}
\hline\noalign{\smallskip}
        ActionName & ActionValence & ActionDegree\\
\noalign{\smallskip}\hline\noalign{\smallskip}
    Greet & POSITIVE & 0.31\\
    Start Conversation & POSITIVE & 0.28\\
    Ignore & NEGATIVE & -0.17\\
    Kick & NEGATIVE & -0.74\\
\noalign{\smallskip}\hline
\end{tabular}
\end{table}

The {\tt{ActionValence}} and {\tt{ActionDegree}} associated with an action may have different values depending on the context. For example, the act of greeting from someone you like is likely to impact positively on the greeted person while the same action from someone you hate is likely to impact negatively. How an autonomous agent or robot would receive such a contextual information? How a robot would recognise complex physical actions like kicking in the first place, when it does not have human-like sensory channel to receive such an information? In general, how an autonomous agent would be able to appraise such actions and lead to the emotion generation? we shall address these questions and offer a solution in Section~\ref{sec:system>emotion_elicitation}. 

\subsubsection{Object Structure}

Objects in EEGS are structured in the following form: 

\vskip 0.1in

\centerline{{\tt  (<ObjectName>, <Familiarity>, <Perception>)}}

\vskip 0.1in

\noindent {\tt{ObjectName}} denotes the name of the object, {\tt{Familiarity}} is a numerical representation showing how familiar is the agent to the object in interaction and {\tt{Perception}} denotes the degree of positivity or negativity of the agent towards another agent/object. {\tt{Familiarity}} can range from 0.0 to 1.0, where 1.0 denotes complete stranger and 0.0 denotes a very familiar person (considering the notion of distance). This choice was made with an analogy that close person would not be far in distance, hence the number `0' for more familiar person. Perception can range from -1.0 to +1.0, where -1.0 denotes very negative perception and +1.0 denotes very positive perception. For example, {\tt{(``PAUL'',  0.5, -0.4)}}, denotes a person named {\tt{PAUL}} who is somewhat familiar (0.5) to the agent and the agent has a negative (-0.4) perception about him. It is important to understand that the values of familiarity and perception are dynamic and change during the interaction with the person. This change applies to the case where the agent interacts with another person/agent or the interaction involves two other persons/agents leaving the model as a mere observer of the situation \cite{Ojha2017b}. 

\begin{table}
\caption{An example of some objects. Adapted from \citet{Ojha2017b} }
\label{tbl:object_structure}       
\begin{tabular}{p{0.3\textwidth}p{0.3\textwidth}p{0.2\textwidth}}
\hline\noalign{\smallskip}
        ObjectName & Familiarity & Perception\\
\noalign{\smallskip}\hline\noalign{\smallskip}
    PAUL  & 0.5 & -0.4 \\
    JOHN & 1.0 & 0.0 \\
    ROBERT  & 0.0 & 1.0 \\
    JESSICA  & 0.3 & 0.6 \\
    ALEX & 0.1 & -0.9 \\
\noalign{\smallskip}\hline
\end{tabular}
\end{table}

Table~\ref{tbl:object_structure} shows an example of list of persons in the memory of EEGS. The person in the first row is named ``PAUL'' about whom the agent has a familiarity of 0.5 and perception of -0.4 towards the person. Initially, when a person is first introduced with the model, the person is considered stranger (i.e. familiarity = 1.0) and the model has a neutral perception (i.e. perception = 0.0) about the person. This design choice was made not to bias the model when a new person is introduced. The perception and familiarity about object changes in the course of interaction. This change is the result of continuous interaction of the object (person) with the agent thereby affecting the goals, standards and attitudes of the system, which will be explained in Section~\ref{sec:system>goals_standards_and_attitudes}.

It would be unwise to argue that the above described notion of event/action/object are complete structures that can represent all the information about a event/action/object that might be relevant to the elicitation of emotions. We acknowledge that this is only one of the many ways an event/action/object can be structured. However, in-depth discussion of the structure events, actions or objects is not in the scope of this paper because the representations presented above are sufficient to validate our hypotheses.


\section{Emotion Elicitation}
\label{sec:system>emotion_elicitation}

In Section~\ref{sec:system>events_actions_and_objects}, we presented the structural representation of events, actions, and objects in the memory of EEGS. We also established how events, actions and objects are related to each other and augment their understanding. In this section, we will discuss how the occurrence of an event leads to the elicitation of simulated emotional experience in EEGS. 

The stage of \emph{emotion elicitation} should be considered as an early process where first-order assessment of the surrounding stimulus is performed \cite{Lambie2002}. This kind of phenomenological experience can be commonly witnessed in case of humans. For example, when someone slaps you, it triggers an instant negative experience \emph{because it is painful to you}. However, in case of autonomous agents (like robots), they are not able to have such an experience because they do not have adequate sensory organs as in humans. How can robots generate phenomenological response in reaction to an stimulus event? One approach is to provide a large amount of sample data with specified labels for various actions and train the system to define an optimal values for the {\tt{ActionValence}} and {\tt{ActionDegree}} for each {\tt{ActionName}} (see action structure in Section~\ref{sec:system>structure_of_events_actions_and_objects}). However, the issue with this approach is that the system will not be able to store the contextual information since these values will get updated with every new data sample. This creates problem when same action has to be simulated in different scenarios with different context, where the emotional responses and behavioural tendencies of the agent might be non-congruent with the situation. Since, the process of emotion elicitation is not the major focus of this research, we have only provided a viable alternative to the above mentioned problem.


The valenced score\footnote{The use of the word \emph{valenced} denotes a quantity that can be positive or negative \ie has a sign.} (\ie {\tt{ActionDegree}}) associated with each action in the given context serve as the first-order phenomenological experience \cite{Lambie2002} of emotion before the occurrence of deliberate and conscious cognitive appraisal \cite{Lambie2002}. This notion may also be related to the instant bodily sensations in response to a stimulus event as explained by \citet{James1884}. For example, a positive action (\ie the action with positive value of {\tt{ActionDegree}}) can lead to bodily sensations that provide a pleasant feeling and a negative action can lead to a bad feeling. This process of attending to the stimulus event is something similar to what \citet{Scherer2001} calls as \emph{intrinsic pleasantness check} \ie the process of determining whether the event is inherently pleasant or unpleasant. Such an experience provides an opportunity to the agent to become aware of the stimulus event and determine whether the event may be significant to affect its current goals \cite{Scherer2001}. Any event that involves a significant impact on the goals of the agent would then promote a second-order deliberate and conscious process of cognitive appraisal \cite{Lambie2002,Scherer2001}.

\section{Cognitive Appraisal}
\label{sec:system>cognitive_appraisal}

\begin{figure}[!t]
\centering
\includegraphics[width=0.99\textwidth]{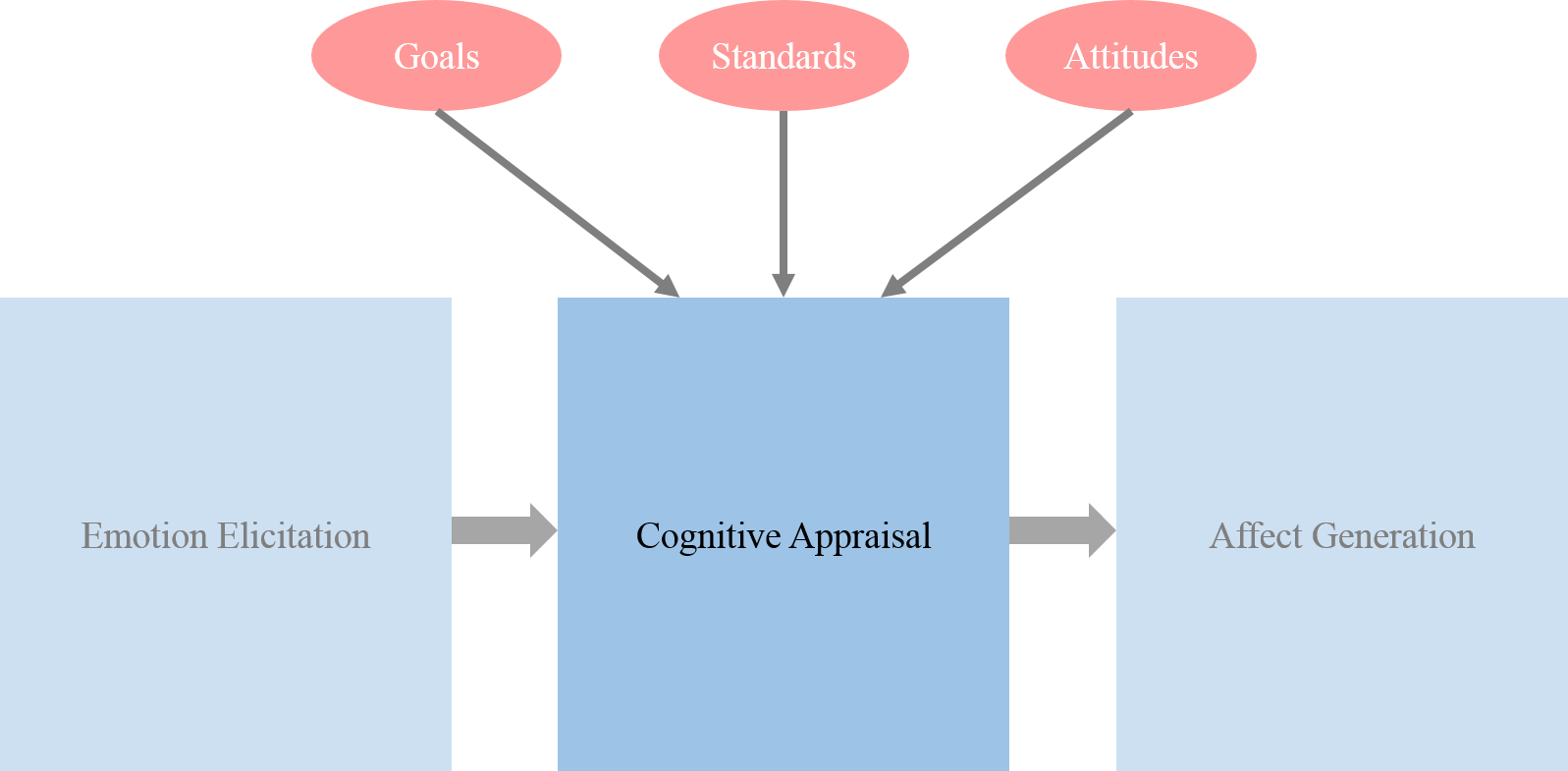}
\caption{Influence of \emph{goals}, \emph{standards} and \emph{attitudes} in cognitive appraisal process as suggested by \citet{Ortony1990}.}
\label{fig:goals_standards_attitudes}
\end{figure}

Appraisal theory is the emotion theory in psychology that relates the process of emotion generation in humans to the cognitive aspects of mind \cite{Lazarus1991, Ortony1990, Roseman1996, Scherer2001}. According to the theory, generation of emotion in an individual is a cognitive process and is determined by the way the emotion inducing situation is appraised (evaluated) by the individual. Cognitive appraisal is usually considered as \emph{second-order} \cite{Lambie2002} \emph{higher-level} \cite{Scherer2001} \emph{deliberate} \cite{Lazarus1982} and \emph{conscious} \cite{Frijda1989} process of evaluating a surrounding stimulus event. Since it is a complex cognitive process, appraisal mechanism is governed by a variety of factors. According to \citet{Ortony1990}, the process of appraisal is determined by the \emph{goals}, \emph{standards} and \emph{attitudes} of the appraising individual. As per the theory, how an event is appraised is determined by the goals, standards and attitudes of the agent. This suggests that the computation of appraisal variables (variables used for the evaluation of a situation) is affected by goals, standards and attitudes of an individual \cite{Ortony1990}. Hence, to provide our model inspired by cognitive appraisal theory, we argue that it is crucial to understand the link between appraisal variables and goals, standards and attitudes as suggested by OCC theory. Although OCC theory describes the relationship of the goals, standards and attitudes to different variables, it does not provide explicit mathematical relation to compute the values of appraisal variables, which are necessary to build a computational model. 

In the following sections, we will present the structure of the goals, standards and attitudes in EEGS and their relationships with various appraisal variables.

\subsection[Goals, Standards and Attitudes]{Goals, Standards and Attitudes\footnote{Most of the content of this section has been adapted from \cite{Ojha2017b}}}
\label{sec:system>goals_standards_and_attitudes}

According to \citet{Ortony1990}, goals, standards and attitudes are the crucial driving factors in emotion generation because these are the ``three major ingredients of appraisal [as] they constitute respectively the criteria for evaluating events, actions of agents, and objects'' \cite[p.~13]{Ortony1990}. In other words, events, actions of agents and objects can be effectively appraised in relation to emotion generation based on what the goals of the agent are, what are the standards of the agent and how is the attitude of the agent with interacting agent.

\subsubsection{Goals}
\label{sec:system>goals}
Goals represent a set of states that an individual wants to achieve. 
we use the term ``set of states'' because there can be more than one goal that an individual aims to accomplish. Moreover,  accomplishment of one goal might help in achieving another goal.  In EEGS, goals are represented in a hierarchy where a goal that helps in accomplishing another goal lies in the lower level of the hierarchy. We have represented the goals of the system as a tree structure in line with the proposal of the OCC theory. Each node of the tree is a goal node and a node may be linked to one or more lower level (child) nodes. 

\begin{figure}[!t]
\centering
\includegraphics[width=1.0\textwidth]{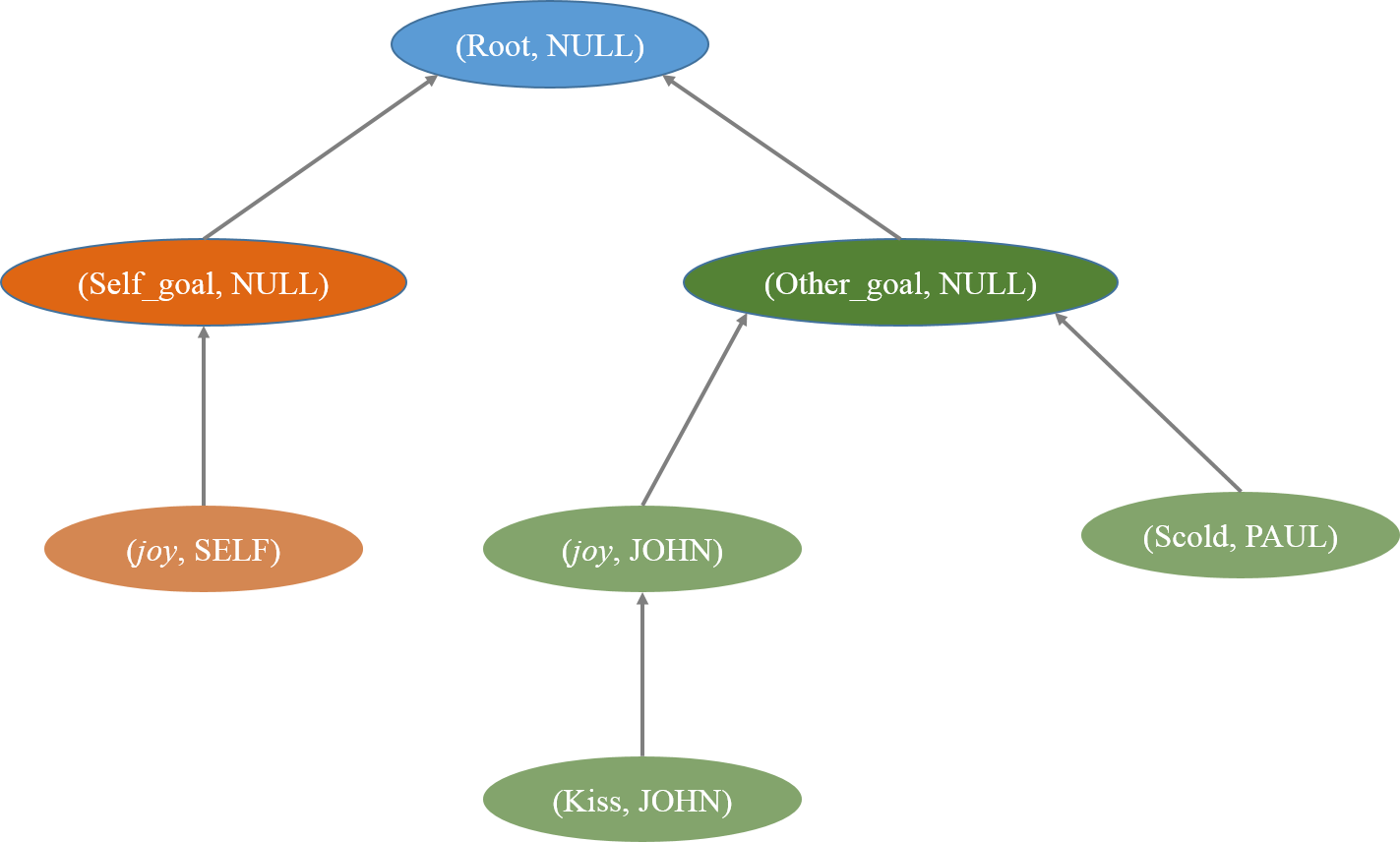}
\caption[An example of a goal tree in EEGS based on OCC theory]{An example of a goal tree in EEGS based on OCC theory \cite{Ortony1990}. Redrawn after \citet[p.~6]{Ojha2017b}.}
\label{fig:goal_tree}
\end{figure}

OCC theory of emotion has described three types of goals: (1) \emph{Active-pursuit (A) goals}, (2) \emph{Interest (I) goals} and (3) \emph{Replenishment (R) goals} \cite{Ortony1990}. \emph{A-goals} refer to the goals that an individual ``wants to accomplish''. As per the theory, such goals are discarded once they are accomplished since the person has no more desire or possibility once it is accomplished. For example, ``goal of a computer scientist to win Turing Award''. \emph{I-goals} refer to the goals that an individual ``wants to see happen'', but does not accomplish it on its own. I-goals are also discarded once they are accomplished. \emph{R-goals} refer to such goals that an individual ``wants to accomplish'' or achieve but does not stop willing to accomplish it once it is achieved. Such goals refer to regular and routing requirements of an individual. For example, a goal (desire) of eating, sleeping, etc. Such goals (desires) are not discarded once they are achieved but are needed routinely \cite{Ortony1990}. our computational model of emotion will be mainly focused on how the R-goals affect the appraisal variables that are related to goals of an individual. However, it should be noted that our mathematical formulation (to be presented following sections) can be adjusted to account for A-goals and I-goals as well.

Figure~\ref{fig:goal_tree} shows an example of a goal tree in EEGS. The goals shown in the goal tree are in the form {\tt(<Action/Emotion>, <Object>)}, where {\tt Action/Emotion} denotes the action to be done or emotional state to be attributed to a particular {\tt Object} (Person)\footnote{Since EEGS is currently intended to interact with humans only, the goals can either be an action performed to a person or an emotional state that the model wants to see in a person. But, it should be noted that this notion of goals can be extended beyond this scope without changing the computational mechanism of EEGS. Also, in line with this assumption, {\tt Object}, now on-wards, shall be considered as {\tt Person} only.}. For example, goal node, {\tt(\emph{joy}, JOHN)} aims to bring ``JOHN'' in state of ``\emph{joy}''. The root node {\tt(Root, NULL)} has two children nodes {\tt(Self\_goal, NULL)} and {\tt(Other\_goal, NULL)}, which denote the goals intended for self and for others respectively. Children of Self\_goal node are the goals that are aimed for the benefit of oneself while the children of Other\_goal node are aimed for the benefit of others. ``NULL'' Person for these goal nodes indicates that there is no specific target person -- they just divide the goals into two categories. Lower level goals are useful for the accomplishment of the higher level goals. For example, the goal {\tt(Kiss, JOHN)} helps in the accomplishment of the goal {\tt(\emph{joy}, JOHN)}. It should be noted that although, here we show {\tt Action/Emotion} and {\tt Person} as strings, the actual computational representation stores them as objects. Similar is the case with the examples in standards and attitudes to be introduced in the following sections.

\subsubsection{Standards}
\label{sec:system>standards}

Standards maintain a collection of norms and values of an individual which may be shaped by the social context or learned concepts. In EEGS, we structure standards in the form:

\vskip 0.1in
\begin{small}
\centerline{{\tt  (<Action/Emotion>, <Source>, <Target>, <Approval>)}}
\end{small}

\noindent which stores a belief that an {\tt Action/Emotion} performed/expressed by the {\tt Source} upon the {\tt Target} has certain level of {\tt Approval} as per the standard. {\tt Approval} is further broken down into the structure {\tt(<Preference>, <ApprovalDegree>)}, where {\tt Preference} indicates if the action from source to target is preferred or not and {\tt ApprovalDegree} indicates the degree of that preference.  For example, {\tt(Slap, PAUL, NEIL, (NO, 0.8))} means ``{\tt PAUL} is {\tt NO}t supposed to {\tt Slap} {\tt NEIL} and the degree of this preference is {\tt 0.8}''. {\tt ApprovalDegree} denotes how strong is the belief that an individual has on the standard. Its value can range from 0 (exclusive) to 1 (inclusive). An {\tt ApprovalDegree} of 1 indicates a very strong belief for that standard and a value close to 0 indicates a very weak belief for that standard. An example of some of the standards of EEGS is shown in Table~\ref{tbl:examples_of_standards}.

\begin{table}
\caption{An example of some emotion and action related standards in EEGS. Adapted from \citet{Ojha2017b}. }
\label{tbl:examples_of_standards}       
\begin{tabular}{p{0.15\textwidth}p{0.15\textwidth}p{0.15\textwidth}p{0.15\textwidth}p{0.15\textwidth}}
\hline\noalign{\smallskip}
        Action/Emotion & Source & Target & Preference & ApprovalDegree\\
\noalign{\smallskip}\hline\noalign{\smallskip}
    Slap & PAUL & NEIL & NO & 0.8 \\
    Scold & SELF & ROBERT & YES & 0.5 \\
    \emph{joy} & SELF & JASMINE & YES & 0.9 \\
    \emph{distress} & SELF & NEIL & NO & 0.4\\
\noalign{\smallskip}\hline
\end{tabular}
\end{table}

The structure of standard presented above may not be the only way to represent the related aspects. However, in the context of OCC theory as well as in the interaction of an agent with people/agents in social contexts, this structure of standard provides necessary ingredients for the computation of appraisal variables and hence in the generation of emotions.

Since standards contain a set of beliefs, the notion of standard should be dynamic as beliefs of a person might change in the course of life experience. For example, let us consider the example we presented in the previous paragraph.  The standard {\tt(Slap, PAUL, NEIL, (NO, 0.8))} might be changed if NEIL does some severely bad action to PAUL. In such a case, the standard might rather change to {\tt(Slap, PAUL, NEIL, (YES, 0.2))}, which means ``it is okay for {\tt PAUL} to {\tt Slap} {\tt NEIL} and the degree of this preference is {\tt 0.2}''. We have maintained this idea of dynamic standards in our computational model that change during the course of interaction with human agents.

It should be noted that an individual (and hence the presented emotion model) can have many recognised persons, many recognised actions and many possible emotions. An individual's standards should account for all of those aspects. The list of standards in Table~\ref{tbl:examples_of_standards} is not exhaustive, it only shows a few representative examples for the understanding of how the standards are structured in EEGS. Moreover, when EEGS runs for the first time, it starts with empty standards. It keeps on building and updating the standards as it interacts with various persons. This makes EEGS completely independent of the implementation domain and can build on its own as per the environmental context.

\subsubsection{Attitudes}
\label{sec:system>attitudes}

Attitudes defined in OCC theory \cite{Ortony1990} can be considered as perception of an individual regarding persons or objects. But unlike the standards, attitudes in EEGS have a slightly different structure. An attitude is structured as {\tt(<Person/Object>, <Perception>)}, where {\tt Person/Object} refers to the person or object about whom the attitude is and {\tt Perception} is the perception about the {\tt Person/Object}. For example, the attitude {\tt(JOHN,0.8)} means ``the model has positive perception about {\tt JOHN} and the degree of the positivity is {\tt 0.8}''. It is important to note that even if the person ``JOHN'' is shown as string in the above example, it is actually represented as an object of class {\tt Person} within computational model. As denoted earlier in the discussion about the structure of an object, {\tt Perception} about a {\tt Person} in EEGS can range from -1 to +1, where -1 indicates an extremely negative perception and +1 indicates extremely positive perception.

\subsection{Appraisal Variables}
\label{sec:system>appraisal_variables}

Cognitive appraisal of an event is achieved through a set of criteria called appraisal variables \cite{Ortony1990,Scherer2001,Roseman1984,Smith1990} which are governed by the goals, standards and attitudes, which were discussed in previous sections. Different theorists have proposed a different set of appraisal variables (see, for example \cite{Ortony1990,Scherer2001,Roseman1990}). Moreover, there are a lot of common appraisal variables among various appraisal theories. The goal of this research is not to determine which set of appraisal variables is valid or sufficient for cognitive appraisal of a situation. So, we will not indulge into the discussion of that matter. In this paper, we aim to present a domain-independent approach of computing appraisal variables irrespective of the appraisal theory used. Although, appraisal theories have recently been recognised as being able to provide a theoretical foundation for achieving domain independence in computational modelling of emotions \cite{Gratch2015}, most existing computational emotion models based on appraisal theory have implemented rule-based domain specific designs \cite{Aylett2005, Dias2005, El2000, Velasquez1997} to achieve cognitive appraisal for the generation of emotion. This is a significant problem because using domain specific pre-defined rules makes it difficult to reuse the model in other domains. We aim to resolve this issue through our research and present a mechanism to appraise events in domain-independent manner.

\subsubsection{Computation of Appraisal Variables}
\label{sec:system>computation_of_appraisal_variables}

According to appraisal theory, emotions are the result of appraisal of a particular situation or event happening in an individual's surrounding \cite{Ortony1990}. Thus, whenever an event occurs, an agent does the evaluation of the situation using several appraisal variables (based on the theory used) and the resulting values of the appraisal variables cause the generation of various emotions. The numeric value of most appraisal variables in EEGS range from the value of -1.0 to +1.0, which is only a design choice and we believe other alternatives should be equally effective (say for example, -100 to +100). The value of +1.0 for appraisal variable, say \emph{desirability}, indicates that a particular event is extremely desirable while the value of -1.0 indicates that the event is extremely undesirable. In the following sections, we will discuss in detail how the numerical values of various appraisal variables are calculated in EEGS. Currently, EEGS is able to compute seven appraisal variables namely \emph{goal conduciveness}, \emph{desirability}, \emph{praiseworthiness}, \emph{appealingness}, \emph{deservingness}, \emph{familiarity} and \emph{unexpectedness}.

\vspace{0.75cm}
\noindent
\textbf{\emph{Desirability}}

\noindent Desirability is the measure of how desirable a particular situation or event is to the appraising individual. In order to evaluate the desirability of an event, it is compared to the goals of the individual \cite{Ortony1990}. If the event is likely to help in achieving goals, then the event is said to be desirable. However, if the event is likely to hinder the accomplishment of the goals, then the event is said to be undesirable. The degree of desirability or undesirability depends on the degree the event helps or hinders the achievement of the goals. An event may not be related to all the goals in the current goal tree (see Section~\ref{sec:system>goals} for the detailed structure of goals and goal tree considered in this research). Desirability of an event in EEGS is computed based on the overall effect the event has on the accomplishment of all the goals in the goal tree. This is determined by considering whether the event is relevant to the goal or not. Before calculating the OCC appraisal variable desirability, we compute a value called \emph{goal conduciveness}, which is an appraisal variable adapted from Scherer's theory of appraisal \cite{Scherer2001} for calculating the degree to which the event helps or hinders the achievement of a particular goal node that is related to the event. When the conduciveness of each goal is calculated, then the numerical value of desirability is computed as the average conduciveness of all the goals in the goal tree.

Suppose there are $N$ goal nodes in the goal tree. If we denote the degree of the action\footnote{In EEGS, an action like \emph{slapping} is considered to have negative degree and an action like \emph{appreciating} is considered to have positive degree. The numeric value of degree of an action depends on how positive or negative the action is. This input value is considered as the result of first order evaluation in line with the arguments of \citet{Lambie2002}. See Section~\ref{sec:system>emotion_elicitation} for more details on this.}/emotion defined in the $i^{th}$ goal node as $d_{g_i}$ $\in$ [-1, 1]; the degree of the action in the recent event that is relevant to the $i^{th}$ goal node as $d_{e_i}$ $\in$ [-1, 1]; height of the $i^{th}$ goal node from root node in the goal tree as $h_i$, then  conduciveness ($GC_i$) of $i^{th}$ goal in the goal tree is given by the following equation.

\begin{equation}
\label{eqn:goal_conduciveness}
	GC_i= 
	\begin{cases}
		1 - \frac{\left|\left|d_{g_i}\right| - \left|d_{e_i}\right|\right|}{h_i} & \mbox{if } sign(d_{g_i}) = sign (d_{e_i}), \\ & \mbox{or} \ d_{g_i}= d_{e_i}=0 \\
		\frac{\left|\left|d_{g_i}\right| - \left|d_{e_i}\right|\right|}{h_i} -1 & \mbox{if } sign(d_{g_i}) \neq sign(d_{e_i})\\
		-\frac{\left|d_{e_i}\right|}{h_i} & \mbox{if } d_{g_i} = 0 \mbox{ \& } d_{e_i} \neq 0\\
		-\frac{\left|d_{g_i}\right|}{h_i} & \mbox{if } d_{g_i} \neq 0 \mbox{ \& } d_{e_i} = 0

	\end{cases}
\end{equation}

\noindent
Where,\newline		
$sign(.)$ is a sign function.

\vspace{0.5cm}
The signed numeric value of $d_{e_i}$, which denotes the degree of the action in the event, is the input received by the cognitive appraisal component when an event occurs. This phenomenon represents what \cite{Lambie2002} consider as the first-order (lower-level) evaluation of an emotion inducing situation, which is performed by \emph{emotion elicitation} component (see Section~\ref{sec:system>emotion_elicitation} for more discussion on emotion elicitation). The signed numeric value of $d_{g_i}$ in a goal node is analogous to the \emph{expected utility} of the achievement of the goal \cite{Gratch2004}. The formula in \Eq~\eqref{eqn:goal_conduciveness} results in a numeric value between -1 and 1 which indicates the degree by which the event helps in attaining the $i^{th}$ goal in the goal tree. A positive value of $GC_i$ indicates that the event helps in achieving the $i^{th}$ goal while a negative value indicates that the event hinders the accomplishment of the goal. Goal conduciveness basically computes the signed deviation of the event from the goal. The reason for dividing this quantity by the height of the node from the root, in \Eq~\eqref{eqn:goal_conduciveness}, is the assumption that if a goal node is closer to the root, its achievement will have more effect on the desirability than a goal node which is farther from the root node. When the conduciveness of each goal in the goal tree is computed, the value of desirability, here denoted with \emph{desi}, is computed as the average goal conduciveness using the equation below. The numeric value of \emph{desi} indicates the degree by which the event is considered (un)desirable in relation to the goals of the individual.

\begin{equation}
\label{eqn:desirability}
desi= \frac{\sum_{i=1}^{N}GC_i}{N}
\end{equation}

\noindent
Where,\newline 	
$N$ is the total number of goal nodes in the goal tree.
		

\vspace{0.75cm}
\noindent
\textbf{\emph{Praiseworthiness}}

\noindent While the appraisal variable \emph{desirability} is measured based on goals, the variable \emph{praiseworthiness} is computed based on the standards \cite{Ortony1990}. An action is considered praiseworthy if it matches closely with the standards of the individual and blameworthy (negative value of the variable praiseworthiness) if it deviates from the standard(s). Praiseworthiness is evaluated by comparing the degree of an action performed by an external agent with the approval degree of that particular action from the given source to the target in the standards of the agent (see Table~\ref{tbl:examples_of_standards} for an idea on how an standard is denoted in this discussion context). If we denote the degree of the action in the event as $d_e$ $\in$ [-1, 1]; the approval degree for the action in a given standard as $d_a$ $\in$ (0, 1], then, praiseworthiness, denoted here with \emph{prai}, is computed using the formula in \Eq~\eqref{eqn:praiseworthiness}.

\begin{equation}
\label{eqn:praiseworthiness}
prai = \begin{cases} 
		\mbox{for} \ d_e < 0;  & \begin{cases}-(d_e * d_a) & \mbox{if} \ pref = YES\\
					 d_e * d_a & \mbox{if} \ pref= NO \end{cases} \\
		\mbox{for} \ d_e > 0; & \begin{cases}d_e * d_a & \mbox{if} \ pref = YES\\
					 -(d_e * d_a) & \mbox {if} \ pref = NO \end{cases}\\
		\mbox{for} \ d_e = 0; & \begin{cases}d_a & \mbox{if} \ pref = YES\\
					 -d_a & \mbox{if} \ pref = NO \end{cases} 
\end{cases}
\end{equation}

\noindent
Where,\newline	
$pref$ is the preference of the action in the standard. $pref$ can be ``YES'' if the action from the given source to target is preferred and ``NO'' if the action is not preferred.
		
The formula in \Eq~\eqref{eqn:praiseworthiness} considers the degree of the action performed, and preference and approval degree for the action from the source to the target in the standard. The value of degree ($d_e$) can range from -1 to +1, where -1 indicates very negative action and +1 indicates very positive action. The value of approval degree ($d_a$) can lie in the range (0, 1]. Approval degree is said to be maximum if it is equal to 1 and minimum if it has a value just greater than 0. $pref$ indicates if the action is preferred action or non-preferred action as per the standard. The formula in \Eq~\eqref{eqn:praiseworthiness} provides a signed numeric value \emph{prai} which denotes how praiseworthy is the action of an external agent as per the standards of the appraising agent.



\vspace{0.75cm}
\noindent
\textbf{\emph{Appealingness}}

\noindent The appraisal variable \emph{appealingness} measures how appealing (likeable) is the person/object to the appraising individual. In EEGS, \emph{appealingness} is determined based on the perception of the model about the person interacting with it. A person who has done nice things in the past is considered to be appealing while a person who has done bad things is not and the degree by which such a perception is maintained depends on the frequency and intensity of the experience. AS such the numeric value of \emph{appealingness}, denoted here with \emph{appe} is given by the following formula.

\begin{equation}
\label{eqn:appealingness}
	appe = object\_perception
\end{equation}

\noindent
Where,\newline
$object\_perception$ $\in$ [-1, 1] is the numeric value of perception about the person/object the model has in its memory (see Section~\ref{sec:system>events_actions_and_objects} for details about how perception of object is stored in agent's memory).
	
	

\vspace{0.75cm}
\noindent
\textbf{\emph{Deservingness}}

\noindent Deservingness is the measure of whether someone deserved to experience an incident or receive an action from someone else \cite{Ortony1990}. People normally determine someone's deservingness of something based on their past deeds. For example, a person who has done numerous good deeds in the past is considered to deserve something good while a person who is evil and does a lot of harm is considered non-deserving for experiencing good things in return. We apply similar assumptions in calculating the appraisal variable \emph{deservingness}. Attribution plays a critical role in the computation of deservingness. For example, anyone would believe to deserve good things or at least thinking to not expecting bad things from happening. Therefore, when the target of the action in an interaction is the agent itself, the agent would consider the good actions to be deserving and bad actions to be non-deserving to itself. Likewise, when it comes to the determination of deservingness of an action/event in relation to another agent (considering a scenario of interaction between two external agents/humans), the model would consider the past actions of the interaction history in its memory. Accordingly, if an external agent has done good deeds to the agent in the model or other external agents liked by the agent in the model, then the model would consider that the external agent deserves to be treated well. If the external agent had done bad instead of good in the past, then the reverse would happen. This discussion leads to the inference that from the agent's perspective: (i) whether `\emph{it}' deserved something to happen depends only on \emph{how positive or negative the event/action is}, and (ii) whether an `\emph{external agent}' deserved something to happen depends on \emph{how positive or negative the event/action is in relation to the past deeds of the external agent}. We suggest the following formula in \Eq~\eqref{eqn:deservingness} for deserviness, denoted by \emph{dese}:

\begin{equation}
\label{eqn:deservingness}
	dese = \begin{cases} 
        \begin{array}{lll}
        \mbox{if} \ target = ``SELF'', & d_e \\
        Otherwise, & d_e + ( \ pi_{pos} + pi_{neg}) + \\ & ( \ pi_{SELF_{pos}} + pi_{SELF_{neg}}) & \mbox{if} \ v_a = POSITIVE\\
        &  d_e - ( \ pi_{pos} + pi_{neg}) - \\ & ( \ pi_{SELF_{pos}} + pi_{SELF_{neg}}) & \mbox{if} \ v_a = NEGATIVE\\
        \end{array} \\
    \end{cases}
\end{equation}

\noindent
Where,\newline
$target$ is the target of action in the interaction,\newline
$d_e$ is the degree of action in the event, \newline
$pi_{pos}$ is the aggregate impact (positive value) of past positive actions from current \emph{target} to the \emph{source} of action,\newline
$pi_{neg}$ is the aggregate impact (negative value) of the past negative actions from \emph{target} to the \emph{source} of action,\newline
$pi_{SELF_{pos}}$ is the aggregate impact (positive value) of past positive actions from current \emph{target} to the \emph{SELF},\newline
$pi_{SELF_{neg}}$ is the aggregate impact (negative value) of the past negative actions from \emph{target} to the \emph{SELF}, and \newline
$v_a$ is the valence of the action in current event.

\vspace{0.5cm}
A reader may be confused about the use of the quantities $pi_{SELF_{pos}}$ and $pi_{SELF_{neg}}$ for the calculation of deservingness even if the target is not `SELF' ( as shown in \Eq~\eqref{eqn:deservingness}). However, these quantities are necessary to signify a natural bias of personal experience a person might feel while measuring deservingness for someone. For example, consider a situation where Paul and Carl work in a company. Carl gets a promotion for his extraordinary outcomes during the year. In a situation where Paul and Carl do not know each other, Paul, by looking at Carl's achievements, will think that Carl deserves the promotion. However, if Paul and Carl know each other and they are in constant competition, with Carl having prevented Paul for getting such promotion by means of dishonest practices, Paul will think Carl does not deserve such promotion. To understand this phenomenon, let us look at the formula in \Eq~\eqref{eqn:deservingness}. In the case where \emph{target} is not the SELF, and the valence of the action $v_a$ is `POSITIVE', considering the above situation of \emph{rewarding the guy} would have a positive value for degree of action in the event $d_e$ and the value of $( \ pi_{pos} + pi_{neg})$ will be zero because Carl and Paul have not met before. But, since Carl's deceiving behaviour has been unhelpful towards the appraising individual (Paul in this case), the value of $pi_{SELF_{pos}}$ will be zero and the value of $pi_{SELF_{neg}}$ will be some negative value. Now, the resulting value of $dese$ will be $d_e + pi_{SELF_{neg}}$. If the absolute value of $d_e$ is higher than or equal to that of $pi_{SELF_{neg}}$, the resulting deservingness will be a non-negative value; however, if that is not the case then the resulting deservingness will be a negative value signifying that Carl did not deserve to get the reward.

\vspace{0.75cm}
\noindent
\textbf{\emph{Familiarity}}

\noindent The appraisal variable \emph{familiarity} measures how well known the person (or object) is to the model. In case of humans, if a person interacts with somebody never encountered before, that person would have no familiarity at all. Indeed, people use the word \emph{stranger} to denote an individual about whom the appraising individual has no familiarity at all. Our familiarity with a person increases as we interact with that person or observe that person interacting with others. 



Unlike, the appraisal variable \emph{appealingness} which shows bidirectional dynamics, \ie increases with the positive experience with a person and decreases with the negative experience with a person, \emph{familiarity} change is unidirectional, \ie you become more and more familiar no matter the experience was positive or negative. In EEGS, familiarity, denoted here with \emph{fami} is determined by the following formula.

\begin{equation} 
\label{eqn:familiarity}
	fami = relationship\_distance
\end{equation}

\noindent
Where,\newline
$relationship\_distance$ $\in$ [0, 1] is the numeric value that denotes how closely the model has known the person/object.

\vspace{0.75cm}
\noindent
\textbf{\emph{Unexpectedness}}

\noindent Appraisal variable \emph{unexpectedness} measures how expected/unexpected a particular event was. In the context of interaction between two persons, a person would expect a behaviour that matches with how the other person behaved him/her in the past. As such, a sudden slapping from a best friend with a lot of fury is an example of an extremely unexpected event when one does not have any idea of wrongdoing. In EEGS, like \emph{familiarity}, the value of the variable \emph{unexpectedness} also lies in the range [0, 1]. But, unlike \emph{familiarity}, where 0 means extremely familiar, in case of \emph{unexpectedness}, 0 means fully expected and 1 means extremely unexpected. In EEGS, \emph{unexpectedness}, denoted here with \emph{unex}, is computed using the formula below.

\begin{equation}
\label{eqn:unexpectedness}
	unex = |d_{e_{avg}} - d_e|
\end{equation}

\noindent
Where,\newline
$unex$ is the unexpectedness of the event,\newline
$d_{e_{avg}}$ $\in$ [-1, 1] is the numeric value that denotes average degree of all the actions from \emph{source} to the \emph{target} in the past, and\newline
$d_e$ $\in$ [-1, 1] is the numeric value that denotes degree of action in the current event.

\vspace {0.5 cm}
As such, $d_{e_{avg}}$ is calculated using the following formula.

\begin{equation}
\label{eqn:average_degree}
    d_{e_{avg}} = \frac{\sum_{i=1}^{N}d_{e_i}}{N}
\end{equation}

\noindent
Where,\newline
$d_{e_i}$ is the degree of the action from \emph{source} to the \emph{target} in the $i^{th}$ event in the memory,\newline
$N$  is the number of events in memory that represent an interaction from \emph{source} to the \emph{target}.

Let us consider the above example to understand the working of \Eq~\eqref{eqn:unexpectedness}. Since the action of slapping has an inherent negative degree associated with it, the value of $d_e$ will be negative. But, a best friend would have done a lot of help and given positive experience to a person. So, the value of $d_{e_{avg}}$ will be positive. Therefore, the absolute difference of $d_{e_{avg}}$ and $d_e$ will be a larger quantity signifying a high degree of unexpectedness. However, if the action was something positive -- say hugging, which would have positive degree. Since, both values are positive the subtraction will decrease the value of the quantity making the event more expected compared to the previous situation.

\vspace{0.5cm}
In this paper, we have presented a detailed computation mechanism of seven appraisal variables. It should be noted that this list of appraisal variables is not exhaustive -- not even in relation to OCC theory \cite{Ortony1990}. The question of ``how many appraisal variables does a model implement?'' is not important. Instead the question of ``whether an autonomous agent implementing an emotion model shows coherent and consistent appraisals'' is more important. The number and types of appraisal variables implemented in a model depends on the intended application of the model. For example, EMA \cite{Gratch2004Domain,Marsella2009} models only three appraisal variables \ie \emph{desirability}, \emph{likelihood} and \emph{coping potential} \cite{Gratch2004} but it is still considered one of the most plausible and influential models of emotion because of its ability to perform cognitive appraisal in a domain-independent manner \cite{Dias2005,Becker2008}. Likewise, EMIA \cite{Jain2015} models only five appraisal variables namely \emph{desirability}, \emph{expectedness}, \emph{outcome probability}, \emph{suddenness}, \emph{cause harm} which were sufficient to achieve the goals of the scenarios simulated in their model. Although, currently, we have implemented only seven appraisal variables, EEGS is flexible enough to account for additional appraisal variables should the mechanism for computation of those variables be available (we will discuss more on this in Section~\ref{sec:system>appraisal-emotion_network}).

\begin{figure}[!t]
\centering
\includegraphics[width=0.99\textwidth]{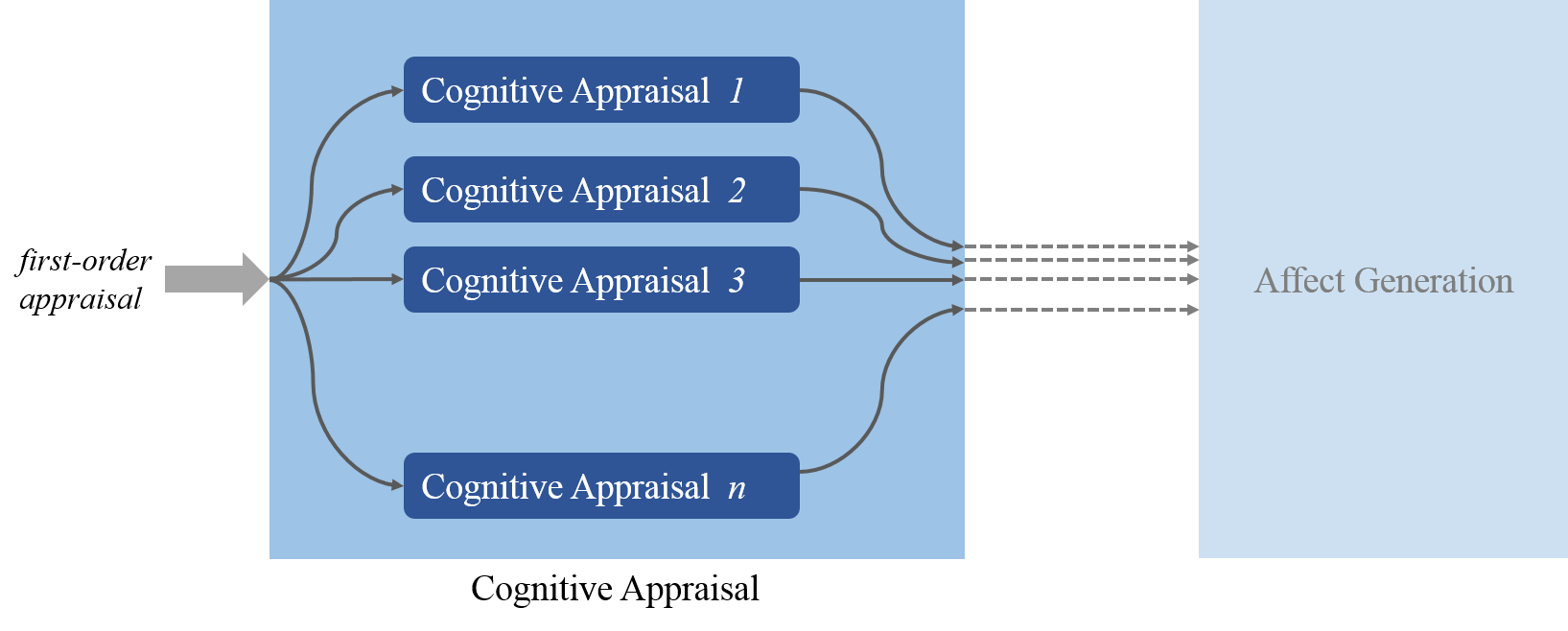}
\caption{Parallel computation of appraisals in EEGS.}
\label{fig:parallel_appraisals}
\end{figure}

\subsubsection{Parallel Processing of Appraisals}
\label{sec:system>parallel_processing_of_appraisals}

\citet{Scherer2001} suggests that appraisal computation follows a sequential pattern and one particular appraisal can not commence until a pre-requisite appraisal check is not completed. For example, the theory of \citet{Scherer2001} assumes that the appraisal of \emph{control check} can only be performed after \emph{goal/need conduciveness check}. However, this proposition is often criticised by psychologists \cite{Ortony1990,Smith2001} as well as computer scientists in affective computing field \cite{Marsella2009}. In EEGS, we follow the assumptions of \citet{Smith2001} that appraisals run in parallel and which appraisal is computed first depends on the complexity of the situation and related history in the appraising individual's memory.

The \emph{first-order} non-cognitive appraisal performed by \emph{emotion elicitation} component (not shown in Figure~\ref{fig:parallel_appraisals}) is received by the \emph{cognitive appraisal component}. The cognitive appraisal component performs various appraisals in parallel (as shown in Figure~\ref{fig:parallel_appraisals}). As stated earlier, which appraisal variable gets computed first depends on the complexity of current \emph{goals}, \emph{standards}, and \emph{attitudes} of EEGS. The individual appraisals then directly influence the corresponding emotions and their intensities in \emph{affect generation} component. Where more than one appraisal variables influence the same emotion, an incremental effect is applied on the emotion intensity. Since, different appraisals may complete at different times, the emotional states keep fluctuating until all the appraisals are determined for the given stimulus event. Such intermediate emotions may cause conflicting affective states to be experienced by the agent which need to converge to a stable emotional state. How such a regulation of emotional states is achieved in EEGS will be discussed in Section~\ref{sec:system>emotion_convergence_and_regulation}.

\section{Affect Generation}
\label{sec:system>affect_generation}

In Section~\ref{sec:system>cognitive_appraisal}, we presented how the mechanism of cognitive appraisal occurs in EEGS and how various appraisal variables are computed. According to appraisal theories, appraisal variables are responsible for the determination of emotions and their intensities in response to the stimulus event \cite{Ortony1990,Scherer2001,Smith1990,Roseman1990}. However, emotion is not an isolated phenomenon. Literature suggests that the mechanism of emotion processing is influenced by individual-specific factors such as \emph{personality} \cite{Corr2008,Revelle1995,Watson1997} and \emph{mood} \cite{Morris1992,Neumann2001}. It is desirable to model the influence of such factors in emotion generation mechanisms of artificial agents, since in practical applications, an intelligent agent should exhibit some difference in emotional response and hence action tendency if it is to be employed in wide range of human-centred situations. For example, an intelligent agent intended to be employed as a personal development assistant is desirable to have an ``organised and systematic'' characteristic (conscientiousness). As such the agent might have to express disappointment or similar emotions if the person under training ignores some routine activity. However, if the agent is to be deployed as an emotional support companion, then it is preferred to forgive such a minor ignorance -- hence it is desirable to have an easy going nature (agreeableness). Similarly, mood can also play an important role in modulating the emotional responses of the agent. For example, if an agent is in a very good mood state (based on recent experience), it might easily forgive an insult from the human interacting with it, while it may not do so if it is in very bad mood. The above examples show only a few of most situations where personality and mood can substantially influence the process of emotion generation. Hence, it is crucial to integrate the relationship of mood and personality in emotion processing mechanism of an intelligent agent. In order to realise such an effective integration of personality and mood in a model of emotion generation, we propose an \emph{appraisal--emotion network} modulated by the factors of personality and mood, which will be detailed in Section~\ref{sec:system>appraisal-emotion_network}.

\begin{figure}
\includegraphics[width=0.99 \textwidth]{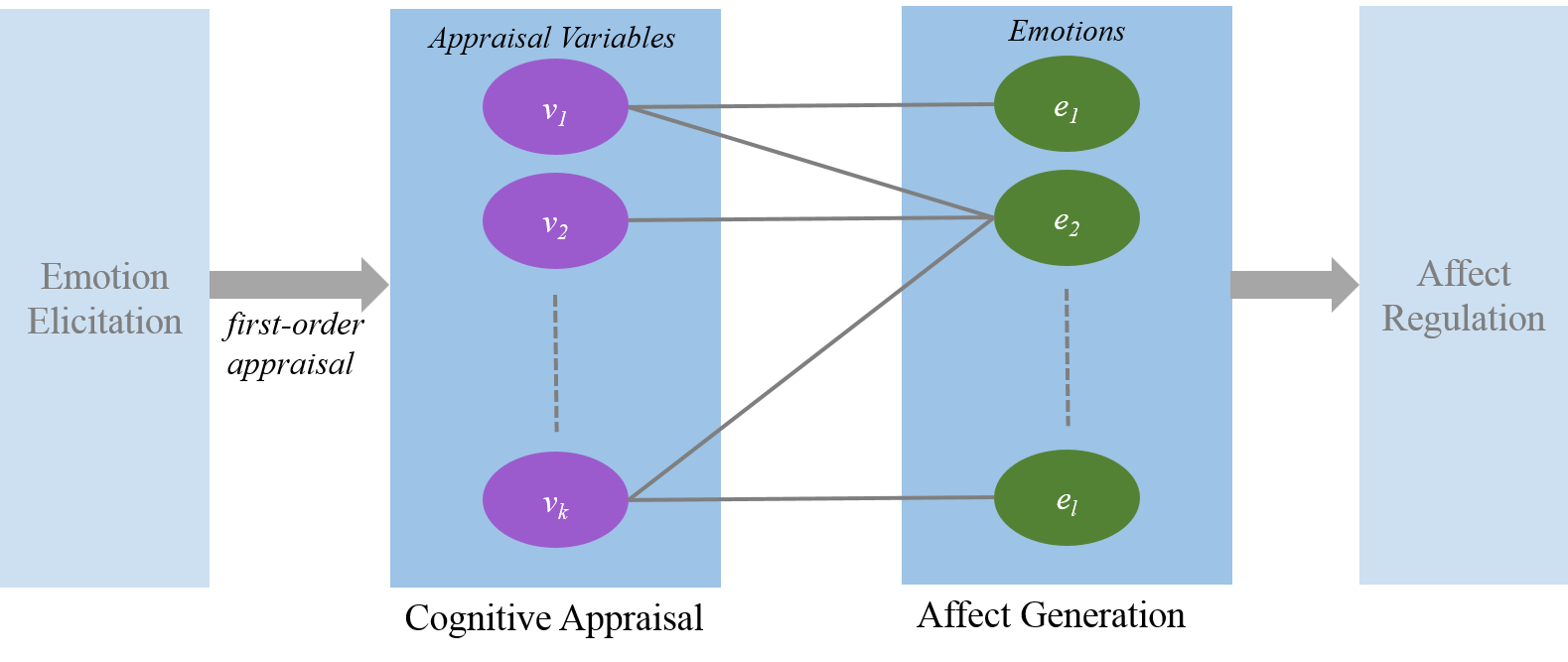}
\caption[Role of cognitive appraisal in affect generation process]{Role of cognitive appraisal in affect generation process. The appraisal variables computed as a result of the cognitive appraisal process help in determining the intensities of the associated emotions.}
\label{fig:affect_generation}
\end{figure}

\subsection{Appraisal--Emotion Network}
\label{sec:system>appraisal-emotion_network}

Once the appraisal variables are computed, the next step for an agent is to exhibit the situation congruent emotional response(s). This objective can be achieved by considering the previously computed appraisal variables for that situation. Studies in emotion research widely support the idea that an individual may generate more than one emotion with varying intensities in response to a single event \cite{Ortony1990,Scherer2001}. This means that an event can cause both \emph{joy} and \emph{distress} at the same time but with different intensities. For example, consider a situation in which you hear a news that your friend met with a fatal car accident in which the friend was injured (but alive) and the car was completely destroyed. This news might give you a mixed feeling. Distress that your friend met with an accident, joy that your friend is alive and safe while at the same time some distress knowing that your friend's new car is completely damaged. Now, suppose an artificial agent faces a situation in which conflicting emotions may be triggered. Although the agent can evaluate the situation using the appraisal functions that return a set of appraisal variables, the agent does not yet know how those variables may be associated with various possible emotions.

\begin{figure}
\includegraphics[width=0.8 \textwidth]{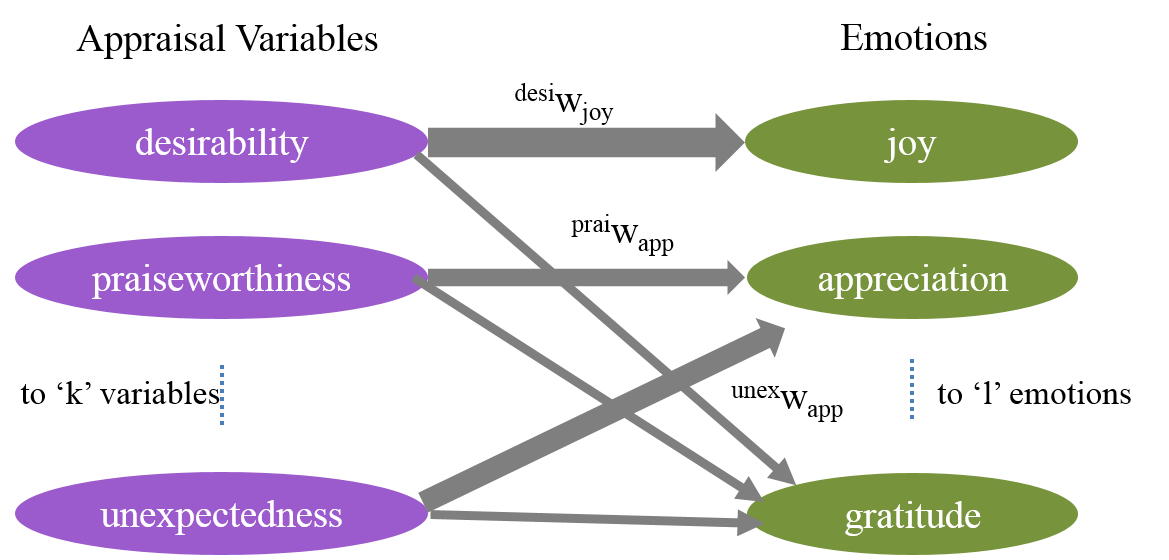}
\caption{An weighted appraisal-emotion network showing many-to-many relationship between appraisal variables and emotions}
\label{fig:appraisal_emotion_network}
\end{figure}

Appraisal theories of emotion propose that appraisal variables and emotions maintain a many-to-many relationship. This means that an appraisal variable may affect more than one emotion and an emotion may be affected by more than one appraisal variable \cite{Ortony1990}. For example,  Figure~\ref{fig:appraisal_emotion_network} shows a generic network of appraisal variables linked to emotions. On the left of the network are appraisal variables. Please note that the network can have $k$ number of appraisal variables (depending on the theory used) although only three appraisal variables are currently shown in Figure~\ref{fig:appraisal_emotion_network}. This choice has been made for the simplicity of explanation. Likewise, on the right of the network are emotions of the artificial agent. There can be $l$ number of emotions based on which emotion theory is implemented in the agent. As previously mentioned, each appraisal variable may influence one or more emotions. Also, an association between an appraisal variable and an emotion is weighted showing the degree by which the appraisal variables affect on the intensity of the emotion. For example, in Figure~\ref{fig:appraisal_emotion_network}, the emotion \emph{joy} is determined by the appraisal variable \emph{desirability} and the weight of association between them is denoted by $\leftidx{^{desi}}{w}{_{joy}}$. This suggests that a desirable event tends to cause the emotion of joy in the agent. Unlike the emotion of \emph{joy}, the emotion of \emph{appreciation} is affected by two appraisal variables. Appraisal variables \emph{praiseworthiness} and \emph{unexpectedness} affect the intensity of emotion \emph{appreciation}. This means if the action of a human actor is praiseworthy from the perspective of the artificial agent and if the action was not expected, this combination may lead the agent to reach to an emotional state of appreciation. The weights of association of appraisal variable \emph{praiseworthiness} and \emph{unexpectedness} with emotion \emph{appreciation} are denoted by $\leftidx{^{prai}}{w}{_{app}}$ and $\leftidx{^{unex}}{w}{_{app}}$ respectively. 

Complex emotions like \emph{gratitude} offers the option of two different kinds of treatment. As explained by most appraisal theories, the emotion of \emph{gratitude} can be considered a combination of the emotions \emph{joy} and \emph{appreciation}. In other words, \emph{gratitude} is a state of emotion in which an agent appraises an event to be desirable to achieve its goals and the action of the human counterpart involved in the event was praiseworthy as per the standards of the agent and this action was quite unexpected for the agent. Therefore, the intensities of complex emotions (including \emph{gratitude}) can be determined by two different ways: (i) \emph{by combining the intensities of simpler constituent emotions} or (ii) \emph{by considering the association with the appraisal variables linked to the constituent emotions}. For example, intensity of the emotion \emph{gratitude} can be determined by combining the intensities of the emotions \emph{joy} and \emph{appreciation}. Alternatively, the same thing can be achieved by considering that the emotion \emph{gratitude} is affected by appraisal variables \emph{desirability}, \emph{praiseworthiness} and \emph{unexpectedness} (as denoted by the arrows from these variables to the emotion \emph{gratitude} in Figure~\ref{fig:appraisal_emotion_network}). 

However, it should be noted that none of the existing appraisal theories present a clear explanation of how the intensities of simpler emotions can be combined to produce the intensity of more complex emotions, with the exception of only a few computational accounts \cite{Reilly2006}. A very simple approach could be to perform an average of the intensities of constituent emotions. But, we anticipate that this approach may not provide an accurate simulation of emotion generation mechanism in artificial agents, in line with the views of \citet{Hudlicka2008}. Whether the contribution of the constituent emotion intensities to complex emotions coincides (or closely resembles) with the intensity given by considering the individual appraisal variables does not have a universal consensus. This shall involve a separate research work the discussion of which lies outside the scope of this paper. In this paper, we are mainly concerned about presenting a \emph{theory-independent} approach to generate intensities of emotions from appraisal variables in an artificial agent. By saying `theory-independent', we mean that our approach can be applicable in an artificial agent implementing any appraisal theory of emotion. One should also note that the appraisal-emotion network presented in Figure~\ref{fig:appraisal_emotion_network} is an example only. The approach we are proposing in this paper can be used for any appraisal theory. Following the mainstream belief on the existing appraisal theories, we assumed that there exists a weight for each association of an appraisal variable to an emotion (as denoted by solid arrows in Figure~\ref{fig:appraisal_emotion_network}). But, how do we operationalise such weights in artificial agents? Although the emotion theories suggest that the process of mapping the appraisal variables to emotion intensities is modulated by factors like \emph{personality} and \emph{mood} \cite{Ortony1990}, only descriptive accounts are available, which cannot be readily implemented in computational models. We shall further discuss how the factors of personality and mood can help in establishing a dynamic relationship of appraisal variables with emotion intensities, in the remaining of the paper.

\subsection{Dynamic Learning of Appraisal--Emotion Association}
\label{sec:system>dynamic_learning}

In the previous sections, we discussed that (i) \emph{an artificial agent implementing appraisal theory for emotion generation needs to have a mechanism to map the computed appraisal variables into various emotion intensities}; (ii) \emph{each appraisal variable may be associated with one or more emotions} and (iii) \emph{each emotion may be influenced by one or more appraisal variables}, and (iv) \emph{these associations are weighted indicating the degree by which the given appraisal variable affects the given emotion}. Additionally, based on the existing literature, we also stated that the weight of association of appraisal variables to emotions is modulated by the factors of personality and mood \cite{Morris1992,Neumann2001,Corr2008}. In this section, we expand the appraisal--emotion association weight ($\leftidx{^{var}}{w}{_{emo}}$) into a simple linear form as shown in \eqref{eqn:weight}.

\begin{equation}
\label{eqn:weight}
\begin{split}
	\leftidx{^{var}}{w}{_{emo}} & = f_O*O + f_C* C + f_E*E + f_A*A + f_N*N + f_M*M
\end{split}
\end{equation}

\noindent
Where, $\leftidx{^{var}}{w}{_{emo}}$ denotes the final weight of association between an appraisal variable and an emotion. $O$ denotes the personality factor of \emph{openness} and $f_O$ denotes the weighting for the factor $O$. $C$ denotes the personality factor of \emph{conscientiousness} and $f_C$ denotes the weighting for the factor $C$. $E$ denotes the personality factor of \emph{extroversion} and $f_E$ denotes the weighting for the factor $E$. $A$ denotes the personality factor of \emph{agreeableness} and $f_A$ denotes the weighting for the factor $A$. $N$ denotes the personality factor of \emph{neuroticism} and $f_N$ denotes the weighting for the factor $N$. $M$ denotes the \emph{mood} factor and $f_M$ denotes the weighting for the factor $M$.

In \Eq~\eqref{eqn:weight}, the weighing factors ($f_O$, $f_C$, $f_E$, $f_A$, $f_N$ and $f_M$) can have positive as well as negative value. This is because a personality factor may have positive, negative or no influence in the relationship between an appraisal variable and an emotion. For example, we can anticipate that the personality of \emph{extroversion} has positive effect on the emotion \emph{joy} and the personality of \emph{neurotocism} may have negative effect on \emph{joy}. Hence, for the link $\leftidx{^{des}}{w}{_{joy}}$ (in Figure~\ref{fig:appraisal_emotion_network}), the value of $f_E$ should be positive while the value of $f_N$ should be negative. Although, we can make use of commonsense to estimate the plausible range for the  values for weighing factors, we can not specify a singular value with certainty. This is the reason we employed a machine learning technique to determine these weighing factors for each appraisal-emotion relationship that can be applied to map any number of appraisal variables to any number of emotions depending on the appraisal theory used to simulate the emotion mechanism in artificial agent. The following section presents the learning algorithm used to establish a dynamic relationship among appraisal variables and emotions.

\begin{figure}[!t]
\centering
\includegraphics[width=0.65\textwidth]{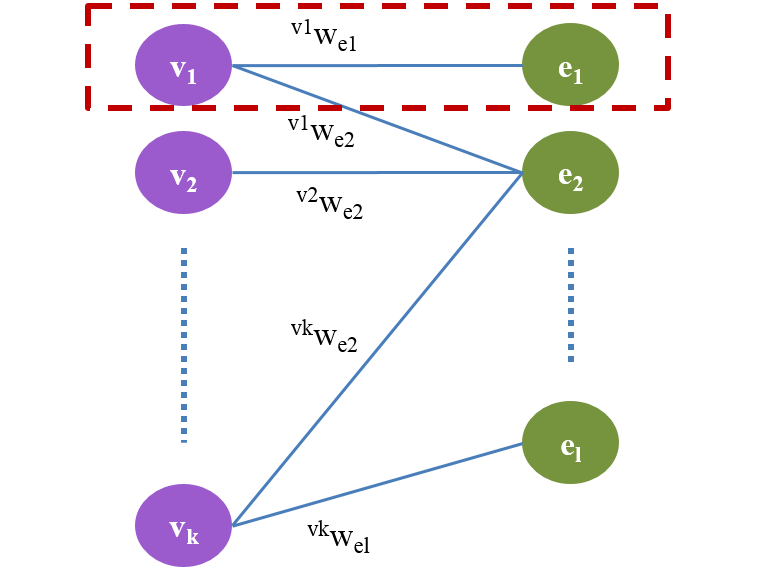}
\caption{A general appraisal-emotion network with $k$ appraisal variables and $l$ emotion types.}
\label{fig:general_appraisal_emotion_network}
\end{figure}

\begin{figure}[!t]
\centering
\includegraphics[width=0.65\textwidth]{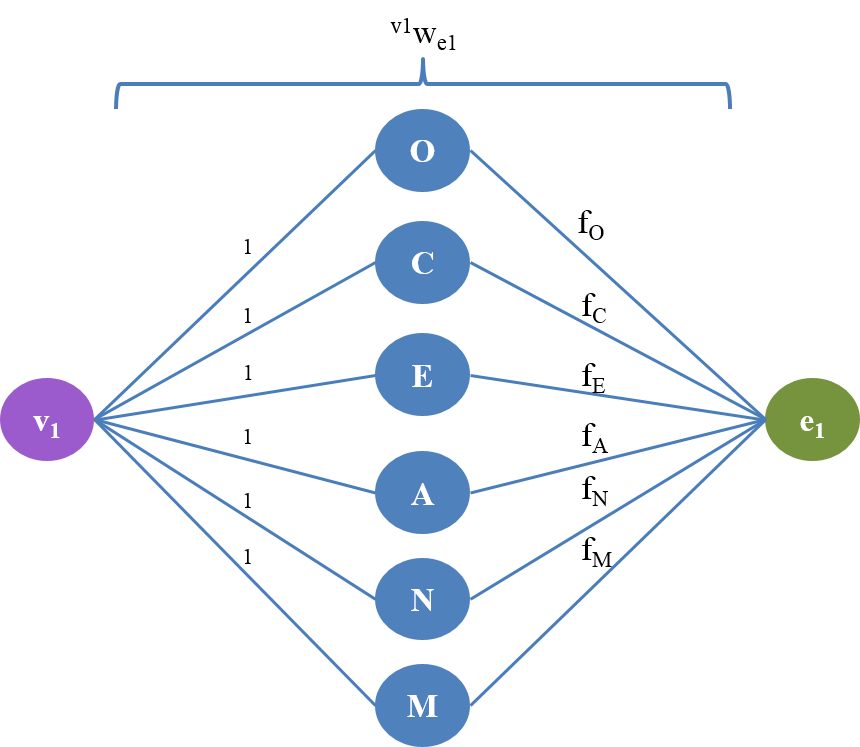}
\caption{Decomposition of the link between appraisal variable $v_1$ and emotion type $e_1$.}
\label{fig:decomposed_network}
\end{figure}

\subsubsection{Learning Algorithm}
\label{sec:system>learning_algorithm}

Considering we have $k$ appraisal variables and $l$ emotion types, the weighted appraisal-emotion network in Figure~\ref{fig:appraisal_emotion_network} can be represented in more generalised form as in Figure~\ref{fig:general_appraisal_emotion_network}. $\leftidx{^{v_1}}{w}{_{e_1}}$ denotes the weight of association between the appraisal variable $v_1$ and the emotion type $e_1$ and so on. As previously mentioned, this weight is affected by the personality factors and mood of the appraising individual. In order to better understand this phenomenon, we can break down the link between $v_1$ and $e_1$ (denoted by the dashed rectangle in Figure~\ref{fig:general_appraisal_emotion_network}), and represent as shown in Figure~\ref{fig:decomposed_network}.

The contribution of an appraisal variable to an emotion intensity is given by the product of the value of the appraisal variable and the weight of association of the appraisal variable to the emotion type. For example, the contribution of the appraisal variable $v_1$ to the emotion type $e_1$ (see Figure~\ref{fig:general_appraisal_emotion_network}) can be defined as the value $v_1$ * $\leftidx{^{v_1}}{w}{_{e_1}}$. This implies the following.

\begin{equation}
\label{eqn:breakdown}
\begin{split}
\hat{e}_{1_{v_1}} & = v_1 * \leftidx{^{v_1}}{w}{_{e_1}} \\
          & = v_1 * (f_O*O + f_C* C + f_E*E + f_A*A + f_N*N + f_M*M)\\
	      & = v_1Of_O + v_1Cf_C + v_1Ef_E + v_1Af_A + v_1Nf_N + v_1Mf_M\\
	      & = p_Of_O + p_Cf_C + p_Ef_E + p_Af_A + p_Nf_N + p_Mf_M
\end{split}
\end{equation}

\noindent Where $p_O$ denotes the product of $v_1$ and the personality factor $O$ i.e. $v_1*O$; $p_C$ denotes $v_1$ * $C$; $p_E$ denotes $v_1$ * $E$; $p_A$ denotes $v_1$ * $A$; $p_N$ denotes $v_1$ * $N$; $p_M$ denotes $v_1$ * $M$.

Alternatively, the contribution of the appraisal variable $v_1$ to the emotion type $e_1$ i.e. $\hat{e}_{1_{v_1}}$ can be denoted as follows.

\begin{equation}
\label{eqn:formalisation}
\begin{split}
\hat{e}_{1_{v_1}} &=  \sum\limits_{x} p_x * f_x \\
            & \mbox{where,}\ x \in \{O, C, E, A, N, M\}\\
            & \ \ \ \ \ \ \ \ \ \ \ \  p_x \in \{p_O, p_C, p_E, p_A, p_N, p_M\}\\
            & \ \ \ \ \ \ \ \ \ \ \ \  f_x \in \{f_O, f_C, f_E, f_A, f_N, f_M\}\\
\end{split}
\end{equation}

The formulation in \Eq~\eqref{eqn:formalisation} considers $p_x$ as the input value and $f_x$ as the weight to be learned for the corresponding $p_x$. If the expected value for the contribution of the appraisal variable $v_1$ to the emotion $e_1$ is denoted by $e_{1_{v_1}}$ and estimated value is denoted by $\hat{e}_{1_{v_1}}$, then the learning mechanism for each $f$ is given by the equation below.

\begin{equation}
\label{eqn:learning_algorithm}
\begin{split}
f'_x &= f_x + \eta (e_{1_{v_1}} - \hat{e}_{1_{v_1}}) * p_x\\
\end{split}
\end{equation}

In \Eq~\eqref{eqn:learning_algorithm}, $\eta$ is decaying learning rate.

\subsubsection{Formalisation of the Learning Algorithm}
\label{sec:system>formalisation_of_the_learning_algorithm}

In the previous section, we presented a demonstration of how a link between an appraisal variable and an emotion can be formulated to determine the weights associated with personality and mood factors for that link. In this section, we shall present a formal model of the algorithm used.

Most emotions $\mathbf{E}=\{e_{1},e_{2},..,e_{l}\}$ could be predicted directly from the appraisal variables $\mathbf{V}=\{v_{1},v_{2},...,v_{k}\}$ using a linear model $\mathbf{W}$ having less than or equal to $k \times l$ links (weights) such that $\hat{e}_{l}=\sum_{k}w_{lk}v_{k}$. However, as previously mentioned, it is believed that the weights $W$ are modulated by factors $F$ of personality and mood $\hat{M}=\{m_O,m_C,m_E,m_A,m_N,m_M\}$ so that

\begin{equation}
w_{lk}=\sum_{x}(f_{lk})_x m_{x}:x\in\{O,C,E,A,N,M\}
\end{equation}

\noindent So, given the data set $D=\{d_{1},d_{2},..,d_{l}\}$ of samples $d$, where each sample $d= \{E,V,\hat{M}\}$ is a set of an emotional state $E$, an appraisal $V$ and personality/mood factors $\hat{M}$, we can define the model to predict $E$ from $V$ and $\hat{M}$ as follows:

\begin{equation}
\hat{e}_{l}=\sum_{k}\left( \sum_{x}(f_{lk})_xm_{x}\right) v_{k}
\end{equation}

\noindent We can learn the parameters of this linear model using stochastic gradient descent \cite{Bottou2010} by minimising the squared error $L=\sum_{l}\left(e_{l}-\hat{e}_{l}\right)^{2}$
summed over the dataset $D$.

\noindent Since:

\begin{equation}
\frac{dL}{df_{lkx}}=2(e_{l}-\hat{e}_{l})\frac{d\hat{e}_{l}}{df_{lkx}}=2(e_{l}-\hat{e}_{l})m_{x}v_{k}
\end{equation}

\noindent we can minimise $\sum_{d\in D}L_{d}$ by iteratively performing the update
using individual lines of data $d$:

\begin{equation}
f_{lkx}\leftarrow f_{lkx}+\eta(e_{l}-\hat{e}_{l})m_{x}v_{k}
\end{equation}
where $\eta$ is a decaying learning rate.

\vspace{0.5cm}
The learning mechanism presented in the previous sections helps in determining (i) \emph{how each of the personality and mood factors quantitatively affect the process of emotion generation} (personality factors are described in Five Factor model \cite{Digman1990}). Running the above mentioned algorithm allows EEGS to (ii) \emph{ dynamically determine the weights of association for each of the factors in the link between an appraisal variable and an emotion}. Combining the sum of the product of these weights and corresponding factors gives the overall weight of association ($\leftidx{^{var}}{w}{_{emo}}$) between an appraisal variable ($var$) and an emotion ($emo$) as presented in \Eq~\eqref{eqn:weight}. This weight is then multiplied by the quantitative value of the appraisal variable in order to determine the intensity contribution of the appraisal variable to the emotion as in \Eq~\eqref{eqn:breakdown} or \eqref{eqn:formalisation}. The final intensity of an emotion is determined by summing the contribution from all the appraisal variables associated, which shall be detailed in Section~\ref{sec:system>computation_of_emotion_and_mood_intensities}. 

\subsection{Implementation of Affect Generation Process in EEGS}
\label{sec:system>computation_of_emotion_and_mood_intensities}

In Sections\ref{sec:system>appraisal-emotion_network} and \ref{sec:system>dynamic_learning}, we (i) presented how different appraisal variables can be linked with various emotions with different degrees \ie weights, and also (ii) explained the machine learning approach used that allows EEGS to learn the parameters of this proposed model, namely the weights between appraisals and emotions. Most importantly, the discussed approach is general enough to consider any combination of appraisal variables and emotional states the system has to consider. As such, the only requirement is an appropriate dataset $D$ including enough samples for such appraisals and emotions. In the following paragraphs, we will describe how emotions are modelled in EEGS.

\subsubsection{Emotion Types in EEGS}
\label{sec:system>emotion_types_in_EEGS}

EEGS is currently able to generate and express ten emotions described in OCC theory \cite{Ortony1990} which are listed below. 

\begin{itemize}
\itemsep0em 
\item \emph{Joy} : A feeling of pleasure or happiness.
\item \emph{Distress} : A feeling of anxiety, sorrow, or pain.
\item \emph{Happy$\_$For} : A feeling of happiness for someone's desirable situation.
\item \emph{Sorry$\_$For} : A feeling of sadness for someone's undesirable situation.
\item \emph{Appreciation} : A feeling when one recognises the good qualities or actions of someone.
\item \emph{Reproach} : To express to (someone) one's disapproval of or disappointment in their actions.
\item \emph{Gratitude} : The state of being grateful to someone.
\item \emph{Anger} : A strong feeling of annoyance, displeasure, or hostility.
\item \emph{Liking} : A feeling when you see someone appealing or interesting.
\item \emph{Disliking} : A feeling when you see someone unappealing or uninteresting.
\end{itemize}

\noindent To effectively process emotions in a computational model like EEGS, an appropriate structure representing various aspects of an emotion is necessary. In this paragraph, we will provide the structure used in EEGS to represent emotional states. According to literature, an emotion can be categorised with a name for its type \cite{Ortony1990} and, therefore, each emotion is addressed by a specific word in a language. This label is used as a proxy to refer to the feeling the person experiences when under the influence of such emotional state. For example, the emotional label \emph{joy}, in the above list, is used to refer to a feeling of internal pleasure. Since our computational model has been heavily inspired by OCC theory \cite{Ortony1990}, our representation considers the assumption of the theory that \emph{emotions are valenced reactions to situations}. Hence, we assume that the emotional state can be perceived as leading to either a negative or positive experience for the subject, \ie the experience underlying the considered emotion is valenced. For example, the emotion \emph{gratitude} is positively valenced, because it leads to a pleasant experience for the subject, whereas the emotion \emph{anger} is negatively valenced. Importantly, to effectively describe and differentiate emotional states, it is not enough to assign them a discrete state of valence either being negative or positive. Indeed, the valenced reaction of the emotion can be situated within different degrees of a valence scale. For example, the emotion \emph{anger} has higher degree of negativity compared to the emotion \emph{reproach}. Detailed discussion about how emotions are differentiated with varying values for the degree of their positivity or negativity will be presented in the following section. In addition to type (name), valence and degree, emotion theories believe that there is a threshold associated with each emotion which represents the minimum intensity required for that emotion to be active, or in other words, to reach the subject's awareness \cite{Ortony1990, Scherer2001}. However, what should be the threshold of a particular emotion from computational perspective is still an unanswered question. Summarising these sentences, an emotion can be described as a valenced reaction to the situation. This reaction can have either negative or positive valence and emotions belonging to the same valenced class can have different degrees of valence within that class. In addition, the intensity of the emotional reaction represents how strongly the valenced reaction is perceived by the subject. As such, in this research, we have assumed that an emotion always has either a positive or negative valence. However, since the degree of valence of such emotional states can be very mild (\ie close to zero), that emotional state may be experienced as neither positive or negative, though still being mildly positive or negative valenced. In addition to these aspects, commonly in emotion modelling literature, the notion of decay time is also evident \cite{Marreiros2010, Padgham1996}. Decay time denotes the time needed for a particular emotion to reach to the level of 0 (zero) intensity.

Based on the existing literature, we have considered the aspects that are essential to define a data structure of emotion and represented an emotion in EEGS in the form of:

\vskip 0.1in

\centerline{{\tt (Name, Valence, Degree, Threshold, Intensity, DecayTime)}}

\vskip 0.1in

\noindent where, {\tt Name} denotes the name for the type of the emotion, {\tt Valence} specifies whether the emotion is positive or negative, {\tt Degree} represents the extent of the positivity and negativity of the emotion, {\tt Threshold} represents the minimum intensity required to trigger the emotion, {\tt Intensity} represents the strength of the emotional experience and {\tt DecayTime} denotes the time required to drop the emotion intensity back to 0. For example, the emotion structure {\tt (\emph{distress}, NEGATIVE, -0.8090, 0.0, 0.5, 10)} denotes the emotion of {\tt Name} {\tt \emph{distress}} which has {\tt NEGATIVE Valence} with {\tt Degree} of {\tt -0.8090}, {\tt Threshold} of {\tt 0.0}, {\tt Intensity} of {\tt 0.5}, and {\tt DecayTime} of {\tt 10} seconds. In EEGS, {\tt Valence} is either ``POSITIVE'' or ``NEGATIVE''; {\tt Degree}\footnote{While the signed value of {\tt Degree} was sufficient to specify the {\tt Valence} as POSITIVE or NEGATIVE, we chose to consider {\tt Valence} as an explicit parameter for the ease of computational mechanism in some conditional checks.} is a number in the range [-1, +1], where -1 denotes extremely negative emotion and +1 denotes extremely positive emotion; {\tt Threshold} is a number in the range [0, 1); {\tt Intensity} is a number in the range [0, 1] and {\tt DecayTime} is a number which currently has been considered between 0 and 10 seconds. We could not find strong evidence on how long the decay time should be considered for an emotion. However, some emotion models were found to use the decay time of less than 10 seconds \cite{Becker2008}. It should be noted that the quantities {\tt Name}, {\tt Valence}, {\tt Degree}, {\tt Threshold}, and {\tt DecayTime} in the above defined emotion structure are constants because a particular type of emotion has a fixed name, fixed valence and degree as well as fixed threshold and decay time since these are inherent properties of an emotion. The only thing that changes in the course of interaction is the {\tt Intensity} of a particular emotion. We can use commonsense to determine the {\tt Valence} of an emotion. For example, it is obvious that an emotion of \emph{joy} has POSITIVE {\tt Valence} and an emotion of \emph{distress} has NEGATIVE {\tt Valence}. However, we cannot say with certainty what would be the {\tt Degree} of the {\tt Valence} of the given emotion \ie how much positive or negative is the experience of the emotion quantitatively. In order to provide a viable solution to this issue, in Section~\ref{sec:system>derivation_of_emotion_valence_dagree}, we propose a mechanism to determine the plausible values for the valence of degrees of each emotion based on the work of \citet{Remington2000}.

\begin{figure}[!t]
\centering
\includegraphics[width=0.7\textwidth]{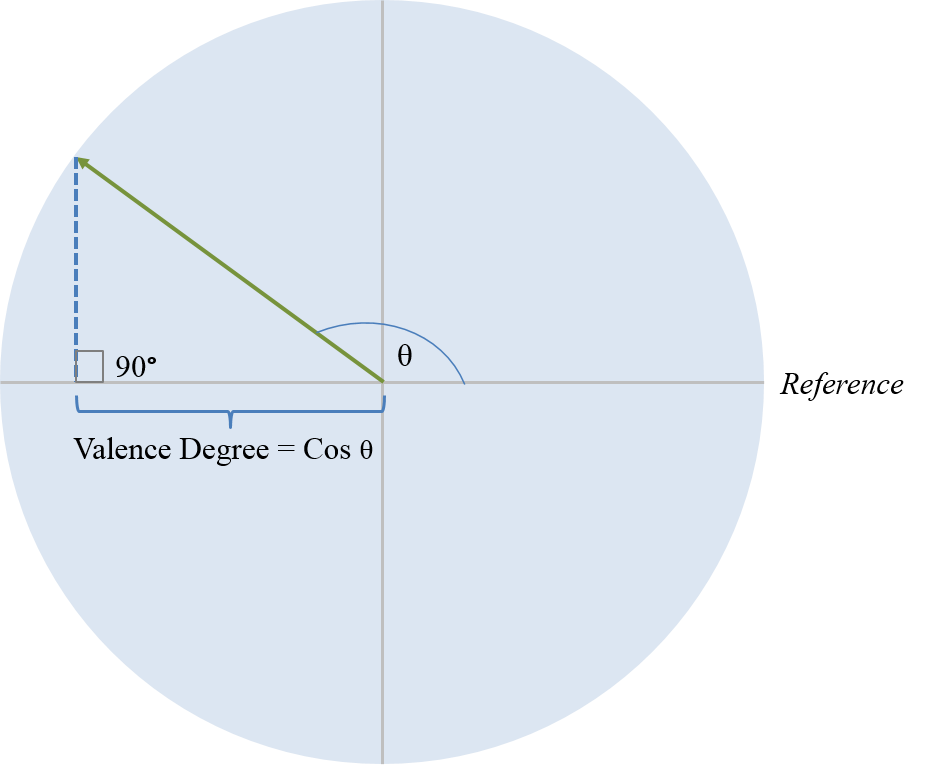}
\caption[Mechanism for mapping the angle of an emotion type into a signed valence degree]{Mechanism for mapping the angle of an emotion type into a signed valence degree. $\theta$ is the average angle for an emotion type derived from \citet{Remington2000}. Horizontal axis in the upper right quadrant is considered as reference for the measurement of angle.}
\label{fig:emotion_valence_derivation}
\end{figure}

\subsubsection{Derivation of Emotion Valence Degree}
\label{sec:system>derivation_of_emotion_valence_dagree}

\citet{Remington2000} present a work that performs a deep reanalysis of the circumplex model of affect and summarises the locations of various affective states in the circumplex based on 10 correlation matrices -- 3 matrices from \citet{Feldman1995}, 1 matrix from \citet{Barrett1996}, 2 matrices from \citet{Mayer1988}, 3 matrices from \citet{Mayer1988b}, and 1 matrix from \citet{Rusting1995}. Each of these matrices provided an angle for the location of an affective state from the $0^\circ$. Since not each matrix provided the location of all the affective states (\ie emotions in the context of this research) considered in EEGS, for some emotions the angles available were less than 10 in number. In order to obtain a more precise angle (location) of an emotion, we calculated the average of the angles provided by various correlation matrices. Since, \citet{Remington2000} only provided the list of angles offered in various correlation matrices, we calculated the corresponding {\tt Valence Degree} using a geometric mapping of the angle to its projection on the horizontal axis as shown in Figure~\ref{fig:emotion_valence_derivation}. It should be noted that the emotion \emph{joy} is considered as the baseline in the correlation matrices provided by \citet{Remington2000}. Therefore, the location angle for \emph{joy} is defined to be $0^\circ$ which leads its {\tt Valence Degree} to be a positive value of 1.0 (as shown in Table~\ref{tbl:emotion_valence_derivation}). 

\begin{table}
\caption{Mapping of the angles of the circumplex into valence degree for various emotions.}
\label{tbl:emotion_valence_derivation}       
\begin{tabular}{p{0.35\textwidth}p{0.25\textwidth}p{0.25\textwidth}}
\hline\noalign{\smallskip}
    Emotion & Angle & Valence Degree\\
\noalign{\smallskip}\hline\noalign{\smallskip}
    \emph{joy} & $0^\circ$ & 1.0\\
    \emph{distress} & $144^\circ$ & -0.8090\\
    \emph{happy$\_$for} & $58^\circ$ & 0.5299\\
    \emph{sorry$\_$for} & $122^\circ$ & -0.5299\\
    \emph{appreciation} & $26^\circ$ & 0.8988\\
    \emph{reproach} & $153.33^\circ$ & -0.8936\\
    \emph{gratitude} & $8^\circ$ & 0.9903\\
    \emph{anger} & $164.75^\circ$ & -0.9648\\
    \emph{liking} & $14.5^\circ$ & 0.9681\\
    \emph{disliking} & $165.5^\circ$ & -0.9681\\
\noalign{\smallskip}\hline
\end{tabular}
\end{table}

In Table~\ref{tbl:emotion_valence_derivation}, the valence degree of the emotion \emph{anger} is determined by first calculating the angle of projection from baseline. For this, four correlation matrices were considered -- Matrix 2 from \cite{Mayer1988} and Matrix 1, 2 and 3 from \cite{Mayer1988b}. The angles considered were 166$^\circ$, 171$^\circ$, 158$^\circ$ and 164$^\circ$ respectively which gave an average angle of 164.75$^\circ$. The angle provided by \citet{Rusting1995} was excluded because it was an outlier \ie only 83$^\circ$. Applying cosine to the obtained angle gave a valence degree of -0.9648 for the emotion of \emph{anger}.

\begin{table}
\caption{Association of various appraisal variables with different emotions as suggested in the OCC theory \cite{Ortony1990}.}
\label{tbl:appraisal_emotion_association}       
\begin{tabular}{p{0.5\textwidth}p{0.4\textwidth}}
\hline\noalign{\smallskip}
    Appraisal Variable & Associated Emotion\\
\noalign{\smallskip}\hline\noalign{\smallskip}
    \multirow{4}{0.5\textwidth}{\emph{desirability}} & \emph{joy} \\
    & \emph{distress}\\
    & \emph{happy$\_$for}\\
    & \emph{sorry$\_$for}\\
    & \emph{gratitude}\\
    & \emph{anger}\\
    
    \hline
    
    \multirow{4}{0.5\textwidth}{\emph{praiseworthiness}} & \emph{appreciation} \\
    & \emph{reproach}\\
    & \emph{gratitude}\\
    & \emph{anger}\\
    
    \hline
    
    \multirow{2}{0.5\textwidth}{\emph{appealingness}} & \emph{liking} \\
    & \emph{disliking}\\
    
    \hline
    
    \multirow{2}{0.5\textwidth}{\emph{deservingness}} & \emph{happy$\_$for} \\
    & \emph{sorry$\_$for}\\
    
    \hline
    
    \multirow{2}{0.5\textwidth}{\emph{familiarity}} & \emph{liking} \\
    & \emph{disliking}\\
    
    \hline

    \multirow{4}{0.5\textwidth}{\emph{unexpectedness}} & \emph{appreciation} \\
    & \emph{reproach}\\
    & \emph{gratitude}\\
    & \emph{anger}\\
\noalign{\smallskip}\hline
\end{tabular}
\end{table}

\subsubsection{From Appraisal to Emotion Intensities}
\label{sec:system>from_appraisal_to_emotion_intensities}

In the previous sections, we discussed how emotions are structured in EEGS and how the {\tt Valence} and {\tt Degree} of various emotion types are determined. In this section, we will present the details of how the values of appraisal variables are mapped into different emotion intensities based on the weights of association identified in Sections~\ref{sec:system>appraisal-emotion_network} and \ref{sec:system>dynamic_learning}. As previously discussed, each appraisal variable may affect the intensity of more than one emotions and each emotion may be affected by more than one appraisal variables. Table~\ref{tbl:appraisal_emotion_association} shows how each appraisal variable in EEGS is linked to different emotions. This association has been defined as per the suggestion of \citet{Ortony1990}. It should be noted that the weight of association between the appraisal variable and an emotion lies in the range of [-1, 1], where a positive value of the weight indicates that the variable affects the emotion intensity positively and a negative value of weight indicates that the variable affects the emotion intensity negatively. For example, the appraisal variable \emph{desirability} which measures how desirable an event is in relation to its goals, should have positive weight of association with emotion \emph{joy} and a negative weight of association with the emotion \emph{distress}. As such, if the value of \emph{desirability} is positive meaning the event is desirable, this appraisal will increase the intensity of \emph{joy} emotion while decreasing the intensity of \emph{distress} emotion at the same time.

The weights associated with each appraisal--emotion pair contributes in determining the degree by which the appraisal variable affects the intensity of the emotion \cite{Ortony1990}. This implies that the `effect' of an appraisal variable on an emotion is the function ($\mathcal{I}$) of the quantitative value of the appraisal variable ($v_i$) and weight of association of the appraisal variable ($\leftidx{^{v_i}}{w}{_{e_j}}$) with the emotion ($e_j$). By saying `effect' we refer to the contribution an appraisal variable makes to the intensity of a particular emotion because its intensity may also be affected by the values of other appraisal variables, as represented by the equation below.

\begin{equation}
\label{eqn:appraisal_intensity_contribution}
\begin{split}
	\hat i_{e_{j_i}} & =  \mathcal{I}_{e}(v_i, \leftidx{^{v_i}}{w}{_{e_j}}): \ i \in[1, k] \mbox{ \& } j \in [1, l] \\
	& = v_i * \leftidx{^{v_i}}w_{e_j}
\end{split}
\end{equation}

\noindent where, $\hat i_{e_{j_i}}$ denotes the contribution of the $i^{th}$ appraisal variable to the intensity of $j^{th}$ emotion. If there are $n$ appraisal variables related to an emotion, then, the final intensity of each emotion ($\hat i_{e_j}$) is determined by the cumulative effect of all the appraisal variables linked to the emotion, as shown in Figure~\ref{fig:appraisal_emotion_network}. This phenomenon is represented by \eqref{eqn:cumulative_intensity}.

\begin{equation}
\label{eqn:cumulative_intensity}
	\hat i_{e_j} = \sum_{i=1}^{n}{ \hat i_{e_{j_i}}}, \forall \ j \in[1, l]
\end{equation}

These computations are performed for all the emotions, it results in a set of emotions $\mathbf{E}=\{e_1, \dots, e_l\}$ with respective intensities $\mathbf{I}=\{\hat i_{e_1}, \dots, \hat i_{e_l}\}$. Hence, the appraisal--emotion network presented in Figure~\ref{fig:appraisal_emotion_network} helps in the computation of the intensities of various emotions of the model based on the \Eqs~\ref{eqn:appraisal_intensity_contribution} and \ref{eqn:cumulative_intensity}. However, it should be noted that not all emotions exhibit a linear combination of product of appraisal values and corresponding weight of association of appraisal variable with the emotion. The computation of the final intensity of some of the emotions like \emph{appreciation}, \emph{reproach}, \emph{gratitude}, \emph{anger}, \emph{liking} and \emph{disliking} may follow non-linear combination as will be discussed in the following sub-sections.

\vspace{0.5cm}
\noindent \textbf {\emph{joy}}\newline
\noindent As suggested by \citet{Ortony1990}, the emotion \emph{joy} is determined by only one appraisal variable -- \emph{desirability}. Therefore the intensity of \emph{joy} emotion is determined by the value of appraisal variable \emph{desirability} and the weight of association of the variable with \emph{joy} emotion. 

\begin{equation}
\label{eqn:joy_intensity}
	\hat i_{joy} = desi * ( \leftidx{^{desi}}w_{joy})
\end{equation}

\noindent \textbf {\emph{distress}}\newline
\noindent Like \emph{joy}, the emotion \emph{distress} is also affected by only one appraisal variable \emph{desirability} \cite{Ortony1990}. Therefore, the intensity of \emph{distress} emotion is also given by the formula similar to that of \emph{joy} emotion.

\begin{equation}
\label{eqn:distress_intensity}
	\hat i_{dist} = desi * ( \leftidx{^{desi}}w_{dist})
\end{equation}

\noindent \textbf {\emph{happy$\_$for}}\newline
\noindent The emotion \emph{happy$\_$for} denotes a feeling of happiness because something desirable happened to someone and the person deserved what happened. As such, the emotion \emph{happy$\_$for} is determined by the appraisal variables \emph{desirability} and \emph{deservingness} \cite{Ortony1990}.

\begin{equation}
\label{eqn:happy_for_intensity}
	\hat i_{hpy\_for} = desi * ( \leftidx{^{desi}}w_{hpy\_for}) + dese * ( \leftidx{^{dese}}w_{hpy\_for})
\end{equation}

\noindent \textbf {\emph{sorry$\_$for}}\newline
\noindent The emotion \emph{sorry$\_$for} denotes a feeling of sadness because something undesirable happened to someone and the person did not deserve what happened. As such, the emotion \emph{sorry$\_$for} is also determined by the appraisal variables \emph{desirability} and \emph{deservingness} \cite{Ortony1990}.

\begin{equation}
\label{eqn:sorry_for_intensity}
	\hat i_{sry\_for} = desi * ( \leftidx{^{desi}}w_{sry\_for}) + dese * ( \leftidx{^{dese}}w_{sry\_for})
\end{equation}

\noindent \textbf {\emph{appreciation}}\newline
\noindent Appreciation is the feeling one experiences when someone does a praiseworthy action from the viewpoint of standards of the assessing person and that action was not expected to happen \cite{Ortony1990}. As such, the emotion of \emph{appreciation} in EEGS, is determined by two appraisal variables -- \emph{praiseworthiness} and \emph{unexpectedness}. If an action of other agent is praiseworthy, then the appraising individual experiences some degree of appreciation. However, how much the action was unexpected largely affects the degree of praiseworthiness as well thereby defining the intensity of appreciation experienced. For this reason, we propose to combine the contributions of the appraisal variables \emph{praiseworthiness} and \emph{unexpectedness} in a non-linear fashion instead of linearly combining their individual contributions to the intensity of \emph{appreciation}. If we denote the contribution of the appraisal variable \emph{praiseworthiness} to emotion \emph{appreciation} as $\hat i_{{appr}_{prai}}$ = $prai * \leftidx{^{prai}}{w}_{appr}$ and the contribution of the appraisal variable \emph{unexpectedness} as $\hat i_{{appr}_{unex}}$ = $unex * \leftidx{^{unex}}{w}_{appr}$, then the overall intensity of the emotion \emph{appreciation} ($\hat i_{appr}$) is given by the following formula.

\begin{equation}
\label{eqn:appreciation_intensity}
	\hat i_{appr} = 
	\begin{cases}
		- {\left | \hat i_{{appr}_{prai}} \right|}^{1 -\hat i_{{appr}_{unex}}} & \mbox{if   } \hat i_{{appr}_{prai}} < 0\\
		{\left (\hat i_{{appr}_{prai}}\right )}^{1 -\hat i_{{appr}_{unex}}} & \mbox{if   } \hat i_{{appr}_{prai}} > 0\\
		0 & \mbox{otherwise   }
	\end{cases}
\end{equation}

\noindent \textbf {\emph{reproach}}\newline
\noindent The feeling of reproach arises when a person disapproves someone's blameworthy (not praiseworthy) action \cite{Ortony1990}. The degree of reproach is also affected by unexpectedness of the event in addition to the degree of blameworthiness. As such, the emotion \emph{reproach} in EEGS is determined by two appraisal variables -- \emph{praiseworthiness} and \emph{unexpectedness} (as in the case of \emph{appreciation} emotion). If we denote the contribution of appraisal variable \emph{praiseworthiness} to emotion \emph{reproach} as $\hat i_{{repr}_{prai}}$ = $prai * \leftidx{^{prai}}{w}_{repr}$ and the contribution of the appraisal variable \emph{unexpectedness} as $\hat i_{{repr}_{unex}}$ = $unex * \leftidx{^{unex}}{w}_{repr}$, then the overall intensity of the emotion \emph{reproach} ($\hat i_{repr}$) is given by the following formula.

\begin{equation}
\label{eqn:reproach_intensity}
	\hat i_{repr} = 
	\begin{cases}
		- {\left | \hat i_{{repr}_{prai}} \right|}^{1 -\hat i_{{repr}_{unex}}} & \mbox{if   } \hat i_{{repr}_{prai}} < 0\\
		{\left (\hat i_{{repr}_{prai}}\right )}^{1 -\hat i_{{repr}_{unex}}} & \mbox{if   } \hat i_{{repr}_{prai}} > 0\\
		0 & \mbox{otherwise   }
	\end{cases}
\end{equation}

\noindent \textbf {\emph{gratitude}}\newline
\noindent Gratitude is a feeling one experiences in response to an unexpected praiseworthy action that is desirable for the achievement of ones goal(s) \cite{Ortony1990}. Therefore, the emotion \emph{gratitude} in EEGS is determined by the appraisal variables \emph{desirability}, \emph{praiseworthiness}, and \emph{unexpectedness}. If we denote the contribution of appraisal variable \emph{desirability} to emotion \emph{gratitude} as $\hat i_{{grat}_{desi}}$ = $desi * \leftidx{^{desi}}{w}_{grat}$, contribution of appraisal variable \emph{praiseworthiness} as $\hat i_{{grat}_{prai}}$ = $prai * \leftidx{^{prai}}{w}_{grat}$ and the contribution of the appraisal variable \emph{unexpectedness} as $\hat i_{{grat}_{unex}}$ = $unex * \leftidx{^{unex}}{w}_{grat}$, then the overall intensity of the emotion \emph{gratitude} ($\hat i_{grat}$) is given by the following formula.

\begin{equation}
\label{eqn:gratitude_intensity}
	\hat i_{grat} = \hat i_{{grat}_{desi}} +
	\begin{cases}
		- {\left | \hat i_{{grat}_{prai}} \right|}^{1 -\hat i_{{grat}_{unex}}} & \mbox{if   } \hat i_{{grat}_{prai}} < 0\\
		{\left (\hat i_{{grat}_{prai}}\right )}^{1 -\hat i_{{grat}_{unex}}} & \mbox{if   } \hat i_{{grat}_{prai}} > 0\\
		0 & \mbox{otherwise   }
	\end{cases}
\end{equation}

\noindent \textbf {\emph{anger}}\newline
\noindent Anger is a feeling one experiences in response to an unexpected blameworthy action that is undesirable for the achievement of ones goal(s) \cite{Ortony1990}. Therefore, the emotion \emph{anger} in EEGS is determined by the appraisal variables \emph{desirability}, \emph{praiseworthiness}, and \emph{unexpectedness}. If we denote the contribution of appraisal variable \emph{desirability} to emotion \emph{anger} as $\hat i_{{angr}_{desi}}$ = $desi * \leftidx{^{desi}}{w}_{angr}$, contribution of appraisal variable \emph{praiseworthiness} as $\hat i_{{angr}_{prai}}$ = $prai * \leftidx{^{prai}}{w}_{angr}$ and the contribution of the appraisal variable \emph{unexpectedness} as $\hat i_{{angr}_{unex}}$ = $unex * \leftidx{^{unex}}{w}_{angr}$, then the overall intensity of the emotion \emph{gratitude} ($\hat i_{angr}$) is given by the following formula.

\begin{equation}
\label{eqn:anger_intensity}
	\hat i_{angr} = \hat i_{{angr}_{desi}} +
	\begin{cases}
		- {\left | \hat i_{{angr}_{prai}} \right|}^{1 -\hat i_{{angr}_{unex}}} & \mbox{if   } \hat i_{{angr}_{prai}} < 0\\
		{\left (\hat i_{{angr}_{prai}}\right )}^{1 -\hat i_{{angr}_{unex}}} & \mbox{if   } \hat i_{{a}ngr_{prai}} > 0\\
		0 & \mbox{otherwise   }
	\end{cases}
\end{equation}

\noindent \textbf {\emph{liking}}\newline
\noindent Liking someone depends on how much an individual thinks the person is appealing \cite{Ortony1990}. This feeling is also affected by the degree of familiarity between the two individuals \cite{Ortony1990}. In line with this, in EEGS, the intensity of the emotion \emph{liking} is determined by the appraisal variables \emph{appealingness} and \emph{familiarity}. If we denote the contribution of appraisal variable \emph{appealingness} to emotion \emph{liking} as $\hat i_{{lkng}_{appl}}$ = $appl * \leftidx{^{appl}}{w}_{lkng}$ and the contribution of appraisal variable \emph{familiarity} as $\hat i_{{lkng}_{fami}}$ = $fami * \leftidx{^{fami}}{w}_{lkng}$, then the overall intensity of the emotion \emph{liking} ($\hat i_{lkng}$) is given by the following formula.

\begin{equation}
\label{eqn:liking_intensity}
	\hat i_{lkng} = 
	\begin{cases}
		- {\left | \hat i_{{lkng}_{appl}} \right|}^{\hat i_{{lkng}_{fami}}} & \mbox{if   } \hat i_{{lkng}_{appl}} < 0\\
		{\left (\hat i_{{lkng}_{appl}}\right )}^{\hat i_{{lkng}_{fami}}} & \mbox{if   } \hat i_{{lkng}_{appl}} > 0\\
		0 & \mbox{otherwise   }
	\end{cases}
\end{equation}

\noindent \textbf {\emph{disliking}}\newline
\noindent Similar to liking, disliking someone also depends on how much an individual thinks the person is appealing \cite{Ortony1990}. This feeling is also affected by the degree of familiarity between the two individuals \cite{Ortony1990}. As such, in EEGS, the intensity of the emotion \emph{disliking} is determined by the appraisal variables \emph{appealingness} and \emph{familiarity}. If we denote the contribution of appraisal variable \emph{appealingness} to emotion \emph{disliking} as $\hat i_{{dlkg}_{appl}}$ = $appl * \leftidx{^{appl}}{w}_{dlkg}$ and the contribution of appraisal variable \emph{familiarity} as $\hat i_{{dlkg}_{fami}}$ = $fami * \leftidx{^{fami}}{w}_{dlkg}$, then the overall intensity of the emotion \emph{disliking} ($\hat i_{dlkg}$) is given by the following formula.

\begin{equation}
\label{eqn:disliking_intensity}
	\hat i_{dlkg} = 
	\begin{cases}
		- {\left | \hat i_{{dlkg}_{appl}} \right|}^{\hat i_{{dlkg}_{fami}}} & \mbox{if   } \hat i_{{dlkg}_{appl}} < 0\\
		{\left (\hat i_{{dlkg}_{appl}}\right )}^{\hat i_{{dlkg}_{fami}}} & \mbox{if   } \hat i_{{dlkg}_{appl}} > 0\\
		0 & \mbox{otherwise   }
	\end{cases}
\end{equation}

\vspace{1.25cm}

It should be noted that although the formulas presented above are based on the theoretical suggestions of \citet{Ortony1990} some assumptions were made to define the proposed formulas. Eventhough some researchers have made some suggestions for the computation of very few emotions \cite{Gratch2004,El2000}, it is not possible to directly compare the accuracy of the currently proposed formulas with the previous proposals because the benchmark tests for previous models is not available. The formulas presented above exhibit a high level of accuracy in predicting emotion intensities collected from human participants (see for example \cite{Ojha2019AAMAS}), which sets a new benchmark for such evaluations for future researchers.

\subsubsection{Emotion Intensity Threshold}
\label{sec:system>emotion_intensity_threshold}

As discussed earlier in Section~\ref{sec:system>computation_of_emotion_and_mood_intensities}, an emotion can have a threshold that specifies a minimum level of activation required for an emotion to be actually present \cite{Ortony1990,Scherer2001}. OCC theory also introduces a concept called \emph{emotion--potential}. Emotion potential is the measure of the extent to which an event can trigger a particular emotion at the given time \cite{Ortony1990}. If emotion--potential of a particular emotion in reaction to an event is below the \emph{emotion--threshold} of that emotion, then the emotion is assumed not to become active at all \cite{Ortony1990}. As such, \cite{Ortony1990} suggest to compute \emph{emotion--potential} ($\hat i_e$) as a function of appraisal variables. This makes the emotion intensities computed in above sections as emotion--potential. OCC theory suggests to subtract the \emph{emotion--threshold} ($thres_e$) of that emotion from the computed emotion--potential in order to obtain the effective \emph{emotion--intensity} ($\hat{i}^{effective}_{e}$). As such the effective intensity for an emotion as suggested by \citet{Ortony1990} would be given as below.

\begin{equation}
\label{eqn:effective_intensity}
	\hat{i}^{effective}_{e} = \hat i_e - thres_e
\end{equation}

Interestingly, \citet{Ortony1990} consider that the value $thres_e$ is a function of time meaning the activation threshold of same emotion may vary with time. However, the authors do not specify how such a variation occurs. Also, the lack of well defined thresholds for various emotions in the first place makes it difficult to decide what should be the activation threshold of each emotion -- if any. Unlike some models that try to realise the notion of threshold for various emotions \cite{Dias2005,Dias2014,Becker2008,Velasquez1997}, we consider any emotion with emotion--potential above the value of zero to be active for the moment and consider the same value as the instantaneous intensity of that emotion. As such, in EEGS, the values of the appraisal variables directly determine the intensities of emotions depending on the weight of association by considering the activation threshold to be zero. However, we provide the flexibility of using non-zero intensity threshold in emotions in EEGS to allow future research and investigation by other researchers.

\subsubsection[]{Interaction among Emotion, Mood and Personality\footnote{Most of the content in this section is adapted from our published work \cite{Ojha2018ACS}}}
\label{sec:system>revisiting_interaction_among_emotion_mood_and_personality} 

\begin{figure}[!t]
\centering
\includegraphics[width=0.5\textwidth]{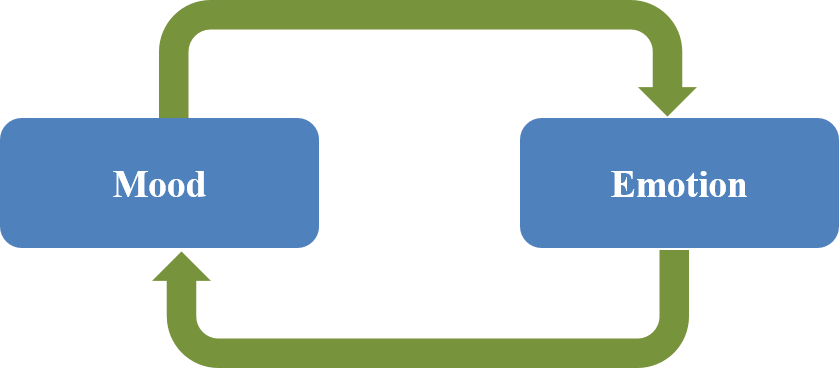}
\caption{Cyclic interaction between emotion and mood.}
\label{fig:cyclic_interaction}
\end{figure}

\begin{figure}[!t]
\centering
\includegraphics[width=0.6\textwidth]{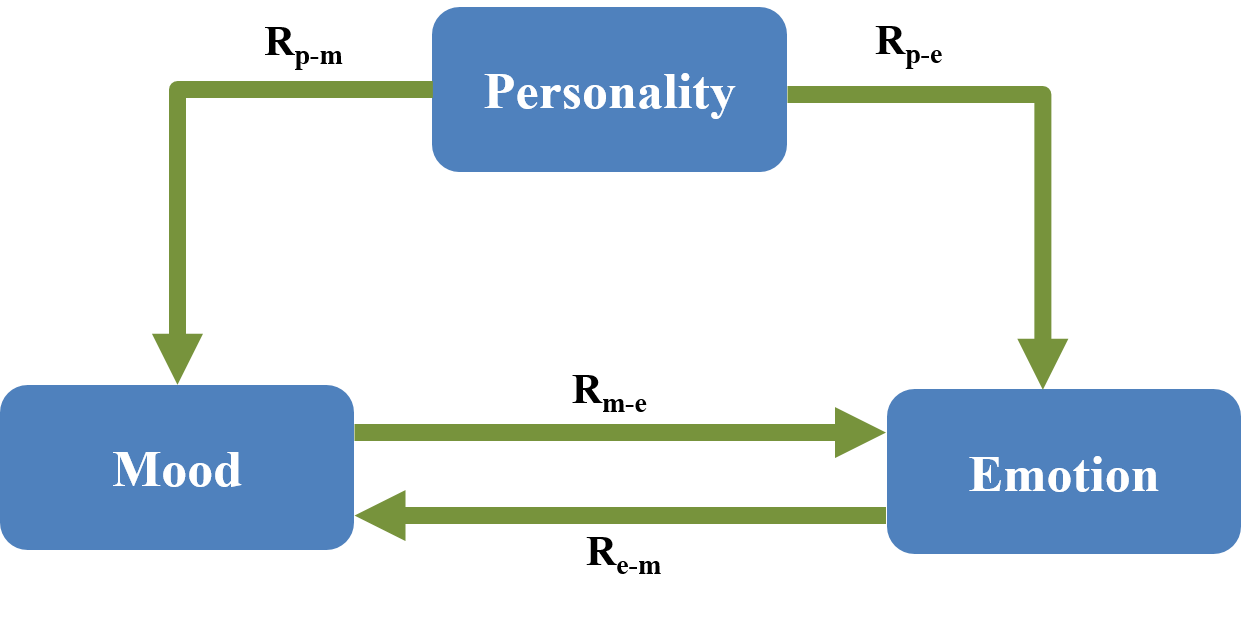}
\caption[Proposed dynamic interaction between emotion, mood and personality]{Proposed dynamic interaction between emotion, mood and personality. Adapted from \citet{Ojha2018ACS}.}
\label{fig:proposed_interaction}
\end{figure}

Researchers consider mood as accumulated effect of multiple emotional episodes \cite{Beedie2005,Ekman1994,Parkinson1996}. Moreover, mood is also said to have significant effect on emotion intensities \cite{Morris1992,Neumann2001}. These facts suggest a cyclic interaction between emotion and mood as opposed to one-way suggested by \citet{Rusting1998}. Therefore, we employ such cyclic interaction between emotion and mood as represented by Figure~\ref{fig:cyclic_interaction}. We suggest the dynamic interaction among emotion, mood and personality as shown in Figure~\ref{fig:proposed_interaction}. A new relationship ($R_{e-m}$) in addition to the ones offered by \citet{Rusting1998} can be seen in Figure~\ref{fig:proposed_interaction}.

Since personality factors do not change significantly with time \cite{Costa1988,Dweck2008}, the substantial effect of personality factors on mood ($R_{p-m}$) only occurs at the initialisation of the model \ie the initial mood state of EEGS is determined by the personality factors, which is functionally represented in \Eq~\eqref{eqn:personality_mood}. If we denote the personality dimensions of the model by $\mathbf{P}_d \in \{O, C, E, A, N\}$,  the five personality dimensions of the model, as described in Five Factor Model of personality \cite{Digman1990,Costa1992}, and mood state by $M$, the initial mood state of EEGS ($M^{initial}$) is given by the function $\mathcal{M}$ below.

\begin{equation}
\label{eqn:personality_mood}
\begin{split}
	M^{initial} & = \mathcal{M}(\mathbf{P_d})\\
	& = \mathcal{M} (O, C, E, A, N)
\end{split}
\end{equation}

The initial mood as obtained from \Eq~\eqref{eqn:personality_mood} not only takes part in mapping the appraisal variables into emotion intensities (as discussed in Sections~\ref{sec:system>appraisal-emotion_network} and \ref{sec:system>dynamic_learning}), but also directly modulates the emotion intensities. Let us denote an emotion as $e$ and its intensity as $\hat i_e$, mood state as $M$ (as previously mentioned) and extent to which a mood affects emotion as $\mathbf{M}_c$, say \emph{mood compensation}, which is a fraction of the current mood state by which the emotion intensities are influenced (see \Eq~\eqref{eqn:mood_compensation}). Now, the resulting emotion intensity after the compensating effect of mood is given by \Eq~\eqref{eqn:mood_effect}.

\begin{equation}
\label{eqn:mood_effect}
\hat i'_e = 
\begin{cases}
\hat i_e + \left|\mathbf{M}_c\right| & \mbox{if } Sign(e) = Sign (M)\\
\hat i_e - \left|\mathbf{M}_c\right| & \mbox{if } Sign(e) \neq Sign (M)
\end{cases}
\end{equation}

The value of mood compensation ($\mathbf{M}_c$) used in \Eq~\eqref{eqn:mood_effect} is given by the following formula.

\begin{equation}
\label{eqn:mood_compensation}
\mathbf{M}_c = \alpha \ M 
\end{equation}

The ration behind the formula in \Eq~\eqref{eqn:mood_effect} is that positive mood tends to increase the intensity of positive emotions and negative mood tends to increase the intensity of negative emotions. Moreover, positive mood tends to decrease the intensity of negative emotions and negative mood tends to decrease the intensity of positive emotions \cite{Neumann2001,Morris1992}. 

The parameter $\alpha$, in \Eq~\eqref{eqn:mood_compensation} should be defined as per the system need and application of the modelled system. In EEGS, we used the value of $\alpha$ as 10\% (i.e. 0.1) because this value provided the most plausible result. It is not always mandatory to use 10\% scaling factor while modelling this phenomenon. Yet, other related work also commonly demonstrate similar scaling (see, for example the work of \citet{Marinier2007}). 

Equation \ref{eqn:mood_effect} only shows how the mood state affects the emotion intensities. Two-way (cyclic) interaction of mood and emotion in EEGS (as discussed earlier -- see Figure~\ref{fig:cyclic_interaction}) should also account for the effect of the resulting emotional state on the mood. Majority of the literature has defined mood as the aggregate effect of numerous continuous emotional experiences \cite{Morris1992, Parkinson1996}. Positive emotional experience tends to shift mood towards positive scale and negative emotional experiences tend to shift mood towards negative scale. EEGS follows this notion of mood in order to model the effect of emotional experience on the subsequent mood. First of all, the impact of the emotional experience on the simulated agent (denoted as $Im$), which is a quantification of the overall influence of the emotional situation, is considered for modelling this relation. Then, an aggregate value is calculated using intensities of all the emotions that are congruent to the impact caused \ie if the impact is positive, the intensities of positively valenced emotion are considered and if the impact is negative, intensities of negatively valenced emotions are considered for the calculation of aggregate intensity. Given $n$ emotions whose valance is congruent to the impact, then, aggregate intensity ($\hat i_{agg}$) can be computed as in  \Eq~\eqref{eqn:aggregate_intensity}.

\begin{equation}
\label{eqn:aggregate_intensity}
\hat i_{agg} = \begin{cases} 
\sum\limits_{j=1}^{n}{\hat i_{e_j}} & \mbox{if } Im > 0 \\
- \sum\limits_{j=1}^{n}{\hat i_{e_j}} & \mbox{Otherwise}
\end{cases}    
\end{equation}

Once the aggregate intensity is calculated, it is converted to mood factor ($M_f$) by passing through a modified Logistic function. $M_f$ is a value in the range [-1, +1] given by \Eq~\eqref{eqn:mood_factor}.

\begin{equation}
\label{eqn:mood_factor}
M_f = \frac{2}{1+e^{-\hat i_{agg}}} \ -\ 1
\end{equation}

The value in the first part of \Eq~\eqref{eqn:mood_factor} is subtracted by 1 to shift the curve 1 step down so that the minimum output value of the function becomes -1 i.e. minimum value of the mood factor. Likewise, making the maximum value in the equation as 2 allows the resulting maximum value to be 1 since the curve is shifted one step down. This results in the value of $M_f$ to lie in the range [-1, +1].

When the mood factor is calculated, new mood is given by the formula in \eqref{eqn:new_mood}.

\begin{equation}
\label{eqn:new_mood}
M' = M + \beta \ M_f
\end{equation}

The quantity $\beta$ in \Eq~\eqref{eqn:new_mood} was chosen to be 10\% in EEGS. New mood state ($M'$) computed by above equation affects the emotion intensities in subsequent emotional experience, thereby maintaining a cyclic interaction, as shown in Figure~\ref{fig:cyclic_interaction}.

The discussion so far in this section demonstrates how EEGS implements the dynamic interaction among emotion, mood and personality.

\subsubsection{Emotion Decay}
\label{sec:system>emotion_decay}

Since the experience of emotion is believed to be instantaneous and short-lived \cite{Forgas1992,Ekman1994,Mayer1992,Rosenberg1998}, the emotions should decay over time \cite{Picard1997,Hudlicka2016}. As such, researchers have proposed various mechanisms to model the decay of emotion intensities in their computational realisations. Most common emotion decay mechanisms proposed in the literature can be categorised as (i) \emph{Linear Decay}, (ii) \emph{Exponential Decay}, (iii) \emph{Logarithmic Decay}, and (iv) \emph{Tan-Hyperbolic Decay} as summarised in Table~\ref{tbl:emotion_decay_functions}.

Majority of the emotion modelling proposals seem to have adopted linear decay function \cite{El2000,Marinier2007,Egges2004}. For example, \citet{El2000} use a decay function of the form $\hat i_e(t+1) = \phi . \hat i_e (t)$ to regulate the dynamics of positive emotions, where, $\hat i_e (t+1)$ represents the intensity of an emotion at time $t+1$, $\hat i_e (t)$ represents the intensity of the emotion at time $t$, and the quantity $\phi$ represents the decay constant that determines the slope of the decay function. For the decay of negative emotions, \citet{El2000} use a different decay constant $\delta$, where $\phi < \delta$. The rationale behind this choice is that positive emotions decay faster than negative emotions \cite{El2000}. Using trial and error, they suggest ``that there was a range of settings [for these constants] that produced a reasonable behaviour for the agent'' \cite[p.~236]{El2000}. The suggested range was $0.1 < \phi < 0.3$ and $0.4 < \delta < 0.5$. Similarly, \citet{Becker2008} uses a linear decay of the form $\hat i_e (t+1) = 1 - \frac{\hat i_e (t)}{10}$, where, $\hat i_e (t+1)$ is the intensity of emotion at time $t+1$, $\hat i_e (t)$ is the intensity of the emotion at time $t$, 10 denotes a decay time of 10 seconds.

\citet{Becker2008} also proposed exponential decay for some of the emotions. The decay function was defined as  $\hat i_e (t+1) = e^{-\hat i_e (t)}$. \citet{Dias2005} are also the proponents of exponential decay of emotion intensities. They define the emotion decay function in their model in a manner similar to that of \citet{Becker2008} in the form of $\hat i_e (t+1) = \hat i_e (t) * e^{-b.t}$, where $e^{-b.t}$ is an exponential function of time $t$ and $b$ is the constant that determines the decay rate. In addition to linear and exponential decay functions, \citet{Gebhard2003} also present a Tan-Hyperbolic function to model the decay of emotion intensities. However, the authors do not explain the internal details of the function used. We anticipate that they used a function of the form $\hat i_e (t+1) = -tanh (\hat i_e (t) - T)$, where, $T$ is the total decay time for an emotion. \citet{Hudlicka2016} describes a possibility of a Logarithmic decay function of emotions, but does not present details necessary to understand how this function might look like and how plausible the decay caused by the function can be.

\begin{table}
\caption{A summary of different emotion decay mechanisms used in various computational models of emotion.}
\label{tbl:emotion_decay_functions}       
\begin{tabular}{p{0.24\textwidth}p{0.4\textwidth}p{0.25\textwidth}}
\hline\noalign{\smallskip}
    Function & Rationale & Suggested by\\
\noalign{\smallskip}\hline\noalign{\smallskip}

    Linear & Emotion decays in a constant rate over the period & \cite{Gebhard2005,Marinier2007,Thagard2002,El2000,Becker2008,Gebhard2003,Egges2004}\\
    Exponential & Emotion intensity stays strong just after its experience and quickly decays after a short while  & \cite{Dias2005,Becker2008,Gebhard2003}\\
    Logarithmic & Emotion starts to decay rapidly as soon as it is experienced & \cite{Hudlicka2016}\\
    Tan-Hyperbolic & Emotion intensity stays strong just after its experience and quickly decays after a short while and tends to stay stable thenafter & \cite{Gebhard2003}\\

\noalign{\smallskip}\hline
\end{tabular}
\end{table}

\begin{figure}[!t]
\centering
\includegraphics[width=0.99\textwidth]{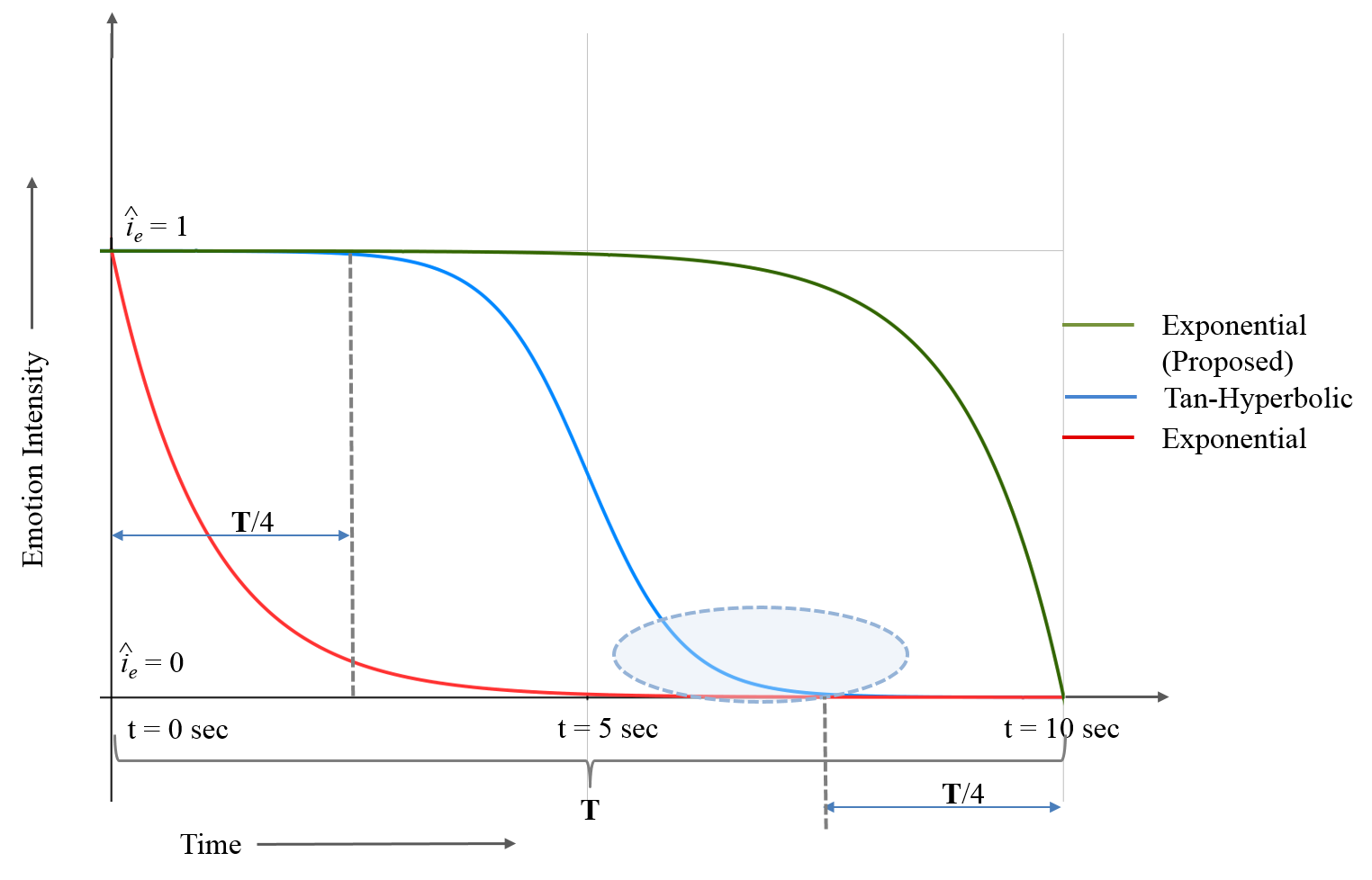}
\caption{Comparison of different emotion decay functions.}
\label{fig:emotion_decay}
\end{figure}

Since linear decay functions are probable to be less accurate to reflect the true mechanism of emotion decay in humans \cite{Hudlicka2016,Dias2005}, we opted for not implementing the linear decay function in EEGS. Logarithmic and Tan-Hyperbolic functions are less reliable because of lack of sufficient number of empirical evaluation of emotion dynamics based on these decay approaches. Therefore, we opted for exponential function to model the decay of emotions in EEGS. However, we use a slightly different approach compared to the functions proposed in existing computational models -- as exhibited by green curve in Figure~\ref{fig:emotion_decay}. This is because the emotion dynamics exhibited by commonly proposed exponential decay functions (as shown by red curve in ~Figure~\ref{fig:emotion_decay}) is questionable. If one closely examines the nature of the red curve in Figure~\ref{fig:emotion_decay}, it can be noticed that the function leads to a rapid drop of emotion intensity just after it is felt (say at time, $t$ = 0). As a result if the total decay time of an emotion is \textbf{T} seconds\footnote{Different emotions may have different length of decay \cite{Picard1997}. Here, we present a generic length of \textbf{T} seconds and leave the flexibility of choosing a decay time for an emotion intensity as empirical evidence may be available from researchers in the future.}, almost all the emotion intensity drops down within one-fourth of the total decay time (as can be seen in Figure~\ref{fig:emotion_decay}, where red curve almost falls to zero within 2.5 seconds of the triggering of the emotion, if \textbf{T}=10 sec). This clearly suggests that the normal exponential curve of the form $e^{-t}$ is not suitable for emotion decay functions. As compared to normal exponential curve, a tan-hyperbolic curve seems to be more promising to achieve a plausible decay of emotion intensities. Such a curve is represented by blue line in Figure~\ref{fig:emotion_decay}. Unlike normal exponential curve, tan-hyperbolic curve does not immediately start to rapidly decay the intensity. Rather, the intensity remains stable for a short period of time (as reasonably expected to happen in humans) and then only starts to decay with about 50\% drop in the intensity within \textbf{T}/2. However, in the second part of the decay, tan-hyperbolic curve inherits the limitation of the regular exponential function \ie remaining intensity drops rapidly almost within half of the remaining decay time (as shown by the shaded region in Figure~\ref{fig:emotion_decay}). But, (i) an emotion intensity should not start decaying immediately after it is triggered. Instead, it should stay stable for a short while and then only start to fall in exponential manner. Moreover, once the emotion intensity starts to fall, (ii) the rate of decrease in intensity should keep on increasing signifying that the effect of that emotion diminishes in increasing rate. In order to address these criteria, we propose a \emph{modified exponential emotion decay function} as defined in \Eq~\eqref{eqn:emotion_decay}.

\begin{equation}
\label{eqn:emotion_decay}
	\hat i_e (t+1)= \hat i_e (t) * \left ( 1- \frac{e^t}{e^T} \right )
\end{equation}

\noindent
Where,\newline		
$\hat i_e (t)$ is the emotion intensity at time $t$,\newline
$\hat i_e (t+1)$ is the emotion intensity at time $t+1$,\newline
$T$ is the total duration (in seconds) for the decay of an emotion intensity, and \newline
$t$ is a point in time at which an emotion intensity is calculated.

\vspace{0.5cm}

\section[Affect Regulation]{Affect Regulation\footnote{Most of the discussion in this section has been adapted from \citet{Ojha2017IJSR} and \citet{Ojha2017a}. It should be noted that although the term `affect regulation' may be used by researchers to denote broader range of processes including emotion regulation, mood regulation, etc., current research is concerned only in the process of emotion regulation.}}
\label{sec:system>emotion_convergence_and_regulation}

Since, there can be more than one emotions active at the same time \cite{Ortony1990,Scherer2001}, it is important for an autonomous agent to regulate the emotions and reach to a stable and regulated emotional state. \citet{Gross1998old,Gross1998} proposes two broad classes of emotion regulation namely (1) \emph{Antecedent-focused} emotion regulation and (2) \emph{Response-focused} emotion regulation, where antecedent-focused regulation revolves around the causes of the emotion elicitation while response-focused regulation is concerned about the handling of the already active emotions. However, the limitation of current implications of response-focused emotion regulation offered by \citet{Gross2007} is that their approach is heavily \emph{non-cognitive}. For example, they suggest that ``drugs may be used [as a regulatory device] to target physiological responses such as muscle tension'' \cite[p.~15]{Gross2007}. They also suggest exercise \cite{Thayer1994} and relaxation \cite{Suinn1971} can regulate physiological and experiential aspects of emotions. Likewise, alcohol \cite{Hull1986} and cigarettes \cite{Brandon1994} are also widely studied to modify the experience of emotion \cite{Gross1998}. Additionally, \citet{Gross1998} argues that \emph{alteration of the expressions} is the most common form of emotion regulation. Although these regulatory mechanisms may be suitable for a human being to mainly calm down negative emotional experiences, these approaches are neither well defined nor appropriate to be realised in a computational model of emotion. In the following sections, we will offer our response-focused emotion regulation mechanism that is based on higher cognitive layer of ethical reasoning and discuss its strengths compared to other emotion convergence and regulation mechanisms used in existing computational models of emotion. But, a key question is -- how can we implement such a regulatory mechanism in autonomous agents that allows them to exhibit socially acceptable emotional and behavioural responses? We will address this question in the following section.

\subsection{Emotion Convergence in Computational Models}
\label{sec:system>emotion_convergence_in_computational_models}

Literature suggests that existing computational models of emotion commonly adopt two approaches to achieve the convergence to a stable emotional state when multiple emotions are activated by the cognitive appraisal process -- either choosing the emotion with (i) \emph{Highest Intensity} \cite{Gratch2004Domain}, or obtaining a (ii) \emph{Blended Intensity} from the intensities of all the active emotions \cite{Reilly2006,Marinier2007}. EMA \cite{Gratch2004Domain} uses the approach of selecting the emotion with highest intensity to determine the final emotional state of the model. A clear disadvantage of the highest intensity approach is that the intensities of several emotions may not significantly differ. Therefore, the higher intensity of an emotion may not be the only sufficient criteria for a more effective choice in that specific situation. For example, consider a situation where \emph{joy} resulted a 0.9 intensity level, whereas distress achieved a 0.85 intensity level. By using the highest intensity approach the emotion \emph{distress} would be completely disregarded and a final emotional state of \emph{joy} with intensity of 0.9 is considered to be operational. Now, consider that the individual experiencing such emotion intensities observed a foe getting fired by their boss. Although the individual may experience slightly higher \emph{joy} for that happening, it is still convenient to regulate and suppress the emotion of \emph{joy} in favour of \emph{distress}, so to prevent an awkward situation at work. However, if we consider a situation where the individual observed a friendly colleague getting a promotion from their boss, this individual may experience similar level of \emph{joy} and \emph{distress} if that individual was also hoping for that promotion, but still regulating the emotions so to manifest \emph{joy} for the colleague's success. From these examples it is clear to see how selecting the emotion with highest intensity without considering other aspects of the situation may lead to poor decisions. In line with this, \citet{Reilly2006} argues that considering only the emotion with highest intensity causes high degree of inaccuracy in emotion processing mechanism. To address this limitation, he suggests an approach that helps in the blending of all the elicited emotions that are congruent to the situation (see \Eq~\eqref{eqn:blended_intensity}). Proponents of the emotion blending approach put forward by \citet{Reilly2006} have followed similar approach in computational models of mood and feelings \cite{Marinier2007}. Although, the approach of blending the emotion intensities proposed by \citet{Reilly2006} seems to a viable way to overcome the highest intensity approach and it is able to consider the contributions of all the constituent emotions, this approach largely fails to attribute the final emotional state to a defined emotion type. In other words, with this approach it is difficult for the individual, or agent, to describe the experienced emotional state with a specific emotion label.

 \begin{equation} 
    \label{eqn:blended_intensity}
	\hat i = 0.1 * \log_2 \sum_{n=1}^{N}2^{10*\hat i_n}
\end{equation}

\noindent where,\newline
$\hat i$ is the resulting intensity, \newline
$N$ is the number of emotions for which intensity is to be combined, \newline
$\hat i_n$ is the intensity of $n^{th}$ emotion, where, $1 \leq n \leq N$.

\vspace{0.5 cm}
In this paper, we propose to address the previously presented limitations by introducing a regulatory \emph{ethical reasoning} mechanism for the selection of a socially appropriate emotional state.

\begin{figure}[!t]
\centering
\includegraphics[width=0.9\textwidth]{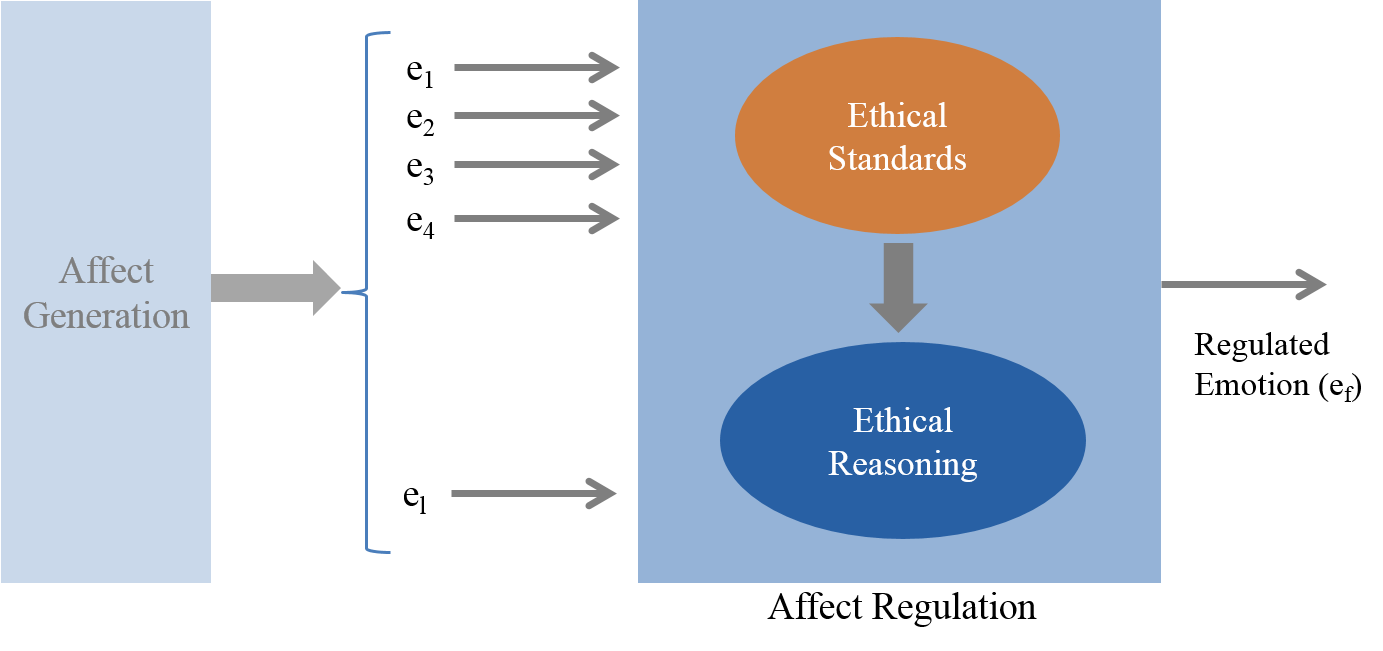}
\caption[Process of affect regulation in EEGS where conflicting emotional states are converged to a final stable and regulated emotional state based on ethical reasoning guided by ethical standards]{Process of affect regulation in EEGS where conflicting emotional states are converged to a final stable and regulated emotional state based on ethical reasoning guided by ethical standards. Redrawn after \cite{Ojha2017a}.}
\label{fig:affect_regulation}
\end{figure}

\subsection{Ethical Reasoning for Emotion Regulation in EEGS}
\label{sec:system>ethical_reasoning_for_emotion_regulation_in_EEGS}

In Section~\ref{sec:system>emotion_convergence_in_computational_models}, we argued that although the \emph{highest intensity} and \emph{blended intensity} approaches can help in converging to a single emotional state and viably assist to regulating the agent's emotions so to lead to a more believable emotional responses, they do not offer effective emotion regulation mechanism to ensure the social acceptance of the emotional responses. We suggest that an ethical reasoning mechanism can assist to converge to a single emotional state and ensure that such emotional state is socially acceptable. By saying socially acceptable emotional response we mean the ones that are in compliance with common human expectations or norms in given social situations. 

Ethical reasoning in EEGS is guided by its \emph{ethical standards}. The ethical standards are modeled as agent's beliefs. Each of these beliefs assesses the degree of approval or disapproval of a specific action directed by a source subject toward a target subject. Although Ethical standards can be generalized to accommodate any aspect of the interaction between two subjects (i.e. any action), in this section, our discussion will revolve around the use of such ethical standards for the purpose of emotion generation and expression. Therefore, the action considered in each standard will describe the expression of a specific emotion. When EEGS runs for the first time, it starts with empty standards (as discussed previously in Section~\ref{sec:system>standards}) \ie it does not have any pre-defined standard. Thus, when a person first interacts with EEGS, it establishes an initial neutral standard that guides in its emotion generation process. Ethical standards can pertain to any aspect of interaction between two persons or between an autonomous agent and a person. However, in this section, our discussion will revolve around the ethical standards in the context of emotion generation and its expression. Thus when a person first interacts with the autonomous agent running EEGS system, the agent builds a set of standards that affect the emotion processing mechanism. Suppose a stranger interacts with the agent. As stated earlier, the agent builds a set of neutral standard. Examples of the agent's standards can be -- ``we should not show anger to him'',  ``we should express joy in interacting with him'' and so on. This can be considered as what the agent believes it is supposed to do or not to do. This belief can have a certain degree depending on who the person is or what is the interaction history of the agent with the person. In other words, whether the internal standard of a robot approves the expression of an emotion to a target also has a degree associated with the approval or disapproval.


Each ethical standard in EEGS is represented as a data structure in the form {\tt (Emotion, Source, Target, Approval)}, where, {\tt Emotion}, in this specific application domain, represents the emotion addressed by the standard, {\tt Source} represents the one that expresses the emotion, {\tt Target} represents the target of the emotion expression. It is important to note that when the standard refers to an emotion the agent should or should not express toward someone, {\tt Source} assumes the value SELF. However, {\tt Source} can also assume other values to refer to other agents or individuals. By doing so, the standards describes what the agent believes about how one person (the {\tt Source}) should behave with another person (the {\tt Target}). Providing to the data structure such flexibility is a viable way to easily accommodate standards for multi-agent interaction applications, which, however, are out of scope for the present work. Moreover, {\tt Approval} denotes whether the expression of emotion is preferred or not and what is the degree of this preference. Approval is further structured as {\tt (Preference, ApprovalDegree)} (see Section~\ref{sec:system>standards}), where {\tt Preference} specifies whether the expression of emotion is preferred or not and {\tt ApprovalDegree} denotes the extent to which the expression of emotion is preferred or not. For example, the standard {\tt (\emph{anger}, SELF, JOHN, (NO, 0.75))} represents ``we should NOt express \emph{anger} to JOHN'' from the agent's perspective and the degree of this belief is 0.75. Similarly, {\tt (\emph{anger}, PAUL, DAVID, (YES, 0.9))} represents ``It is okay (YES) for PAUL to express \emph{anger} to DAVID'' and the degree of this belief is 0.9.

As addressed in Section~\ref{sec:system>standards}, it should be noted that the notion of standards in EEGS is not static quantity. Even though the autonomous agent starts the interaction with neutral standards, the standards change in the course of interaction depending on how the person interacts with the agent. Recall the example of a standard in the previous paragraph -- {\tt (\emph{anger}, SELF, JOHN, (NO, 0.75))}. As per the standard, the agent (SELF) is not supposed to express anger towards JOHN. However, if JOHN constantly misbehaves with the agent, the standards adapt to the new situation by becoming more negative and ultimately leading to the agent's belief that it is fine to express anger towards JOHN. Using ethical standards enables EEGS to converge to more socially acceptable emotional states, thus regulating the agent's emotions by ensuring more conscious and ethical emotional responses.  Table~\ref{tbl:set_of_standards} shows some examples of the standards in the memory of EEGS related to the emotion \emph{anger}.

\begin{table}
\caption{An example of a set of ethical standards for \emph{anger} emotion. Adapted from \citet{Ojha2017IJSR}. }
\label{tbl:set_of_standards}       
\begin{tabular}{p{0.15\textwidth}p{0.15\textwidth}p{0.15\textwidth}p{0.2\textwidth}p{0.15\textwidth}}
\hline\noalign{\smallskip}
    Emotion & Source & Target & Preference & Degree\\
\noalign{\smallskip}\hline\noalign{\smallskip}
    \emph{anger} & SELF & JOHN & NO & 0.8 \\
    \emph{anger} & PAUL & JOHN & YES & 0.25 \\
    \emph{anger} & DAVID & JOHN & NO & 0.5 \\
\noalign{\smallskip}\hline
\end{tabular}
\end{table}

\subsection{Reasoning Mechanism in EEGS}
\label{sec:system>reasoning_mechanism_in_EEGS}

In the current implementation, EEGS is able to generate ten different emotions in response to an event. In a particular situation, one or more emotions might be triggered in reaction to the event \cite{Ortony1990}. As discussed previously, an autonomous agent must be able to converge to a final emotional state which can then be expressed through a behavioural responses congruent to the appraised emotional situation. In Section~\ref{sec:system>ethical_reasoning_for_emotion_regulation_in_EEGS}, we discussed how such emotional response should not only be believable, but also socially acceptable. To address the gaps in existing models of emotion generation, we provided a data structure able to store ethical standards. This data structure is the foundation for the modeling of a higher cognitive layer of ethical reasoning in EEGS \cite{Ojha2017a}. In this paper, we insist that when there are multiple emotions triggered by an event at the same time by the appraisal of an event \cite{Ortony1990}, an ethical reasoning process can assist an emotion generation model to converge to a stable emotional state that is not only plausible for the given situation, but also socially acceptable. In the remainder of this section we will present the computational details to model such ethical reasoning process in EEGS.

We introduce the term \emph{Coefficient of Standard} (CoS), which is the measure of positive significance of all the standards related to an emotion being considered. This coefficient is calculated for only the standards in which the person interacting with the agent (SELF) is represented as {\tt Target}. In other words, CoS is a cumulative value of the signed approval degrees for the expression of an emotion by all (including SELF) towards the person currently interacting with the agent itself. For example, let us consider the standards in Table~\ref{tbl:set_of_standards}. If JOHN is currently interacting with the robot and \emph{anger} is one of the elicited emotions, then the coefficient of standard for the \emph{anger} emotion is computed as the average approval degree of all the standards of \emph{anger} emotion where JOHN is the target.

Suppose, there are $n$ elicited emotions from which the most appropriate final emotional state is to be determined. If there are $N$ standards related to the $j^{th}$ emotion :  $1 \leq j \leq n$ and we denote the degree of approval of $i^{th}$ standard as $d_{a_i}$ : $1 \leq i \leq N$, and preference associated with a standard as $pref$, then, the coefficient of standard of the $j^{th}$ emotion is given by \Eq~\eqref{eqn:cos}. 

\begin{equation}
\label{eqn:cos}
CoS_j = \frac{\mathlarger \sum_{i=1}^{N}{ }
    \begin{cases}
        d_{a_i},\ \ \ \mbox{ if } \ pref = ``YES"\\
        - d_{a_i},\  \mbox{ if } \ pref = ``NO"\\
    \end{cases}}{N}    
\end{equation}

\Eq~\eqref{eqn:cos} shows that coefficient of standard is the average of signed approval degree for the expression of the $j^{th}$ emotion from all the recognised persons (including "SELF") to the person interacting with the agent. This, in fact, measures how much the internal standards of the agent support the expression of an emotion. For example, if a standard has preference "YES" then it is okay to express the emotion -- hence the positive summation in \Eq~\eqref{eqn:cos}. Likewise, if a standard has preference "NO" then it is not okay to express the emotion -- hence the negative summation in \Eq~\eqref{eqn:cos}. As such, the higher the coefficient of standard (including sign), the better the emotion for expression in the given social context.

The notion of the concepts of deontological and consequentialist ethics is efficiently captured by the formula in \Eq~\eqref{eqn:cos}. The formula considers the duties in the form of standards of the agent, thereby capturing the essence of deontological ethics \cite{Alexander2007,Robbins2007}. All the standards related to each emotion are considered for the computation of coefficient of standard. Moreover, in addition to the standards related to itself, the agent also considers the standards related to other recognised persons and the person interacting to the agent (see Table~\ref{tbl:set_of_standards} for example). By doing this, the agent becomes able to address the consequence of the expression of a particular emotion on the target as well as other related persons, thereby capturing the notion of consequentialist ethics as well. 

However, considering only the internal standards for the determination of final emotional state may still lead to unethical or socially unacceptable emotions. For example, consider a person who is really nice and has done plenty of good things to you. Many other people also have positive thoughts about the person and have high regards for the person. Naturally, as per the standard, expressing anger to such a person should be discouraged. Nevertheless, there can be situations where an angry or aggressive response is the most appropriate reaction in response to an action of such a person -- say he tries to stab your best friend with a knife. You would definitely become angry and respond in defensive and aggressive manner even if you had high standards for the person. In order to overcome the presented limitation of expressing unethical emotional responses, we also consider emotion intensities elicited by the appraisal process together with the coefficients of standard of each emotion.

As such, we compute a numeric quantity denoted with \emph{Quantified Emotion} to take into account the degree and intensity of the elicited emotions. If we denote the degree of valence of $j^{th}$ emotion by $d_{v_j}$ and the intensity of $j^{th}$ emotion as $\hat{i}_j$, then the quantified value of the $j^{th}$ emotion is given by \eqref{eqn:qe}.

\begin{equation}
\label{eqn:qe}
QE_j =  d_{v_j} \ * \  \hat{i}_j
\end{equation}

Now, the absolute value of the $j^{th}$ quantified emotion is multiplied to its corresponding coefficient of standard to compute the \emph{Coefficient of Ethics (CoE)} as shown in \eqref{eqn:coe}. The reason for using absolute value of $QE_j$ is to avoid the undesirable sign change when the signed value of $CoS_j$ is multiplied by signed value of $QE_j$. This helps to consider only the strength of the emotion based on its degree and intensity (without any regards to its sign).

\begin{equation}
\label{eqn:coe}
CoE_j =  CoS_j\ * \  \left| QE_j \right|
\end{equation}

When the coefficient of ethics for each elicited emotion is computed, the \emph{emotion with the highest value of coefficient of ethics is selected} as the most ethical emotional state in the given situation. In other words, the $CoS$ acts as a regulation mechanism based on ethics to assist the selection of more socially acceptable emotional responses.

In order to test the validity of our claim that ethical reasoning in EEGS can help an autonomous agent to reach to a socially appropriate emotional state, we compared the emotion dynamics of EEGS using three different approaches to reach to final emotional state, which were introduced in Section~\ref{sec:system>emotion_convergence_in_computational_models} as (i) \emph{Highest Intensity Approach} -- where the emotion with the highest intensity is considered as the final emotional state, (ii) \emph{Blended Intensity Approach} -- where the intensities of the elicited emotions are blended to determine a new intensity value and a final emotion type to be attributed, and (iii) \emph{Ethical Reasoning Approach} -- the proposed approach where the final emotional state is determined by reasoning ethically, which we presented earlier in this section. The details of the result can be found in our published work \cite{Ojha2017IJSR,Ojha2017a}.

\section{A Guideline for the Implementation of EEGS Modules}
\label{sec:system>a_guideline_for_the_implementation}

Previous sections in this paper described the computational details of the presented emotion model EEGS. In this section, we will describe how the computational details have been realised as an implemented model. These details are expected to provide a better understanding of how the proposed model can be implemented and replicated by other researchers for the purpose of comparison and bench-marking. 

Overall computational model is implemented in Java Programming language with embedded Apache Derby Database\footnote{A reader may find more information about Apache Derby at https://db.apache.org/derby} for the storage and retrieval of data. These were our personal implementation choices. A reader should be able to achieve a successful implementation of the model described in this paper using other languages or frameworks. 

Since the proposed model consists of four main `processing' modules namely (1) emotion elicitation module, (2) cognitive appraisal module, (3) affect generation module and (4) affect regulation module, the subsections below will follow discussion of the implementation details of the given modules in the same order.

\subsection{Implementing the Emotion Elicitation Module}
\label{sec:system>implementing_emotion_elicitation_module}

The emotion elicitation module in EEGS represents the first-order non-cognitive appraisal of the situation \cite{Lambie2002} leading to an experience of valenced bodily reaction denoting the positivity or negativity of the event \cite{James1884,Lange1885}. In EEGS, the emotion elicitation module is realised more on a functional level than computational. In other words, in the current model, the underlying complexities of the mechanism to compute first-order phenomenological reaction \cite{Lambie2002} is not implemented. The valenced reaction for a particular event in the given context was obtained from a survey data where people were asked to rate the positivity or negativity of the given action in the given context (see \cite{Ojha2017a} for details on how the data was collected). As such, the actions in a particular experimental scenario are assigned the average score provided by the survey participants and that score is considered as the first-order phenomenological reaction \cite{Lambie2002}. Thus   the input to the emotion elicitation module is a valenced score for the action in the given context.

Alternatively, instead of relying on the average scores assigned to the actions, it is also possible to employ machine learning techniques to enable a system to learn how to map contextual information of the emotional event into a signed numeric score that can be fed to the emotion elicitation module. The lack of an appropriate level of details on how to computationally realise this module is a limitation of the present study. However, the main objective of our discussion is to investigate the cognitive aspects of an emotional process and the data collection methodology used to overcome this limitation was sufficient to test the behaviour of our model. Although it is possible to computationally realise EEGS model without introducing emotion elicitation module, this notion was provided to allow an easy investigation of the aspect for future researchers.

\subsection{Implementing the Cognitive Appraisal Module}
\label{sec:system>implementing_cognitive_appraisal_module}

The cognitive appraisal module takes the first-order phenomenological reaction (which represents the contextual information about the event) as the input and computes the appraisal variables with the help of goals, standards and attitudes by means of the formulas provided in Section~\ref{sec:system>computation_of_appraisal_variables} for each considered appraisal variable (see Section~\ref{sec:system>goals_standards_and_attitudes} for discussion on how goals, standards and attitudes are defined in EEGS). The table below summarises the inputs and output of the cognitive appraisal module.

\begin{table}
\caption{Input(s) and output(s) of cognitive appraisal module. }
\label{tbl:io_cognitive_appraisal_module}       
\begin{tabular}{p{0.35\textwidth}p{0.25\textwidth}p{0.3\textwidth}}
\hline\noalign{\smallskip}
        Input Parameter(s) & Supporting Parameter(s) & Computed/Output Parameters(s)\\
\noalign{\smallskip}\hline\noalign{\smallskip}
    (1) Action Scores (first-order phenomenological reaction \cite{Lambie2002}) & (1) Goals \newline (2) Standards \newline (3) Attitudes & (1) Set of Appraisal Variables \\
\noalign{\smallskip}\hline
\end{tabular}
\end{table}

The set of appraisal variables computed by the cognitive appraisal module are provided as input to the affect generation module where each appraisal variable may be linked to more than one emotion intensities. However, before the computed appraisal variables are sent to the affect generation module, the values of these variables need to be normalised in a particular range, which is explained below.

\subsubsection{Normalisation of Appraisal Variables}
\label{sec:system>normalisation_of_appraisal_variables}

The appraisal variables computed using the formulas discussed in Section~\ref{sec:system>computation_of_appraisal_variables} may sometimes lead to unexpected values that lie outside the expected range (see Table~\ref{tbl:appraisal_variables_and_value_ranges} for expected range of values). Therefore it is important to normalise the values obtained by the given formulas in the specified range. We use a modified Logistic function to obtain the appraisal values in the specified range.

\begin{equation}
\label{eqn:appraisal_variable_normalisation}
    normalised\_appraisal = \frac{range\_gap}{1+e^{-m \ * \ (appraisal - midpoint)}} \ + \ \gamma
\end{equation}

\noindent
Where,\newline
$normalised\_appraisal$ is the normalised value of appraisal variable,\newline
$range\_gap$ is the gap between the min and max expected value of the appraisal variable,\newline
$m$  is the slope of part of the curve where it exhibits linear mapping,\newline
$appraisal$ is the non-normalised value of the appraisal variable, \newline
$midpoint$ is the mid-point is the mid value of the appraisal value range, and \newline
$\gamma$ is the offset used to shift the value of the Logistic function up or down as per the requirement. A positive value of $\gamma$ shifts the normalised value up and a negative value shifts it down.

\begin{table}
\caption{Appraisal variables in EEGS and their value ranges.}
\label{tbl:appraisal_variables_and_value_ranges}       
\begin{tabular}{p{0.35\textwidth}p{0.25\textwidth}}
\hline\noalign{\smallskip}
        Appraisal Variable & Expected Value Range \\
\noalign{\smallskip}\hline\noalign{\smallskip}

    \emph{goal conduciveness} & [-1, 1]\\
    \emph{desirability} & [-1, 1]\\
    \emph{praiseworthiness} & [-1, 1]\\
    \emph{appealingness} & [-1, 1]\\
    \emph{deservingness} & [-1, 1]\\
    \emph{familiarity} & [0, 1]\\
    \emph{unexpectedness} & [0, 1]\\

\noalign{\smallskip}\hline
\end{tabular}
\end{table}

\begin{figure}[!t]
\centering
\includegraphics[width=0.99\textwidth]{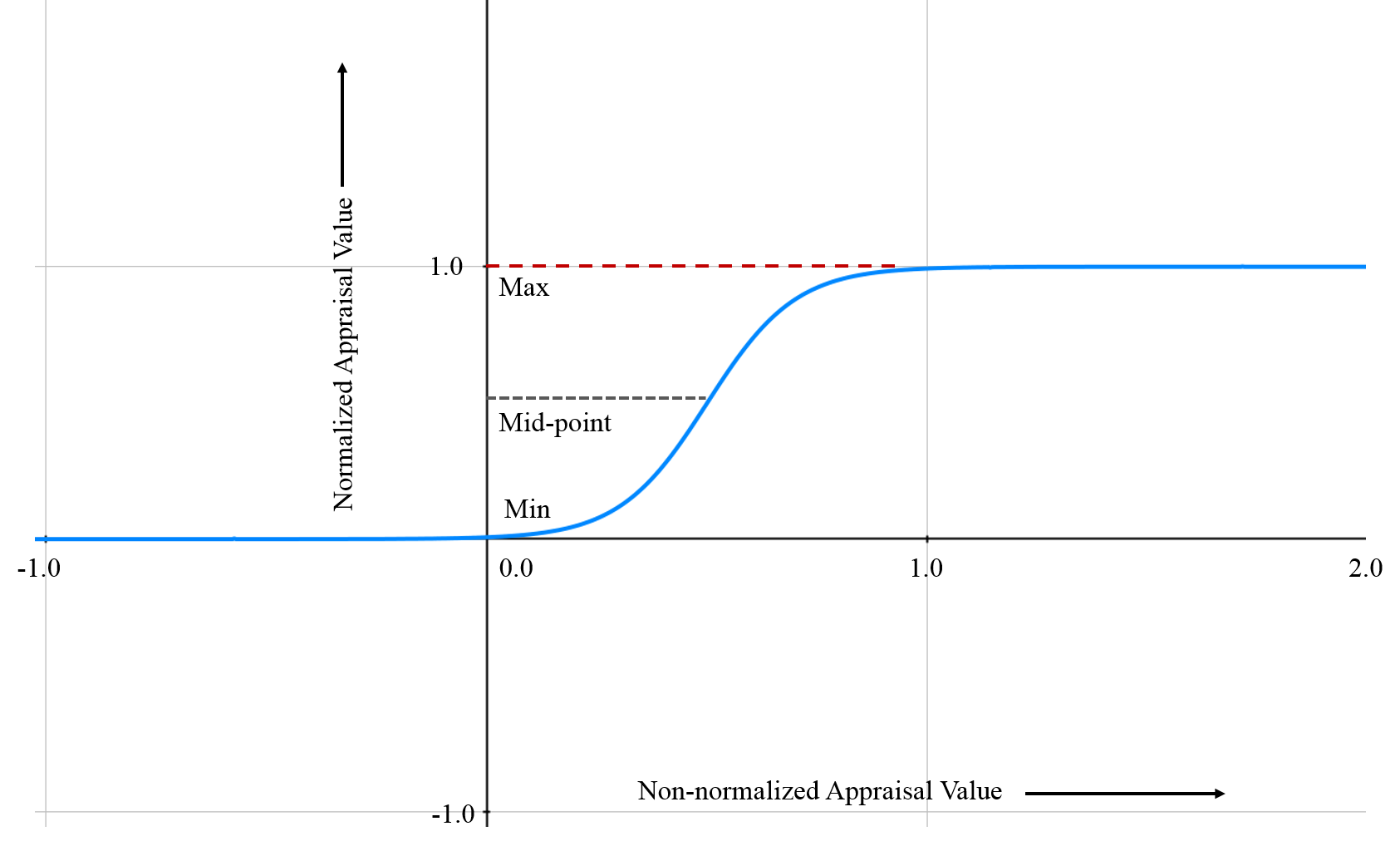}
\caption{Normalisation function for appraisal variables in the range [0,1].}
\label{fig:norm_0_1}
\end{figure}

\begin{figure}[!t]
\centering
\includegraphics[width=0.99\textwidth]{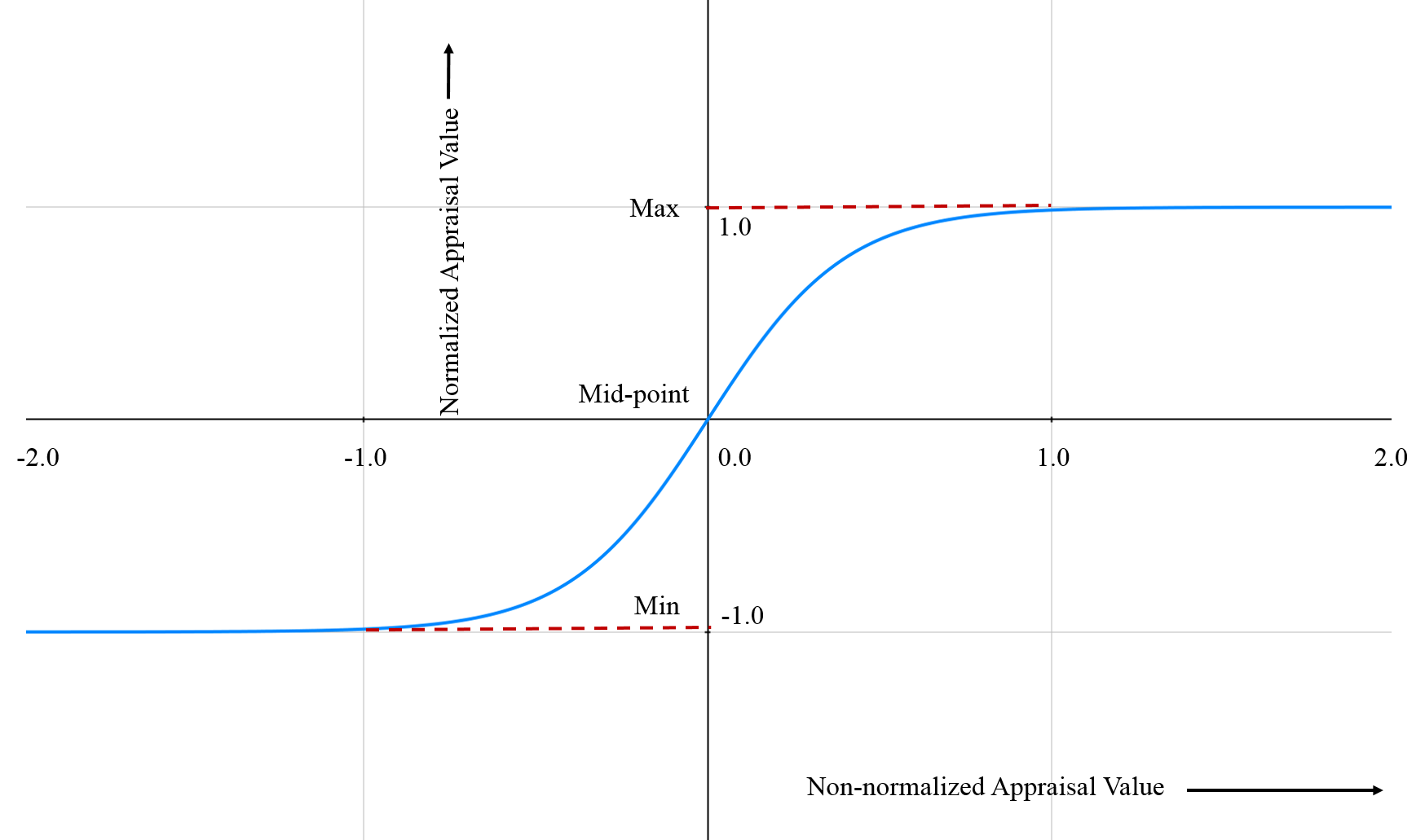}
\caption{Normalisation function for appraisal variables in the range [-1,1].}
\label{fig:norm_-1_1}
\end{figure}

Figure~\ref{fig:norm_0_1} shows the logistic function to normalise the computed appraisal variables that are supposed to be in the range [0,1] \ie \emph{familiarity} and \emph{unexpectedness} ($range\_gap$ = 1, $m$ = 10, $midpoint$ = 0.5 and $\gamma$ = 0). Horizontal axis represents the value of the appraisal variable before normalisation and the vertical axis represents the value of the appraisal variable after normalisation. In case of the appraisal variables lying in the range [0,1], the midpoint of mapping should lie at 0.5 (as shown in Figure~\ref{fig:norm_0_1}). Similarly, Figure~\ref{fig:norm_-1_1} shows the normalisation function for the appraisal variables lying in the range [-1,1], where $range\_gap$ = 2, $m$ = 5, $midpoint$ = 0.0 and $\gamma$ = -1. The reason for using the value of $range\_gap$ as 2 is because the the difference between 1 and -1 is 2. Slope ($m$) is slightly lower to maintain a consistent mapping. $\gamma$ is set to -1 in order to shift the curve 1 step below and making $midpoint$ to be zero.

\begin{figure}[!t]
\centering
\includegraphics[width=0.99\textwidth]{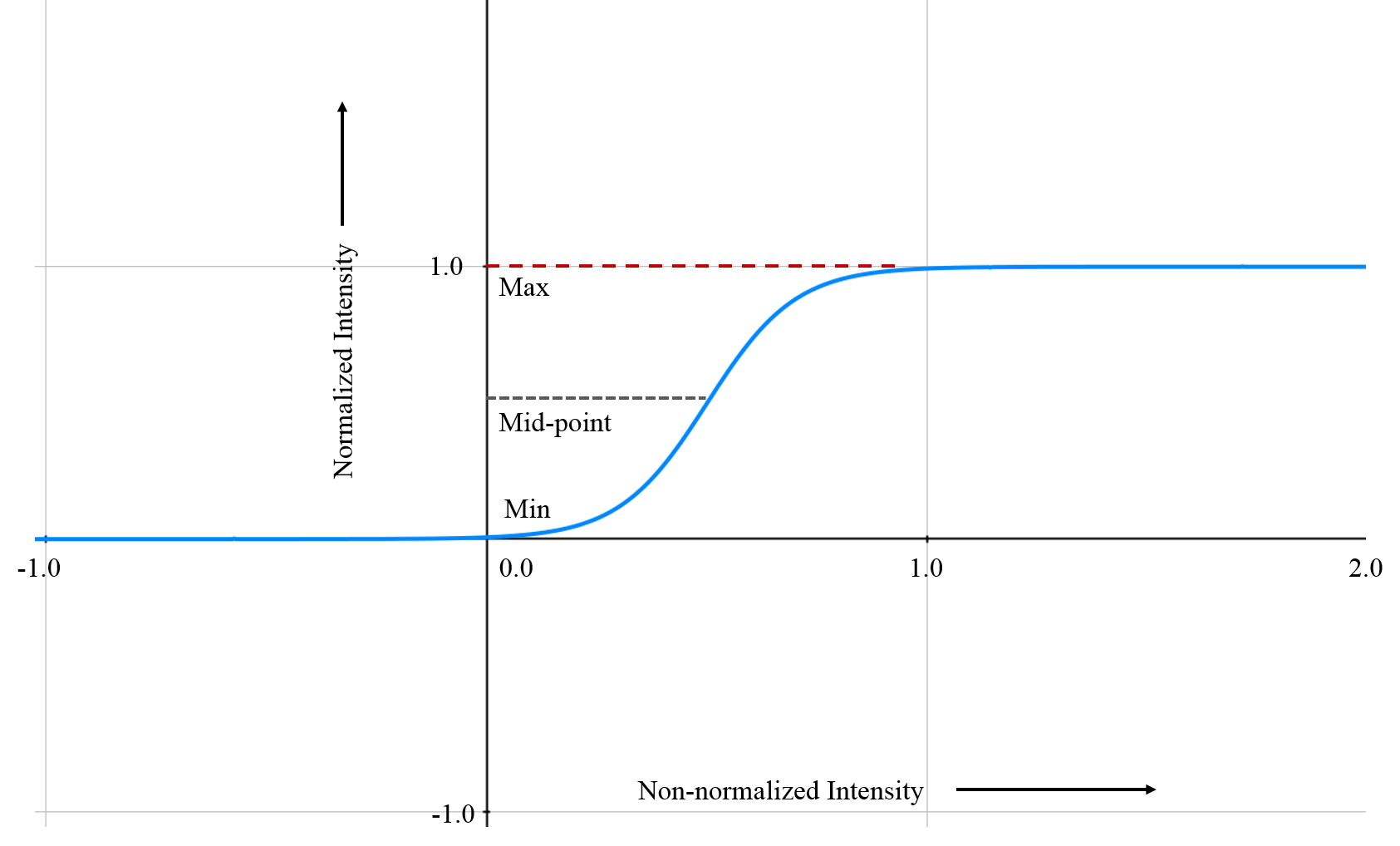}
\caption{Normalisation function for emotion intensities.}
\label{fig:intensity_norm}
\end{figure}

\begin{table}
\caption{Input(s) and output(s) of affect generation module.}
\label{tbl:io_affect_generation_module}       
\begin{tabular}{p{0.35\textwidth}p{0.25\textwidth}p{0.25\textwidth}}
\hline\noalign{\smallskip}
       Input Parameter(s) & Supporting Parameter(s) & Computed/Output Parameters(s)\\
\noalign{\smallskip}\hline\noalign{\smallskip}

    (1) Set of Appraisal Variables & (1) Mood\newline (2) Personality & (1) Set of Emotions\newline Mood State \\

\noalign{\smallskip}\hline
\end{tabular}
\end{table}

\subsection{Implementing the Affect Generation Module}
\label{sec:system>implementing_affect_generation_module}

The affect generation module mainly deals with mapping the computed appraisal variables into emotion intensities and also alters the mood state. As such, affect generation module takes the set of appraisal variables output from cognitive appraisal module as an input and computes intensities of various emotions with the help of personality and mood factors that was used to learn the association of appraisal variables to emotions (see Sections\ref{sec:system>appraisal-emotion_network} and ~\ref{sec:system>dynamic_learning} for details on how such an association is determined in EEGS). The formulas presented between \Eqs~\ref{eqn:joy_intensity}--\ref{eqn:disliking_intensity} are used to calculate the emotion intensities of various emotions. Moreover, two-way interaction between emotion and mood was realised by implementing the formulas presented in the \Eqs~\ref{eqn:personality_mood}--\ref{eqn:new_mood}. Table~\ref{tbl:io_affect_generation_module} summarises the input and output parameters of the affect generation module.

\subsubsection{Emotion Intensity Update and Normalisation}
\label{sec:system>emotion_intensity_update_and_normalisation}

In contrast to some other models that discard the previous intensity of an emotion once new appraisals are performed \cite{Dias2014}, EEGS opts for an incremental approach where every new appraisal does not undo the effect of prior appraisals on emotion intensities but rather performs an increment or decrement on the intensity based on current appraisal -- in line with the suggestions of \citet{Scherer2001}. This allows EEGS to preserve the true dynamics of emotional experience and support effectively in the mood update process in a natural manner. However, the drawback of such an incremental approach is that it may lead to ever increasing intensity of an emotion, when the subsequent events are congruent to the valence of the emotion. This kind of ever increasing intensity may lead to incoherent behaviour of the agent. Therefore, a suitable mechanism should be employed to ensure that the intensity of a particular emotion always lies in the standard range of [0,1]. In order to achieve this, EEGS normalises the computed emotion intensities using the modified Logistic function similar to the one in \Eq~\eqref{eqn:appraisal_variable_normalisation}.

\begin{equation}
\label{eqn:emotion_intensity_normalisation}
    \hat i^{norm}_{e} = \frac{range\_gap}{1+e^{-m \ * \ (\hat i_e - midpoint)}} \ + \ \gamma
\end{equation}

\noindent
Where,\newline
$\hat i^{norm}_{e}$ is the normalised value of emotion intensity,\newline
$range\_gap$ is the gap between the min and max expected value of the intensity,\newline
$m$  is the slope of part of the curve where it exhibits linear mapping,\newline
$\hat i_e$ is the non-normalised value of the emotion intensity, \newline
$midpoint$ is the mid-point is the mid value of the intensity range, and \newline
$\gamma$ is the offset used to shift the value of the Logistic function up or down as per the requirement. A positive value of $\gamma$ shifts the normalised value up and a negative value shifts it down.

The function in equation \ref{eqn:emotion_intensity_normalisation} and Figure~\ref{fig:intensity_norm} help in normalising an emotion intensity to a stable value in the range [0, 1].

\begin{table}
\caption{Input(s) and output(s) of affect regulation module.}
\label{tbl:io_affect_regulation_module}       
\begin{tabular}{p{0.2\textwidth}p{0.3\textwidth}p{0.3\textwidth}}
\hline\noalign{\smallskip}
       Input Parameter(s) & Supporting Parameter(s) & Computed/Output Parameters(s)\\
\noalign{\smallskip}\hline\noalign{\smallskip}

    (1) Set of Emotions & (1) (Ethical) Standards & (1) Regulated Emotional State \\

\noalign{\smallskip}\hline
\end{tabular}
\end{table}

\subsection{Implementing the Affect Regulation Module}
\label{sec:system>implementing_affect_regulation_module}

As discussed in Section~\ref{sec:system>emotion_convergence_and_regulation}, multiple active emotional states generated by affect generation module need to be regulated and converged to a stable emotional state to allow socially acceptable behaviour by the agent. This is achieved by taking the emotions output from affect generation module and determining a final emotional state with the help of ethical standards using the \Eqs~\ref{eqn:cos}--\ref{eqn:coe} and selecting the emotional state with highest coefficient of ethics (see Section~\ref{sec:system>ethical_reasoning_for_emotion_regulation_in_EEGS} for details on the formulas and computational mechanism provided).

The regulated emotional state obtained as the output of affect regulation module can be considered to control other cognitive components, if implemented in intelligent systems affected by emotions such as decision systems \cite{Holtzman1988}.

\section{Model Evaluation}
\label{sec:model_evaluation}

As previously discussed, EEGS undergoes three main stages of processing involving (i) \emph{calculation of appraisal variables} to appraise the emotion eliciting stimulus, (ii) \emph{mapping the appraisal variables into the intensities of various emotions}, and (iii) \emph{reaching to a stable emotional state by regulating the active emotions}. Therefore, we propose an approach that allows the evaluation of each of these processing stages individually, which we call a \emph{3-Stage Evaluation} of a computational model of emotion.

\begin{figure}[!t]
\centering
\includegraphics[width=0.8\textwidth]{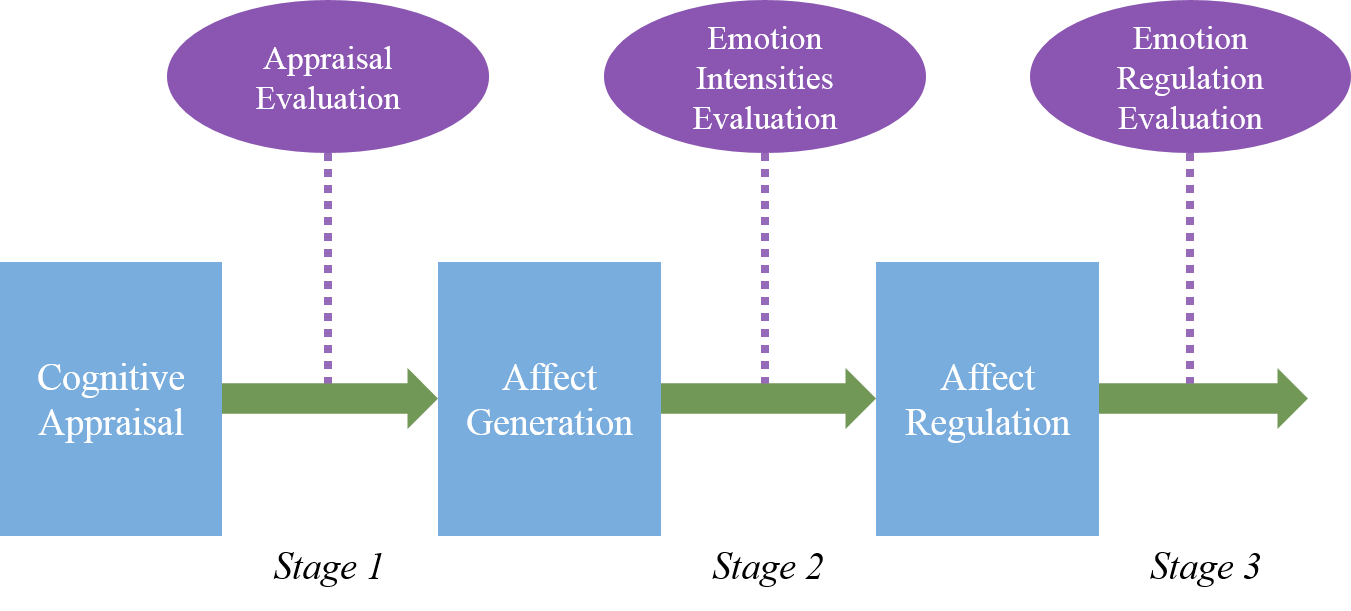}
\caption{Proposed \emph{3-Stage Evaluation} approach for computational models of emotion.}
\label{fig:3-stage_evaluation}
\end{figure}

\fig~\ref{fig:3-stage_evaluation} provides a graphical visualisation of the proposed 3-stage evaluation approach used to evaluate our computational model of emotion -- EEGS. Stage 1 evaluation measures the accuracy of the computation of different appraisal variables in EEGS and also examines if EEGS can compute appraisal variables in multiple domains without the need of changing the model's rules and parameters or not. Stage 2 evaluation measures whether the appraisal variables are mapped accurately into emotion intensities and how the factors of mood and personality affect generated emotion dynamics. Finally, Stage 3 evaluation measures whether the final emotional state reached by EEGS represents human-like emotions and  enhances the social acceptance of the emotion or not.

In this paper, our goal was to present the computational details of our emotion model, EEGS, making it a transparent model of emotions and enabling other researchers to replicate the behaviour of our model and benchmark with existing or new models thereby leading to the further evolution of the emotion modelling field. Therefore, we have not devoted much space in this paper to the evaluation of the various aspects of our model. However, we would like to redirect our readers to our published work for a discussion of the evaluation of various components of the presented emotion model. Our paper published in Annual Meeting on Advances in Cognitive Systems (ACS) \cite{Ojha2017b} discusses how the appraisal mechanism in EEGS is performed in a domain-independent manner for some representative appraisal variables of OCC theory \cite{Ortony1990} (Stage 1 evaluation). In a paper published in International Conference on Autonomous Agents and Multi-Agent Systems (AAMAS) \cite{Ojha2019AAMAS}, we present a new benchmark on the emotion intensity computation accuracy with a overall accuracy of about 80\%, which is much higher compared to previous attempts \cite{Meuleman2013} (Stage 2 evaluation). We also present an analysis of how the personality and mood factors can influence the emotion dynamics of EEGS model in a paper published in ACS (see \cite{Ojha2018ACS} for more details) (Stage 2 evaluation). We also present an analysis of how the ethical reasoning mechanism in EEGS allows the model to reach to a regulated and socially acceptable emotional state in our publishes work (see \citet{Ojha2017a} and \citet{Ojha2017IJSR} for details) (Stage 3 evaluation).

\section{Summary and Conclusion}

This paper presented a detailed description of various modules of the proposed Ethical Emotion Generation System (EEGS). The discussion followed the sequence of theoretical steps involved in the process of emotion generation and regulation. Also, the description of the processes in various modules was accompanied by complete technical/mathematical representation of the corresponding aspect.

The paper started with a brief overview of EEGS in \sect~\ref{sec:system>EEGS_ethical_emotion_generation_system} leading to the discussion of the overall process of appraisal dynamics and emotion generation as shown in \fig~\ref{fig:appraisal_dynamics_revisited}. Then, in \sect~\ref{sec:system>overall_system_srchitecture}, we presented an overall architecture of EEGS, which is composed of various modules as shown in \fig~\ref{fig:system_architecture_EEGS}, namely (i) \emph{Emotion Elicitation Module}, (ii) \emph{Cognitive Appraisal Module}, (iii) \emph{Memory Module}, (iv) \emph{Characteristics Module}, (v) \emph{Affect Generation Module}, and (vi) \emph{Affect Regulation Module}. As explained previously, the \emph{emotion elicitation module} performs the first-order \cite{Lambie2002} lower-level \cite{Scherer2001} appraisal of the event that does not involve conscious cognitive processing. This process is followed by the computation in \emph{cognitive appraisal module} where different appraisal variables are calculated based on the goals, standards and attitudes of the agent stored in \emph{memory module}. The process of mapping the appraisal variables into emotion intensities in \emph{affect generation module} is modulated by several factors like personality and mood present in the \emph{characteristics module}. The multiple emotions activated by the cognitive appraisal process are converged to a stable state and regulated by the \emph{affect regulation module}.

The discussion was then followed by the technical description of \emph{events}, \emph{actions} and \emph{objects} in relation to the \emph{emotion elicitation} process was presented in \sect~\ref{sec:system>events_actions_and_objects}. Then the \emph{cognitive appraisal} process in EEGS was presented where the concepts of \emph{goals}, \emph{standards} and \emph{attitudes} were introduced. \emph{Goals}  defined  a set of states an individual wants to achieve. \emph{Standards} in EEGS represent the beliefs of the agent regarding various actions between any two agent (including itself). An standard encourages or prevents an agent from performing some action to another agent. Additionally, an standard defines what an agent believes what kind of actions can be performed by one agent to another agent. \emph{Attitudes} of an agent reflect what it feels about an object/person. An agent can have positive attitude towards an external agent if is has positive experience of interaction with the agent and negative attitude otherwise. These goals, standards and attitudes play a central role in the computation of appraisal variables \cite{Ortony1990}. We present the details of a novel mechanism for the computation of appraisal variables in \sect~\ref{sec:system>computation_of_appraisal_variables}. While other models of emotion compute appraisal variables based on domain-specific rules \cite{Aylett2005,Dias2005,Velasquez1997}, we use a domain-independent mechanism to compute the appraisal variables in EEGS (details of the formulae for the calculation of individual appraisal variable can be found in \sect~\ref{sec:system>computation_of_appraisal_variables}).

After the completion of all the appraisal processes (which run in parallel as explained in \sect~\ref{sec:system>parallel_processing_of_appraisals}), the quantitative values of appraisal variables are mapped into emotion intensities. This mapping process in EEGS is influenced by factors like mood \cite{Morris1992,Neumann2001} and personality \cite{Corr2008,Revelle1995,Watson1997}. EEGS implements a machine learning approach to dynamically determine the \emph{weight of association} of various appraisal variables to emotion intensities based on personality and mood factors, the details of which is presented in \sect~\ref{sec:system>dynamic_learning}. These weights and quantitative values of various associated appraisal variables are used to compute the intensities of various emotions in EEGS (see \sect~\ref{sec:system>from_appraisal_to_emotion_intensities} for technical details). EEGS not only realises the effect of mood and personality on emotion, but also operationalises the influence of personality on mood and effect of emotion intensities on the mood dynamics of the agent -- as previously discussed in \sect~\ref{sec:system>revisiting_interaction_among_emotion_mood_and_personality}, \fig~\ref{fig:cyclic_interaction} and \fig~\ref{fig:proposed_interaction}. Since, the experience of emotion is often believed to be short-lived, EEGS employs a mechanism to decay the intensity of emotion after a certain duration. Several emotion decay functions have been put forward by researchers namely (i) \emph{Linear} \cite{Becker2008,Egges2004,El2000,Gebhard2005}, (ii) \emph{Exponential} \cite{Becker2008,Dias2005}, (iii) \emph{Logarithmic} \cite{Hudlicka2016}, and (iv) \emph{Tan-Hyperbolic} \cite{Gebhard2003}. In \fig~\ref{fig:emotion_decay}, we presented a comparison of various emotion decay strategies and proposed a more plausible and realistic decay function in the form of a modified exponential as shown in \Eq~\eqref{eqn:emotion_decay}.

Because of multiple associations between appraisal variables and emotions, more than one emotional states can be active as a result of cognitive appraisal process \cite{Ortony1990,Scherer2001}. Due to this reason, an agent should go through two important processes to yield an expected emotional response -- (i) \emph{emotion convergence} and (ii) \emph{emotion regulation}. Emotion convergence is the processing of reaching to a stable emotional state when more than one conflicting emotional states are active at the same time. Emotion regulation is the process of ensuring that the experience of emotion helps in achieving personal as well as social benefits \cite{Gross1998}. Although previous proposals offered (i) \emph{highest intensity} approach \cite{Gratch2004Domain} or (ii) \emph{blended intensity} approach \cite{Reilly2006} for emotion convergence, we opted to adopt an (iii) \emph{ethical reasoning} approach to achieve the goals of both the emotion convergence as well as regulation (where the former two approaches are not able to achieve the goal of emotion regulation).

We hope that the presentation of the internal details of our computational model of emotion, EEGS, will allow researchers to replicate our emotion model and compare and benchmark the appraisal and emotion generation/regulation mechanism with emerging models of emotion. We also believe that the mathematical formulation presented in this paper may have limitations because of complex nature of emotion and related processes. However, we take this as an opportunity for allowing the researchers and experts in the field for deeper investigation into the matter and make a significant step towards the further evolution of the field of `emotion modelling' -- a rather slowly advancing area compared to other areas of artificial intelligence.   

\acknowledgement
We would like to acknowledge that the content of this paper has been extracted from the doctoral dissertation of Suman Ojha submitted to the University of Technology Sydney (UTS), Australia. This research was supported by an Australian Government Research Training Program Scholarship.

\bibliographystyle{spbasic}      
\bibliography{references.bib}   

\end{document}